\documentclass[a4paper,11pt]{article}

\usepackage[english]{babel}
\usepackage[utf8]{inputenc}
\usepackage{lmodern}

\usepackage{color,amsmath,amsthm,amssymb,mathtools,nicefrac,bbm,graphicx}

\usepackage{stackengine}

\usepackage{tikz}
\usetikzlibrary{matrix,chains,positioning,decorations.pathreplacing,arrows}
\usetikzlibrary{shapes,arrows}
\tikzset{font={\fontsize{9pt}{12}\selectfont}}
\usepackage{adjustbox}

\usepackage[us]{datetime}

\DeclareFontEncoding{LS1}{}{}
\DeclareFontSubstitution{LS1}{stix}{m}{n}
\DeclareMathAlphabet{\mathscr}{LS1}{stixscr}{m}{n}

\usepackage[margin=1in]{geometry}

\usepackage{hyperref}
\hypersetup{colorlinks=true,linkcolor=blue}

\usepackage[shortlabels,inline]{enumitem}
\usepackage[capitalise]{cleveref} % clever reference

\crefname{enumi}{item}{items}
\crefname{equation}{}{}

\newcommand{\N}{\mathbb{N}}

\newcommand{\R}{\mathbb{R}}

\DeclareMathOperator{\rank}{rank}

\newcommand{\ind}{\mathbbm{1}}

\newcommand{\interior}[1]{{\kern0pt#1}^{\mathrm{o}}}
\newcommand*\diff{\mathop{}\!\mathrm{d}}

\newcommand{\is}{\curvearrowleft}
\newcommand{\smallsum}{\textstyle \sum}

\newcommand{\smallbigcap}{\textstyle \bigcap}
\newcommand{\smallbigcup}{\textstyle \bigcup}

\newcommand{\Exists}{\exists\,}

\newcommand{\mc}[1]{\mathcal{#1}}
\newcommand{\mf}[1]{\mathfrak{#1}}

\newcommand{\ms}[1]{\mathscr{#1}}

\newcommand{\inp}{d}
\newcommand{\width}{\mathfrak{h}}
\newcommand{\dimension}{\mathfrak{d}}

\newcommand{\scalprod}[2]{\cfadd{def:scalar_product_norm} \langle #1, #2 \rangle}
\newcommand{\bscalprod}[2]{\cfadd{def:scalar_product_norm} \big \langle #1, #2 \big \rangle}

\newcommand{\eucnorm}[1]{\cfadd{def:scalar_product_norm} \lVert #1 \rVert}

\newcommand{\half}{\cfadd{def:half_spaces_hyperplane}\mc H}
\newcommand{\hyper}{\cfadd{def:half_spaces_hyperplane}\mc G}

\makeatletter
\ExplSyntaxOn
\seq_new:N \g_abbrs
\prop_new:N \g_abbr_counts
\tl_new:N \l_abbr_count_tl

\NewDocumentCommand{\abbr}{m m O{#1} m m O{#4} m}{
	\expandafter\newcommand\csname#3\endcsname[1][]{
		\seq_if_in:NnTF \g_abbrs {#1} {
			\prop_get:NnN \g_abbr_counts {#1} \l_abbr_count_tl
			\prop_gput:Nnx \g_abbr_counts {#1} {\int_eval:n {\l_abbr_count_tl + 1}}
			\hyperref[#1]{#7}
		} {
			\seq_gput_left:Nn \g_abbrs {#1}
			\prop_gput:Nnn \g_abbr_counts {#1} {1}
			\expandafter\gdef\csname#1@def\endcsname{#2}
			\phantomsection\label{#1}
			\str_if_eq:nnTF{##1}{}{\emph{#2}}{##1}~(\hyperref[#1]{#7})
		}
	}
	\expandafter\newcommand\csname#6\endcsname[1][]{
		\seq_if_in:NnTF \g_abbrs {#1} {
			\prop_get:NnN \g_abbr_counts {#1} \l_abbr_count_tl
			\prop_gput:Nnx \g_abbr_counts {#1} {\int_eval:n {\l_abbr_count_tl + 1}}
			\hyperref[#1]{#4}
		} {
			\expandafter\gdef\csname#1@def\endcsname{#5}
			\seq_gput_left:Nn \g_abbrs {#1}
			\prop_gput:Nnn \g_abbr_counts {#1} {1}
			\phantomsection\label{#1}
			\str_if_eq:nnTF{##1}{}{\emph{#5}}{##1}~(\hyperref[#1]{#4})
		}
	}
}

\ExplSyntaxOff
\makeatother

\abbr{ANN}{artificial neural network}{ANNs}{artificial neural networks}{ANN}
\abbr{ReLU}{rectified linear unit}{ReLUs}{rectified linear units}{ReLU}
\abbr{GF}{gradient flow}{GFs}{gradient flows}{GF}
\abbr{GD}{gradient descent}{GDs}{gradient descents}{GD}
\abbr{SGD}{stochastic gradient descent}{SGDs}{stochastic gradient descents}{SGD}
\abbr{MSE}{mean squared error}{MSEs}{mean squared errors}{MSE}

\DeclareFontFamily{U}{matha}{\hyphenchar\font45}
\DeclareFontShape{U}{matha}{m}{n}{
<-6> matha5 <6-7> matha6 <7-8> matha7
<8-9> matha8 <9-10> matha9
<10-12> matha10 <12-> matha12
}{}
\DeclareSymbolFont{matha}{U}{matha}{m}{n}

\DeclareFontFamily{U}{mathx}{\hyphenchar\font45}
\DeclareFontShape{U}{mathx}{m}{n}{
<-6> mathx5 <6-7> mathx6 <7-8> mathx7
<8-9> mathx8 <9-10> mathx9
<10-12> mathx10 <12-> mathx12
}{}
\DeclareSymbolFont{mathx}{U}{mathx}{m}{n}

\DeclareMathDelimiter{\vvvert} {0}{matha}{"7E}{mathx}{"17}%

\newcommand{\lipnorm}[2]{\cfadd{def:lipschitz_norm} \vvvert #1 \vvvert_{#2}}
\newcommand{\plipnorm}[1]{\vvvert #1 \vvvert}

\DeclareFontFamily{OMX}{MnSymbolE}{}

\DeclareFontShape{OMX}{MnSymbolE}{m}{n}{
    <-6>  MnSymbolE5
   <6-7>  MnSymbolE6
   <7-8>  MnSymbolE7
   <8-9>  MnSymbolE8
   <9-10> MnSymbolE9
  <10-12> MnSymbolE10
  <12->   MnSymbolE12}{}
\DeclareFontShape{OMX}{MnSymbolE}{b}{n}{
    <-6>  MnSymbolE-Bold5
   <6-7>  MnSymbolE-Bold6
   <7-8>  MnSymbolE-Bold7
   <8-9>  MnSymbolE-Bold8
   <9-10> MnSymbolE-Bold9
  <10-12> MnSymbolE-Bold10
  <12->   MnSymbolE-Bold12}{}

\DeclareSymbolFont{largeMN}  {OMX}{MnSymbolE}{m}{n}

\DeclareMathDelimiter{\lwavy}{\mathopen}{largeMN}{'136}{largeMN}{'136}
\DeclareMathDelimiter{\rwavy}{\mathclose}{largeMN}{'136}{largeMN}{'136}
\DeclareMathDelimiter{\lWavy}{\mathopen}{largeMN}{'137}{largeMN}{'137}
\DeclareMathDelimiter{\rWavy}{\mathclose}{largeMN}{'137}{largeMN}{'137}

\newcommand{\hoeldernorm}[3]{\cfadd{def:hoelder_norm} \stackanchor[0pt]{$\lwavy$}{$\lwavy$} {#1} \stackanchor[0pt]{$\rwavy$}{$\rwavy$}_{#2,#3}}
\newcommand{\sobnorm}[3]{\cfadd{def:sobolev-slobodeckij_norm} \stackanchor[0pt]{$\lWavy$}{$\lWavy$} {#1} \stackanchor[0pt]{$\rWavy$}{$\rWavy$}_{#2,#3}}

\newcommand{\abs}[1]{\lvert #1 \rvert}
\newcommand{\babs}[1]{\bigl\lvert #1 \bigr\rvert}
\newcommand{\bbabs}[1]{\Bigl\lvert #1 \Bigr\rvert}

\newcommand{\qandq}{\quad\text{and}\quad}
\newcommand{\qqandqq}{\qquad\text{and}\qquad}

\theoremstyle{plain}
\newtheorem{theorem}{Theorem}[section]
\newtheorem{lemma}[theorem]{Lemma}
\newtheorem{prop}[theorem]{Proposition}
\newtheorem{cor}[theorem]{Corollary}

\newtheorem{setting}[theorem]{Setting}
\theoremstyle{definition}
\newtheorem{definition} [theorem]{Definition}

\usepackage{xparse}
\usepackage{etoolbox}

\NewDocumentEnvironment{cproof}{m}
{\begin{proof}[Proof of \cref{#1}]}%
{\noindent The proof of \cref{#1} is thus complete.
\end{proof}}

\NewDocumentEnvironment{cproof2}{m}
{\begin{proof}[Proof of \cref{#1}]}%
{\noindent This completes the proof of \cref{#1}.
\end{proof}}

\ExplSyntaxOn

\seq_new:N \g_cflist_loaded
\seq_new:N \g_cflist_pending

\NewDocumentCommand{\cfadd}{ m }
{
  \seq_if_in:NnF \g_cflist_loaded { #1 } {
    \seq_if_in:NnF \g_cflist_pending { #1 } {
      \seq_gput_right:Nn \g_cflist_pending { #1 }
    }
  }
}

\NewDocumentCommand{\cfload}{ o }
{
  \seq_if_empty:NTF \g_cflist_pending {\unskip} {
    (cf.\ \cref{\seq_use:Nn \g_cflist_pending {,}})\IfValueTF{#1}{#1~}{\unskip}
    \seq_gconcat:NNN \g_cflist_loaded \g_cflist_loaded \g_cflist_pending
    \seq_gclear:N \g_cflist_pending
  }
}

\NewDocumentCommand{\cfclear} {} {
  \seq_gclear:N \g_cflist_loaded
  \seq_gclear:N \g_cflist_pending
}

\NewDocumentCommand{\cfout}{ o }
{
  \seq_if_empty:NTF \g_cflist_pending {\unskip} {
    (cf.\ \cref{\seq_use:Nn \g_cflist_pending {,}})\IfValueTF{#1}{#1~}{\unskip}
    \seq_gclear:N \g_cflist_pending
  }
}

\NewDocumentCommand{\ifnocf} { m } {
  \seq_if_empty:NT \g_cflist_pending { #1 }
}

\NewDocumentCommand{\cfconsiderloaded}{ m }{

  \seq_gput_right:Nn \g_cflist_loaded {#1}

}

\ExplSyntaxOff

\ExplSyntaxOn

\bool_new:N \g_noteobserve

\NewDocumentCommand{\nobs}{}{
  \bool_if:nTF { \g_noteobserve } {
    \bool_gset_false:N \g_noteobserve
    note~
  } {
    \bool_gset_true:N \g_noteobserve
    observe~
  }
}

\NewDocumentCommand{\Nobs}{}{
  \bool_if:nTF { \g_noteobserve } {
    \bool_gset_false:N \g_noteobserve
    Note~
  } {
    \bool_gset_true:N \g_noteobserve
    Observe~
  }
}

\ExplSyntaxOff

\ExplSyntaxOn

\bool_new:N \g_hencetherefore

\NewDocumentCommand{\hence}{}{
  \bool_if:nTF { \g_hencetherefore } {
    \bool_gset_false:N \g_hencetherefore
    hence~
  } {
    \bool_gset_true:N \g_hencetherefore
    therefore~
  }
}

\NewDocumentCommand{\Hence}{}{
  \bool_if:nTF { \g_hencetherefore } {
    \bool_gset_false:N \g_hencetherefore
    Hence,~we~obtain~
  } {
    \bool_gset_true:N \g_hencetherefore
    Therefore,~we~obtain~
  }
}

\ExplSyntaxOff

\ExplSyntaxOn

\int_new:N \g_furthermore

\NewDocumentCommand{\Moreover}{ o o }{
  \IfValueT{#1}{
    \str_case:nn {#1} {
      {Furthermore} {\int_set:Nn {\g_furthermore} {0}}
      {Moreover} {\int_set:Nn {\g_furthermore} {1}}
      {In~addition} {\int_set:Nn {\g_furthermore} {2}}
      {note} {\bool_gset_true:N \g_noteobserve}
      {observe} {\bool_gset_false:N \g_noteobserve}
    }
    \IfValueT{#2}{
      \str_case:nn {#2} {
        {Furthermore} {\int_set:Nn {\g_furthermore} {0}}
        {Moreover} {\int_set:Nn {\g_furthermore} {1}}
        {In~addition} {\int_set:Nn {\g_furthermore} {2}}
        {note} {\bool_gset_true:N \g_noteobserve}
        {observe} {\bool_gset_false:N \g_noteobserve}
      }
    }
  }
  \int_case:nn { \int_mod:nn {\g_furthermore} {3} } {
    { 0 } { Furthermore,~\nobs that}
    { 1 } { Moreover,~\nobs that}
    { 2 } { In~addition,~\nobs that}
  }
  \int_incr:N \g_furthermore
  \IfValueF{#1}{~}
}

\ExplSyntaxOff

\ExplSyntaxOn

\prop_const_from_keyval:Nn \l__verbs
{
show = shows ,
imply = implies ,
demonstrate = demonstrates ,
prove = proves ,
establish = establishes ,
ensure = ensures ,
assure = assures
}

\seq_const_from_clist:Nn \g_prove_mru {
   establish,
   demonstrate,
   prove,
   show,
   imply,
   ensure
}

\tl_new:N \g_wordtmp
\seq_new:N \l_mytmps

\cs_generate_variant:Nn \str_if_in:nnTF { nVTF }

\NewDocumentCommand{\prove}{ o }{
   \IfValueTF{#1}{
     \seq_clear:N \l_mytmps
     \seq_map_inline:Nn \g_prove_mru {
       \str_if_eq:nnTF {##1} {ensure} {
         \str_set:Nn \l_temps {n}
       } {
         \str_set:Nx \l_temps {\str_head_ignore_spaces:n {##1}}
       }
       \str_if_in:nVTF {#1} \l_temps {
         \seq_put_right:Nn \l_mytmps {##1}
       } { }
     }
     \seq_get_right:NN \l_mytmps \g_wordtmp
   } {
     \seq_get_right:NN \g_prove_mru \g_wordtmp
   }
   \tl_use:N \g_wordtmp
   \seq_gput_left:NV \g_prove_mru \g_wordtmp
   \seq_gremove_duplicates:N \g_prove_mru
   \IfValueF{#1}{~}
}

\NewDocumentCommand{\proves}{ o }{
   \IfValueTF{#1}{
     \seq_clear:N \l_mytmps
     \seq_map_inline:Nn \g_prove_mru {
       \str_if_eq:nnTF {##1} {ensure} {
         \str_set:Nn \l_temps {n}
       } {
         \str_set:Nx \l_temps {\str_head_ignore_spaces:n {##1}}
       }
       \str_if_in:nVTF {#1} \l_temps {
         \seq_put_right:Nn \l_mytmps {##1}
       } { }
     }
     \seq_get_right:NN \l_mytmps \g_wordtmp
   } {
     \seq_get_right:NN \g_prove_mru \g_wordtmp
   }
   \str_set:NV \l_tmpa_str \g_wordtmp
   \prop_get:NVN \l__verbs \l_tmpa_str \l_tmpa_tl
   \tl_use:N \l_tmpa_tl
   \seq_gput_left:NV \g_prove_mru \g_wordtmp
   \seq_gremove_duplicates:N \g_prove_mru
   \IfValueF{#1}{~}
}

\ExplSyntaxOff

\ExplSyntaxOn

\NewDocumentEnvironment{flexmath}{ m o }{
  \str_if_eq:noTF {a} {#1} {
    \begin{equation}
    \IfValueT{#2}{\label{eq:\loc.#2}}
    \begin{aligned}
  } {
    \catcode`&=9
    \renewcommand{\\}{}
    \str_if_eq:noTF {d} {#1} {
      \begin{equation}
      \IfValueT{#2}{\label{eq:\loc.#2}}
    } {
      \begin{math}
    }
  }
}{
  \str_if_eq:noTF {i} {#1} {
    \end{math}
    \catcode`&=4
  } {
    \str_if_eq:noTF {d} {#1} {
      \end{equation}
    } {
      \end{aligned}
      \end{equation}
    }
  }
}

\ExplSyntaxOff

\begin{document}

\title{On bounds for norms of reparameterized ReLU artificial \\ neural network parameters: sums of fractional powers of \\ the Lipschitz norm control the network parameter vector}

\author{Authors}

\author{Arnulf Jentzen$^{1,2}$ and Timo Kr\"oger$^{3}$\bigskip\\
\small{$^1$ School of Data Science and School of Artificial Intelligence,} \vspace{-0.1cm}\\
\small{The Chinese University of Hong Kong, Shenzhen (CUHK-Shenzhen),}\vspace{-0.1cm}\\
\small{Shenzhen, China; e-mail: \texttt{ajentzen}\textcircled{\texttt{a}}\texttt{cuhk.edu.cn}}\smallskip\\
\small{$^2$ Applied Mathematics: Institute for Analysis and Numerics,}\vspace{-0.1cm}\\
\small{Faculty of Mathematics and Computer Science, University of M\"unster,}\vspace{-0.1cm}\\
\small{M\"unster, Germany; e-mail: \texttt{ajentzen}\textcircled{\texttt{a}}\texttt{uni-muenster.de}}\smallskip\\
\small{$^3$ Applied Mathematics: Institute for Analysis and Numerics,}\vspace{-0.1cm}\\
\small{Faculty of Mathematics and Computer Science, University of M\"unster,}\vspace{-0.1cm}\\
\small{M\"unster, Germany; e-mail: \texttt{timo.kroeger}\textcircled{\texttt{a}}\texttt{uni-muenster.de}}}

\date{November 14, 2025}
%\date{\today}

\maketitle

\vspace{-0.6cm}

\begin{abstract}
\vspace{0.2cm}
It is an elementary fact in the scientific literature that the Lipschitz norm of the realization function of a feedforward fully connected rectified linear unit (ReLU) artificial neural network (ANN) can, up to a multiplicative constant, be bounded from above by sums of powers of the norm of the ANN parameter vector. Roughly speaking, in this work we reveal in the case of shallow ANNs that the converse inequality is also true. More formally, we prove that the norm of the equivalence class of ANN parameter vectors with the same realization function is, up to a multiplicative constant, bounded from above by the sum of powers of the Lipschitz norm of the ANN realization function (with the exponents $ \nicefrac{1}{2} $ and $ 1 $). Moreover, we prove that this upper bound only holds when employing the Lipschitz norm but does neither hold for Hölder norms nor for Sobolev-Slobodeckij norms. Furthermore, we prove that this upper bound only holds for sums of powers of the Lipschitz norm with the exponents $ \nicefrac{1}{2} $ and $ 1 $ but does not hold for the Lipschitz norm alone.
\end{abstract}

\vspace{-0.28cm}

\tableofcontents

\section{Introduction}
\label{sec:introduction}

In recent years, \ANNs\ have become an extremely powerful tool for tackling a wide variety of complex tasks, such as recognizing natural language, handwritten text, or objects in images, as well as controlling motor vehicles or robotic devices in general. Although \GD\ optimization schemes have often proven to be highly effective for training \ANNs\ in practice, it remains a fundamental open problem in research to rigorously prove under which conditions \GD\ optimization schemes converge or diverge. However, there are several promising mathematical analysis approaches in the scientific literature that provide a step in this area of research and prove the convergence of various optimization schemes under suitable assumptions. In the following, we want to briefly outline some of the findings in a selection of these works and we refer to the references mentioned below for details and further reading.

One of the most well-known and fundamental results in the field of time-continuous \GD\ optimization methods goes back to \L ojasiewicz~\cite{Lojasiewicz1984}, in which it was shown that a non-divergent solution of a \GF\ associated with a real analytic risk function (which is often referred to as the energy function in the context of \GFs) converges to a single limit point. The basic idea is to prove that for real analytic risk functions the so-called \L ojasiewicz inequality holds and, using this, to control the length of non-divergent \GF\ trajectories around their limit points (see also Absil et al.~\cite[Section 2]{AbsilMahonyAndrews2005}). This argument was extended, for example, in Bolte et al.~\cite{BolteDaniilidis2006} to a broad class of nonsmooth risk functions by replacing the differential with a subdifferential, so that the convergence of bounded \GF\ trajectories of corresponding subgradient dynamical systems could be shown.

Furthermore, there are several results in the scientific literature that employ \L ojasiewicz's original idea and analyze time-discrete descent methods. In particular, in Attouch \& Bolte~\cite[Theorem~1]{AttouchBolte2009} it was shown that every bounded sequence generated by a proximal algorithm, applied to a risk function that satisfies the \L ojasiewicz inequality around its generalized critical points, converges to a generalized critical point. This abstract convergence result was further extended in Attouch et al.~\cite[Theorems 3.2, 4.2, 4.3, 5.1, 5.3, 5.6, and 6.2]{AttouchBolteSvaiter2013} to achieve various convergence results for bounded sequences of descent methods such as inexact gradient methods, inexact proximal algorithms, forward-backward splitting algorithms, gradient projection methods, and regularized Gauss-Seidel methods satisfying sufficient decrease assumptions and allowing a relative error tolerance. In addition, in Absil et al.~\cite{AbsilMahonyAndrews2005} there are abstract convergence results for analytic risk functions and non-divergent sequences generated by general time-discrete descent methods.

Several convergence results can be applied to the training of \ANNs\ using \GD\ optimization schemes. Specifically, under suitable assumptions, in the context of training \ANNs\ with finitely many training data, it was shown that every limit point of a bounded sequence generated by the stochastic subgradient method is a critical point of the risk function and that the risk function values converge (see Davis et al.~\cite[Corollary 5.11]{Davis2020stochastic}). Moreover, in Dereich \& Kassing~\cite{JML-3-3} the convergence of bounded stochastic gradient descent schemes was studied, in particular, in the case of deep \ANNs\ with an analytic activation function, compactly supported input data, and compactly supported output data. In addition, in Jentzen \& Riekert~\cite[Theorem~1.3]{JML-1-2} (cf. Eberle et al.~\cite[Theorem~1.2]{EberleJentzenRiekertWeiss2021ERA}) it was recently proved that every non-divergent \GF\ trajectory in the training of deep \ANNs\ with \ReLU\ activation, under the assumption that the unnormalized probability density function and the target function are piecewise polynomial, converges with a strictly positive rate of convergence to a generalized critical point in the sense of the limiting Fr\'{e}chet subdifferential. In the case of constant target functions in the training of deep \ANNs\ with \ReLU\ activation, the boundedness and convergence for \SGD\ processes were demonstrated (see Hutzenthaler et al.~\cite{Hutzenthaler2021} and the references mentioned therein). We also want to mention results in the area of inertial Bregman proximal gradient methods and block coordinate descent methods with a possibly variable metric (cf., e.g., Mukkamala et al.~\cite{MR4131349}, Ochs~\cite{Ochs2019}, Xu \& Yin~\cite{XuYin2013}, and Zeng et al.~\cite{pmlr-v97-zeng19a}). For additional references on \GD\ optimization schemes, we refer, for example, to the overview articles Bottou et al.~\cite{doi:10.1137/16M1080173}, E~et al.~\cite{CSIAM-AM-1-561}, and Ruder~\cite{Ruder2016}.

In view of these scientific findings and the frequently made assumption that the sequence of \ANN\ parameter vectors generated by the optimization method is bounded, it is a key contribution of this article to discover a new relationship between norms of \ANN\ parameter vectors and sums of powers of the Lipschitz norm of the \ANN\ realization function. More formally, it is an elementary fact in the scientific literature that the Lipschitz norm of the realization function of a deep rectified linear unit \ANN\ with $ L \in \N $ many affine linear transformations can, up to a multiplicative constant, be bounded from above by sums of powers of the norm of the \ANN\ parameter vector with the exponents $ 1 $ and $ L $ (cf., e.g., Beck et al.~\cite[Corollary 2.37]{doi:10.1142/S021902572150020X} and Miyato et al.~\cite[Section 2.1]{arxiv.1802.05957}). Roughly speaking, in this work we reveal in the case of shallow \ANNs\ that the converse inequality is also true (but with the exponents $ \nicefrac{1}{2} $ and $ 1 $ instead of $ 1 $ and $ 2 $). While the inequality that the Lipschitz norm of the realization function of shallow \ANNs\ can be controlled by sums of powers of the norm of the \ANN\ parameter vector is an elementary fact (see \cref{lem:inverse:bound} below), the converse inequality (see \cref{eq:thm:positive} below) is non-trivial and has a much more involved proof. To illustrate this converse inequality in a more accurate form, we now present the first main result of our article, \cref{thm:positive} below, and we refer to Subsection~\ref{subsec:equivalence} below for more explicit estimates.

\vspace{0.2cm}

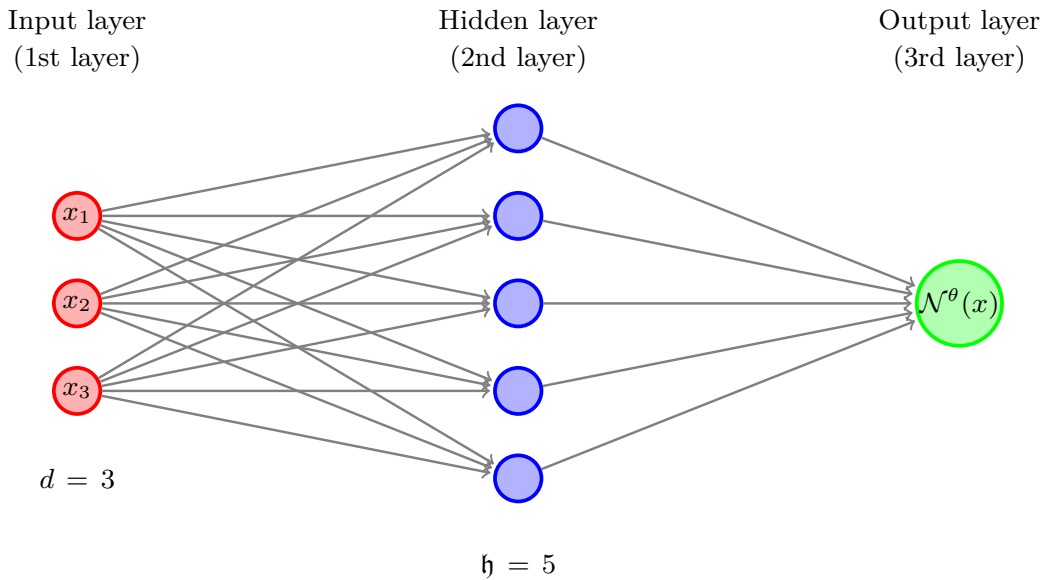
\begin{figure}[h!]
	\centering
	\begin{adjustbox}{width=\textwidth}
		\begin{tikzpicture}[shorten >=1pt,->,draw=black!50, node distance=\layersep]
		\tikzstyle{every pin edge}=[<-,shorten <=1pt]
		\tikzstyle{input neuron}=[very thick, circle,draw=red, fill=red!30, minimum size=15pt,inner sep=0pt] 
		\tikzstyle{output neuron}=[very thick, circle,draw=green, fill=green!30, minimum size=20pt,inner sep=0pt]
		\tikzstyle{hidden neuron}=[very thick, circle, draw=blue, fill=blue!30, minimum size=15pt,inner sep=0pt]
		\tikzstyle{annot} = [text width=9em, text centered]
		\tikzstyle{annot2} = [text width=4em, text centered]
		\node[input neuron](I-1) at (0,-1.5 cm) { $ x_1 $ };
		\node[input neuron](I-2) at (0,-2.5 cm) { $ x_2 $ };
		\node[input neuron](I-3) at (0,-3.5 cm) { $ x_3 $ };
		\foreach \name / \y in {1,...,5}
		\path[yshift=0.5cm]
		node[hidden neuron] (H-\name) at (5 cm,-\y cm) {};      
		\path[yshift=-1cm]
		node[output neuron](O) at (10 cm,-1.5 cm) { $ \mc N^{\theta}(x) $ };   
		\foreach \dest in {1,...,5}
		{\path[line width = 0.8] (I-1) edge (H-\dest);
		\path[line width = 0.8] (I-2) edge (H-\dest);
		\path[line width = 0.8] (I-3) edge (H-\dest);}
		\foreach \source in {1,...,5}
		\path[line width = 0.8] (H-\source) edge (O);     
		\node[annot,above of=H-1, node distance=1cm, align=center] (hl) {Hidden layer\\(2nd layer)};
		\node[annot, above of=I-1, node distance = 2cm, align=center] {Input layer\\ (1st layer)};
		\node[annot, above of=O, node distance=3cm, align=center] {Output layer\\(3rd layer)};		
		\node[annot2,below of=H-5, node distance=1cm, align=center] (sl) { $ \width = 5 $ };
		\node[annot2, below of=I-3, node distance=1cm, align=center] (s2) { $ \inp = 3 $ };
		\end{tikzpicture}
	\end{adjustbox}
	\caption{Graphical illustration of the considered shallow \ANN\ architecture in \cref{thm:positive} and \cref{thm:equivalence}
             in the special case of an \ANN\ with $ \inp = 3 $ neurons on the input layer and $ \width = 5 $ neurons on the hidden layer.
             In this situation, there are $ \inp \width = 15 $ real weight parameters and $ \width = 5 $ real bias parameters for the first
             affine linear transformation from the three-dimensional input layer to the five-dimensional hidden layer, and there are
             $ \width = 5 $ real weight parameters and $ 1 $ real bias parameter for the second affine linear transformation from the
             five-dimensional hidden layer to the one-dimensional output layer.
             The total number of parameters of this \ANN\ thus satisfies $ \dimension = \inp \width + 2 \width + 1 = 26 $.
	         We have that for every \ANN\ parameter vector $ \theta = ( \theta_1, \ldots, \theta_{\dimension} ) \in \R^{\dimension} = \R^{26} $
	         the associated realization function $ \R^3 \ni x \mapsto \mc N^{\theta}(x) \in \R $ maps the three-dimensional input vector
	         $ x = ( x_1, x_2, x_3 ) \in \R^3 $ to the scalar output
	         $ \mc N^{\theta}(x) = \theta_{\dimension} + \sum_{i=1}^{5} \theta_{ \inp \width + \width + i }
	         \max \{ \theta_{ \inp \width + i } + \sum_{j=1}^{3} \theta_{ (i-1) \inp + j } x_j, 0 \} \in \R $.}
	\label{figure:architecture}
\end{figure}

\begin{samepage}
\begin{theorem}\label{thm:positive}
  Let $ \inp, \width, \dimension \in \N $, $ \ms a \in \R $, $ \ms b \in (\ms a, \infty) $ satisfy $ \dimension = \inp \width + 2 \width + 1 $,
  for every $ \theta = ( \theta_1, \ldots, \theta_{\dimension} ) \in \R^{\dimension} $ let
  $ \mc N^{\theta} \in C( [\ms a, \ms b]^{\inp}, \R ) $ satisfy for all $ x = ( x_1, \ldots, x_{\inp} ) \in [\ms a, \ms b]^{\inp} $ that
  \begin{equation}
    \mc N^{ \theta } (x) = \theta_{\dimension} + \smallsum\nolimits_{i=1}^{\width} \theta_{ \inp \width + \width + i }
    \max \bigl\{ \theta_{ \inp \width + i } + \smallsum\nolimits_{j=1}^{\inp} \theta_{ (i-1) \inp + j } x_j, 0 \bigr\},
  \end{equation}
  for every $ n \in \N $, $ x = ( x_1, \ldots, x_n ) \in \R^n $ let $ \eucnorm{ x } = ( \smallsum_{j=1}^n \abs{ x_j }^2 )^{1/2} \in \R $,
  let $ z \in [\ms a, \ms b]^{\inp} $, and for every $ f \colon [\ms a, \ms b]^{\inp} \to \R $ let
  $ \plipnorm{ f } = \abs{ f ( z ) } + \sup\nolimits_{ x, y \in [\ms a, \ms b]^{\inp}, \, x \neq y }
    \nicefrac{ \abs{ f(x) - f(y) } }{ \eucnorm{ x - y } } \in [0,\infty] $.
  Then there exist $ \ms c, \ms C \in \R $ such that for all $ \theta \in \R^{\dimension} $ there exists $ \vartheta \in \R^{\dimension} $
  such that $ \mc N^{\vartheta} = \mc N^{\theta} $ and
  \begin{equation}\label{eq:thm:positive}
    \eucnorm{ \vartheta }
    \le \ms c \bigl( \plipnorm{ \mc N^{\theta} }^{\nicefrac{1}{2}} + \plipnorm{ \mc N^{\theta} } \bigr)
    \le \ms C \bigl( \eucnorm{ \vartheta }^{\nicefrac{1}{2}} + \eucnorm{ \vartheta }^2 \bigr).
  \end{equation}
\end{theorem}
\end{samepage}

\cref{thm:positive} is an immediate consequence of \cref{cor:bound:combined} in Subsection~\ref{subsec:equivalence} below combined with the fact that for all $ x, y \in [0,\infty) $ it holds that $ \max \{ x, y \} \le x + y \le 2 \max\{ x, y \} $. \cref{cor:bound:combined} follows from \cref{cor:bound:explicit} in Subsection~\ref{subsec:upper_bounds_lipschitz_norms} below, which, in turn, builds on \cref{thm:bound} in Subsection~\ref{subsec:upper_bounds_lipschitz_norms}. In the following, we add some explanatory comments regarding the mathematical objects that appear in \cref{thm:positive} above.

The natural number $ \inp \in \N = \{ 1, 2, 3, \ldots \} $ in \cref{thm:positive} specifies the number of neurons on the input layer, whereas the natural number $ \width \in \N $ specifies the number of neurons on the hidden layer. There are $ \inp \width $ real weight parameters and $ \width $ real bias parameters for the first affine linear transformation from the $ \inp $-dimension input layer to the $ \width $-dimensional hidden layer, and there are $ \width $ real weight parameters and $ 1 $ real bias parameter for the second affine linear transformation from the $ \width $-dimensional hidden layer to the one-dimensional output layer (cf.~also Figure~\ref{figure:architecture} above for a graphical illustration of the considered shallow \ANN\ architecture). The total number of parameters, specified by the natural number $ \dimension \in \N $ in \cref{thm:positive}, thus satisfies
\begin{equation}
  \dimension = ( \inp \width + \width ) + ( \width + 1 ) = \inp \width + 2 \width + 1.
\end{equation}
The range of the permissible input data of the \ANNs\ considered in \cref{thm:positive} is described by the real parameters $ \ms a \in \R $ and $ \ms b \in (\ms a, \infty) $. Note that for every \ANN\ parameter vector $ \theta \in \R^{\dimension} $ we have that the function
\begin{equation}
  [\ms a, \ms b]^{\inp} \ni x \mapsto \mc N^{\theta} (x) \in \R
\end{equation}
in \cref{thm:positive} constitutes the realization function associated with the \ANN\ parameter vector \nolinebreak $ \theta $.
Moreover, in \cref{thm:positive}, for a fixed point $ z \in [\ms a, \ms b]^{\inp} $ we have that for every function $ f \colon [\ms a, \ms b]^{\inp} \to \R $ the extended real number
\begin{equation}\label{eq:lipschitz_norm}
  \plipnorm{ f } = \abs{ f ( z ) } + \sup\nolimits_{ x, y \in [\ms a, \ms b]^{\inp}, \, x \neq y }
    \tfrac{ \abs{ f(x) - f(y) } }{ \eucnorm{ x - y } } \in [0,\infty]
\end{equation}
specifies the Lipschitz norm of $ f $. We note that there are several definitions of the Lipschitz norm in the scientific literature; however, all of these Lipschitz norms are equivalent.

Under these conditions, \cref{thm:positive} establishes that there exist real numbers $ \ms c, \ms C \in \R $ such that for every \ANN\ parameter vector $ \theta \in \R^{\dimension} $ there exists an \ANN\ parameter vector $ \vartheta \in \R^{\dimension} $ such that $ \mc N^{\vartheta} = \mc N^{\theta} $ and
\begin{equation}\label{eq:explanatory:comments}
    \eucnorm{ \vartheta }
    \le \ms c \bigl( \plipnorm{ \mc N^{\theta} }^{\nicefrac{1}{2}} + \plipnorm{ \mc N^{\theta} } \bigr)
    \le \ms C \bigl( \eucnorm{ \vartheta }^{\nicefrac{1}{2}} + \eucnorm{ \vartheta }^2 \bigr).
\end{equation}
Thus, for every \ANN\ parameter vector there exists a reparameterization, by which we mean an \ANN\ parameter vector with the same realization function, such that the standard norm of the parameters is bounded, up to a multiplicative constant, by the sum of powers of the Lipschitz norm of the realization function with the exponents $ \nicefrac{1}{2} $ and $ 1 $. Note, however, that due to the fact that all norms on $ \R^{\inp} $ are equivalent (can up to a multiplicative constant be estimated against each other), we have that the statement of \cref{thm:positive} with the standard norm replaced by another norm is also true.

Furthermore, observe that the right inequality in \cref{eq:explanatory:comments} is elementary and follows from the well-known fact that there exists a real number $ \mf c \in \R $ such that for all \ANN\ parameter vectors $ \theta \in \R^{\dimension} $ it holds that
\begin{equation}
  \plipnorm{ \mc N^{\theta} } \le \mf c \bigl( \eucnorm{ \theta } + \eucnorm{ \theta }^2 \bigr)
\end{equation}
(see above \cref{thm:positive}).
The left inequality, on the other hand, or a reparameterization bound comparable in kind, is to the best of our knowledge not known in the scientific literature and is one of the key contributions of this article. We emphasize that a reparameterization of the \ANNs\ is mandatory for the left inequality to hold, since parts of the parameters of every \ANN\ can be chosen arbitrarily large without changing its realization function, for example, by scaling the input weights and output weights of hidden neurons. Both inequalities combined roughly give a kind of equivalence for the class of \ANN\ parameter vectors with the same realization function and the Lipschitz norm of the \ANN\ realization function.

The upper bounds for the reparameterized network parameters from \cref{thm:positive} and its more general version in \cref{thm:bound}, respectively, are also relevant in other aspects. For example, in the training of \ANNs\ with one hidden layer and \ReLU\ activation in a supervised learning problem the position of global minima of the underlying risk function can be specified in more detail. Specifically, in \cref{cor:global_minimum} below we show in the special situation where there is only one neuron on the input layer (corresponding to the case $ d = 1 $ in \cref{thm:positive}) and where the target function $ f \colon [\ms a, \ms b] \to \R $ is Lipschitz continuous that there exists a global minimum of the risk function within an area that depends on the permissible input domain specified by $ \ms a, \ms b \in \R $, the network width $ \width \in \N $, the Lipschitz constant of the target function, and the supremum norm of the target function. We now present the precise statement of \cref{cor:global_minimum}.

\begin{cor}\label{cor:global_minimum}
  Let $ \width, \dimension \in \N $, $ \ms a, L, \ms C \in \R $, $ \ms b \in (\ms a, \infty) $ satisfy $ \dimension = 3 \width + 1 $,
  let $ f \colon [\ms a, \ms b] \to \R $ satisfy for all $ x, y \in [\ms a, \ms b] $ that $ \abs{ f(x) - f(y) } \le L \abs{ x - y } $ and
  \begin{equation}
    \ms C \ge \max \bigl\{ \max\{ 2, \abs{ \ms a }, \abs{ \ms b } \} \width^{\nicefrac{1}{2}} L^{\nicefrac{1}{2}},
       ( \ms b - \ms a ) ( 2 \width^2 + \width ) L + \sup\nolimits_{ z \in [\ms a, \ms b] } \abs{ f(z) } \bigr\},
  \end{equation}
  let $ \mu \colon \mc B( [\ms a, \ms b] ) \to [0,\infty] $ be a measure, and let $ \mc L \colon \R^{\dimension} \to \R $ satisfy for all
  $ \theta = ( \theta_1, \ldots, \theta_{\dimension} ) \in \R^{\dimension} $ that
  \begin{equation}\label{eq:cor:global_minimum}
    \mc L ( \theta ) = \smallint\nolimits_{\ms a}^{\ms b} ( f(x) - \theta_{\dimension} - \smallsum_{i=1}^{\width}
      \theta_{ 2 \width + i } \max\{ \theta_{ \width + i } + \theta_i x, 0 \} )^2 \, \mu( \diff x ).
  \end{equation}
  Then there exists $ \theta \in [ - \ms C , \ms C ]^{\dimension} $ such that
  $ \mc L ( \theta ) = \inf_{ \vartheta \in \R^{\dimension} } \mc L ( \vartheta ) $.
\end{cor}

\cref{cor:global_minimum} is a direct consequence of \cite[Theorem~2.2]{JML-1-2} combined with \cref{thm:bound} in Subsection~\ref{subsec:upper_bounds_lipschitz_norms} below. Note that \cref{cor:global_minimum}, for example, ensures that in the special situation in the training of an \ANN\ with $ \width = 5 $ neurons on the hidden layer and where there are $ m \in \N $ many input-output data pairs given by the input data $ x_1, x_2, \ldots, x_m \in [0,1] $ and the output data $ y_1, y_2, \ldots, y_m \in [-1,1] $, which satisfy that for all $ i, j \in \{ 1, 2, \ldots, m \} $ with $ i \neq j $ it holds that $ \abs{ y_i - y_j } \le \abs{ x_i - x_j } $, there exists a global minimum point $ \theta = ( \theta_1, \ldots, \theta_{\dimension} ) \in \R^{\dimension} = \R^{16} $ of the \MSE\ risk function
\begin{equation}
  \R^{\dimension} \ni \vartheta = ( \vartheta_1, \ldots, \vartheta_{\dimension} ) \mapsto \mc L ( \vartheta )
    = \tfrac{1}{m} \smallsum_{i=1}^m \abs{ y_i - \vartheta_{\dimension} - \smallsum_{j=1}^{\width}
      \vartheta_{ 2 \width + j } \max\{ \vartheta_{ \width + j } + \vartheta_j x_i, 0 \} }^2 \in \R
\end{equation}
which satisfies $ \max_{ i \in \{ 1, 2, \ldots, \dimension \} } \abs{ \theta_i } \le 56 $.

We also want to mention the relationship of \cref{thm:positive} to the concept of the so-called inverse stability of the realization map. This concept deals with the question under which circumstances \ANNs\ with similar realization functions can be reparameterized so that their new representatives are close together (cf.~Berner et al.~\cite[Definition~1.1]{berner2019}). In Petersen et al.~\cite[Section~4]{MR4243432} it was shown that the inverse stability of the realization map for deep \ANNs\ with non-affine linear Lipschitz continuous activation functions fails with respect to the uniform norm. Berner et al.~\cite{berner2019}, on the other hand, demonstrates that the inverse stability does hold on a restricted parameterization space for shallow \ANNs\ with \ReLU\ activation without biases with respect to the Sobolev semi-norm.

The second main result of our article, \cref{thm:equivalence} below, addresses the optimality of \cref{eq:thm:positive} in \cref{thm:positive}. In the following, we show, on the one hand, that the Lipschitz norm in \cref{eq:thm:positive} cannot be replaced by Hölder norms and, on the other hand, that the range of the exponents of the powers of the Lipschitz norm cannot be attenuated. For the precise statement, we now present \cref{thm:equivalence}.

\begin{samepage}
\begin{theorem}\label{thm:equivalence}
  Let $ \inp, \width, \dimension \in \N $, $ \ms a \in \R $, $ \ms b \in (\ms a, \infty) $ satisfy $ \dimension = \inp \width + 2 \width + 1 $,
  for every $ \theta = ( \theta_1, \ldots, \theta_{\dimension} ) \in \R^{\dimension} $ let
  $ \mc N^{\theta} \in C( [\ms a, \ms b]^{\inp}, \R ) $ satisfy for all $ x = ( x_1, \ldots, x_{\inp} ) \in [\ms a, \ms b]^{\inp} $ that
  \begin{equation}
    \mc N^{ \theta } (x) = \theta_{\dimension} + \smallsum\nolimits_{i=1}^{\width} \theta_{ \inp \width + \width + i }
    \max \bigl\{ \theta_{ \inp \width + i } + \smallsum\nolimits_{j=1}^{\inp} \theta_{ (i-1) \inp + j } x_j, 0 \bigr\},
  \end{equation}
  for every $ n \in \N $, $ x = ( x_1, \ldots, x_n ) \in \R^n $ let $ \eucnorm{ x } = ( \smallsum_{j=1}^n \abs{ x_j }^2 )^{1/2} \in \R $,
  for every $ f \colon [\ms a, \ms b]^{\inp} \to \R $ and every $ \gamma \in [0,1] $ let
  $ \plipnorm{ f }_{\gamma} = \sup\nolimits_{ x \in [\ms a, \ms b]^{\inp} } \abs{ f ( x ) }
    + \sup\nolimits_{ x, y \in [\ms a, \ms b]^{\inp}, \, x \neq y } \nicefrac{ \abs{ f(x) - f(y) } }{ \eucnorm{ x - y }^{\gamma} }
    \in [0,\infty] $,
  and let $ n \in \N $, $ \gamma_1, \gamma_2, \ldots, \gamma_n \in [0,1] $, $ \delta_1, \delta_2, \ldots, \delta_n \in [0,\infty) $ satisfy
  $ \max\{ \gamma_1, \gamma_2, \ldots, \allowbreak \gamma_n \} \ind_{ [0,1) }
    ( \min\{ \gamma_1, \allowbreak \gamma_2, \allowbreak \ldots, \gamma_n \} ) < 1 $.
  Then the following two statements are equivalent:
  \begin{enumerate}[(i)]
    \item There exists $ \ms c \in \R $ such that for all $ \theta \in \R^{\dimension} $ there exists $ \vartheta \in \R^{\dimension} $
    such that $ \mc N^{\vartheta} = \mc N^{\theta} $ and
      \begin{equation}
        \eucnorm{ \vartheta } \le \ms c \bigl( \plipnorm{ \mc N^{\theta} }_{\gamma_1}^{\delta_1} + \ldots
          + \plipnorm{ \mc N^{\theta} }_{\gamma_n}^{\delta_n} \bigr).
      \end{equation}
    \item There exist $ i, j \in \{ 1, 2, \ldots, n \} $ such that $ \gamma_i = \gamma_j = 1 $, $ \delta_i \le \nicefrac{1}{2} $, and
      $ \delta_j \ge 1 $.
  \end{enumerate}
\end{theorem}
\end{samepage}

\cref{thm:equivalence} is a direct consequence of \cref{cor:lipnorm:range:positive} in Subsection~\ref{subsec:upper_bounds_lipschitz_norms} below, \cref{cor:lipnorm:range:negative} in Subsection~\ref{subsec:lower_bounds_lipschitz_norms} below, and \cref{cor:hoelder:sob:sum} in Subsection~\ref{subsec:lower_bounds_hoelder_norms_sob-slob_norms} below. We have that for every function $ f \colon [\ms a, \ms b]^{\inp} \to \R $ and every $ \gamma \in [0,1] $ the extended real number
\begin{equation}\label{eq:hoelder_norm}
  \plipnorm{ f }_{\gamma} = \sup\nolimits_{ x \in [\ms a, \ms b]^{\inp} } \abs{ f(x) }
    + \sup\nolimits_{ x, y \in [\ms a, \ms b]^{\inp}, \, x \neq y }
      \tfrac{ \abs{ f(x) - f(y) } }{ \eucnorm{ x - y }^{\gamma} } \in [0,\infty]
\end{equation}
in \cref{thm:equivalence} above specifies the Hölder norm of $ f $. Observe that, in the case that $ \gamma = 1 $, the Hölder norm in \cref{eq:hoelder_norm} is equivalent to the Lipschitz norm in \cref{eq:lipschitz_norm} and can therefore be considered as the Lipschitz norm.
In the following, we want to explain \cref{thm:equivalence} in more detail.

Note that \cref{thm:equivalence}, in the case that $ \min\{ \gamma_1, \gamma_2, \ldots, \gamma_n \} = 1 $, shows that the range of the exponents of the powers of the Lipschitz norm of the \ANN\ realization function must extend at least from  $ \nicefrac{1}{2} $ to $ 1 $ for the upper bound of the reparameterized network parameters to hold. In particular, this implies that the upper bound for the reparameterized network parameters in \cref{eq:thm:positive} only holds for sums of powers of the Lipschitz norm with the exponents $ \nicefrac{1}{2} $ and $ 1 $ but does not hold for the Lipschitz norm alone. Moreover, \cref{thm:equivalence} above, in the case that $ \min\{ \gamma_1, \gamma_2, \ldots, \gamma_n \} < 1 $, demonstrates that it is not possible to control the network parameters of reparameterized \ANNs\ using sums of powers of the Hölder norm of the realization function with arbitrary exponents. In \cref{cor:hoelder:sob:sum} in Subsection~\ref{subsec:lower_bounds_hoelder_norms_sob-slob_norms}, we show that this does also hold for Sobolev-Slobodeckij norms. Specifically, the realization map for shallow \ANNs\ with \ReLU\ activation is not inverse stable with respect to Hölder norms and Sobolev-Slobodeckij norms.

The remainder of this article is organized in the following way. In \cref{sec:upper_bounds_lipschitz_norms}, we establish upper bounds for norms of reparameterized \ANNs\ using Lipschitz norms. In \cref{sec:lower_bounds_lipschitz_norms}, we address the optimality of the bounds from \cref{sec:upper_bounds_lipschitz_norms} and prove lower bounds for norms of reparameterized \ANNs\ using Lipschitz norms. Finally, in \cref{sec:lower_bounds_hoelder_norms_sob-slob_norms}, we consider different norms for the realization function and establish lower bounds for norms of reparameterized \ANNs\ using Hölder norms and Sobolev-Slobodeckij norms.

\newpage

\section{Upper bounds for norms of reparameterized artificial neural networks (ANNs) using Lipschitz norms}
\label{sec:upper_bounds_lipschitz_norms}

In this section, we establish in \cref{cor:bound:explicit} in Subsection~\ref{subsec:upper_bounds_lipschitz_norms} below upper bounds for norms of reparameterized \ANN\ parameter vectors using Lipschitz norms. In particular, we show that every \ANN\ parameter vector $ \theta \in \R^{\dimension} = \R^{ \inp \width + 2 \width + 1 } $ can be reparameterized by an \ANN\ $ \vartheta \in \R^{\dimension} $ such that the maximum norm of $ \vartheta $ is, up to a multiplicative constant, bounded by the maximum of powers of the Lipschitz norm of the realization function $ \mc N^{\theta} \colon [\ms a, \ms b]^{\inp} \to \R $ with the exponents $ \nicefrac{1}{2} $ and $ 1 $. The proof of \cref{cor:bound:explicit} uses our main result of this section, the upper bounds for norms of reparameterized \ANNs\ in \cref{thm:bound} in Subsection~\ref{subsec:upper_bounds_lipschitz_norms}. \cref{thm:bound}, in turn, builds on the well-known properties of tessellations of convex polytopes in compact cubes established in \cref{lem:tesselations} in Subsection~\ref{subsec:properties_tessellations} below, on the well-known properties of affine hyperplanes established in \cref{lem:hyperplane:scalar} and \cref{lem:hyperplanes:non-parallism} in Subsection~\ref{subsec:properties_affine_hyperplanes} below, and on the essentially well-known ability to isolate points of affine linear hyperplanes in compact cubes presented in \cref{lem:hyperplanes:isolation} in Subsection~\ref{subsec:properties_affine_hyperplanes}. In \cref{cor:bound:combined} in Subsection~\ref{subsec:equivalence} below, we combine \cref{cor:bound:explicit} and the well-known upper bounds of the Lipschitz constant and the Lipschitz norm of the realization function of an \ANN\ established in \cref{lem:lipschitzconstant:bound} and \cref{lem:inverse:bound} in Subsection~\ref{subsec:equivalence}, respectively, to obtain a kind of equivalence for the class of \ANN\ parameter vectors with the same realization function and the Lipschitz norm of the \ANN\ realization function.

In \cref{setting:main} in Subsection~\ref{subsec:upper_bounds_lipschitz_norms}, we describe our mathematical setup to introduce the architecture of the considered shallow \ANNs, specified by the number of input neurons $ \inp \in \N $ and the number of hidden neurons $ \width \in \N $, the dimension of the parameter space $ \dimension = \inp \width + 2 \width + 1 \in \N $, and the realization function $ \mc N^{\theta} \colon [\ms a, \ms b]^{\inp} \to \R $ associated with every \ANN\ parameter vector $ \theta \in \R^{\dimension} $. For the convenience of the reader, we recall the notions of the standard scalar product and of the standard norm in \cref{def:scalar_product_norm} in  Subsection~\ref{subsec:properties_tessellations}, and for every $ A \subseteq [\ms a, \ms b]^{\inp} $ with $ A \neq \emptyset $ and every $ f \colon [\ms a, \ms b]^{\inp} \to \R $ we introduce in \cref{def:lipschitz_norm} in Subsection~\ref{subsec:upper_bounds_lipschitz_norms} the extended real number $ \lipnorm{f}{A} \in [0,\infty] $, which corresponds to the Lipschitz norm of $ f $ in the case that $ A = \{ z \} $ for a fixed point $ z \in [\ms a, \ms b]^{\inp} $.

\subsection{Properties of tessellations of convex polytopes in compact cubes}
\label{subsec:properties_tessellations}

\cfclear
\begin{definition}\label{def:scalar_product_norm}
  For every $ d \in \N $, $ x = ( x_1, \ldots, x_d ) $, $ y = ( y_1, \ldots, y_d ) \in \R^d $ we denote by
  $ \scalprod{ x }{ y } \in \R $ and $ \eucnorm{ x } \in \R $ the real numbers which satisfy that
  $ \scalprod{ x }{ y } = \sum_{i=1}^d x_i y_i $ and $ \eucnorm{ x } = ( \sum_{i=1}^d \abs{ x_i }^2 )^{\nicefrac{1}{2}} $.
\end{definition}

\cfclear
\begin{definition}\label{def:half_spaces_hyperplane}
  \cfconsiderloaded{def:half_spaces_hyperplane}
  For every $ d \in \N $, $ w = ( w_1, \ldots, w_d ) \in \R^d $, $ b \in \R $, $ \ell \in \{ 0, 1 \} $
  we denote by $ \half_{w,b}^{\ell} \subseteq \R^d $ and $ \hyper_{w,b} \subseteq \R^d $
  the sets given by
  \begin{equation}
    \half_{w,b}^{\ell} = \bigl\{ x \in \R^d \colon (-1)^{\ell} ( b + \scalprod{ w }{ x } ) \le 0 \bigr\}
    \qqandqq
    \hyper_{w,b} = \bigl\{ x \in \R^d \colon b + \scalprod{ w }{ x } = 0 \bigr\}
  \end{equation}
  \cfload.
\end{definition}

\cfclear
\begin{lemma}\label{lem:tesselations}
  Let $ d, N \in \N $, $ \ms a \in \R $, $ \ms b \in (\ms a, \infty) $, $ w_1, w_2, \ldots, w_N \in \R^d $, $ b_1, b_2, \ldots, b_N \in \R $.
  Then for all $ x \in [\ms a, \ms b]^d $ there exist $ y \in (\ms a, \ms b)^d $, $ \ell_1, \ell_2, \ldots, \ell_N \in \{ 0, 1 \} $,
  $ \varepsilon \in (0,\infty) $ such that
  \begin{equation}
    x \in \Bigl( \smallbigcap_{i=1}^N \half_{w_i,b_i}^{\ell_i} \Bigr)
    \qqandqq
    \{ u \in \R^d \colon \eucnorm{ y - u } \le \varepsilon \}
      \subseteq \Bigl( \smallbigcap_{i=1}^N \half_{w_i,b_i}^{\ell_i} \Bigr)
  \end{equation}
  \cfout.
\end{lemma}

\begin{cproof}{lem:tesselations}
  Throughout this proof let $ \mu \colon \mc B( \R^d ) \to [0,\infty] $ be the Lebesgue measure and
  let $ P_{\ell} \subseteq \R^d $, $ \ell \in \{ 0, 1 \}^N $, and $ A_x \subseteq \{ 0, 1 \}^N $, $ x \in \R^d $,
  satisfy for all $ \ell = ( \ell_1, \ldots, \ell_N ) \in \{ 0, 1 \}^N $, $ x \in \R^d $ that
  \begin{equation}
    P_{\ell} = \bigl( \smallbigcap_{i=1}^N \half_{w_i,b_i}^{\ell_i} \bigr)
    \qqandqq
    A_x = \bigl\{ \ell \in \{ 0, 1 \}^N \colon x \in P_{\ell} \bigr\}
  \end{equation}
  \cfload.
  \Nobs that the fact that for all $ i \in \{ 1, 2, \ldots, N \} $, $ \ell \in \{ 0, 1 \} $ it holds that
  $ \half_{w_i,b_i}^{\ell_i} \subseteq \R^d $ is closed \proves that for all $ \ell \in \{ 0, 1 \}^N $ it holds that
  $ P_{\ell} \subseteq \R^d $ is closed.
  \Hence that for all $ x \in \R^d $, $ \ell \in \{ 0, 1 \}^N \backslash A_x $ there exists $ \varepsilon \in (0,\infty) $ such that
  \begin{equation}
    \{ u \in \R^d \colon \eucnorm{ x - u } \le \varepsilon \} \cap P_{\ell} = \emptyset
  \end{equation}
  \cfload.
  This \proves that for all $ x \in \R^d $ there exists $ \varepsilon \in (0,\infty) $ such that it holds that
  \begin{equation}
    \{ u \in \R^d \colon \eucnorm{ x - u } \le \varepsilon \} \cap
    \bigl( \smallbigcup_{ \ell \in \{ 0, 1 \}^N \backslash A_x } P_{\ell} \bigr) = \emptyset.
  \end{equation}
  The fact that $ \smallbigcup_{ \ell \in \{ 0, 1 \}^N } P_{\ell} = \R^d $ \hence
  \proves that for all $ x \in \R^d $ there exists $ \varepsilon \in (0,\infty) $ such that it holds that
  \begin{equation}
    \{ u \in \R^d \colon \eucnorm{ x - u } \le \varepsilon \} \subseteq \bigl( \smallbigcup_{ \ell \in A_x } P_{\ell} \bigr).
  \end{equation}
  \Hence that for all $ x \in [\ms a, \ms b]^d $ there exists $ \varepsilon \in (0,\infty) $ such that
  \begin{equation}
    \mu \bigl( ( \smallbigcup_{ \ell \in A_x } P_{\ell} ) \cap [\ms a, \ms b]^d \bigr)
    \ge \mu \bigl( \{ u \in \R^d \colon \eucnorm{ x - u } \le \varepsilon \} \cap [\ms a, \ms b]^d \bigr) > 0.
  \end{equation}
  This \proves that for all $ x \in [\ms a, \ms b]^d $ there exists $ \ell \in A_x $ such that
  $ \mu( P_{\ell} \cap [\ms a, \ms b]^d ) > 0 $.
  \Hence that for all $ x \in [\ms a, \ms b]^d $ there exist $ y \in (\ms a, \ms b)^d $, $ \ell_1, \ell_2, \ldots, \ell_N \in \{ 0, 1 \} $,
  $ \varepsilon \in (0,\infty) $ such that
  \begin{equation}
    x \in \Bigl( \smallbigcap_{i=1}^N \half_{w_i,b_i}^{\ell_i} \Bigr)
    \qqandqq
    \{ u \in \R^d \colon \eucnorm{ x - u } \le \varepsilon \}
      \subseteq \Bigl( \smallbigcap_{i=1}^N \half_{w_i,b_i}^{\ell_i} \Bigr).
  \end{equation}
\end{cproof}

\subsection{Properties of affine hyperplanes in compact cubes}
\label{subsec:properties_affine_hyperplanes}

\cfclear
\begin{lemma}\label{lem:hyperplane:scalar}
  Let $ d \in \N $, $ z \in \R^d $, $ w_1, w_2 \in \R^d \backslash \{ 0 \} $, $ b_1, b_2 \in \R $
  satisfy $ \hyper_{w_1,b_1} = \hyper_{w_2,b_2} $ and $ z \notin ( \half_{w_1,b_1}^1 \cup \half_{w_2,b_2}^1 ) $
  \cfload.
  Then it holds that $ \eucnorm{ w_1 } w_2 = \eucnorm{ w_2 } w_1 $ and $ \eucnorm{ w_1 } b_2 = \eucnorm{ w_2 } b_1 $
  \cfout.
\end{lemma}

\begin{cproof}{lem:hyperplane:scalar}
  Throughout this proof let $ A = ( A_1, A_2 ) \in \R^{ 2 \times d } $ satisfy that
  \begin{equation}
    A_1 = w_1
    \qqandqq
    A_2 = w_2.
  \end{equation}
  \Nobs that the fact that $ w_1 \neq 0 $ and the assumption that $ \hyper_{w_1,b_1} = \hyper_{w_2,b_2} $
  \prove that there exists $ u \in \R^d $ which satisfies for all $ i \in \{ 1, 2 \}$ that
  \begin{equation}\label{eq:lem:hyperplane:scalar:u}
    b_i + \scalprod{ w_i }{ u } = 0
  \end{equation}
  \cfload.
  \Nobs that \cref{eq:lem:hyperplane:scalar:u} \proves that for all $ i \in \{ 1, 2 \} $, $ x \in \hyper_{w_i,0} $ it holds that
  $ b_i + \scalprod{ w_i }{ u + x } = b_i + \scalprod{ w_i }{ u } + \scalprod{ w_i }{ x } = 0 $.
  Combining this with \cref{eq:lem:hyperplane:scalar:u} and the assumption that $ \hyper_{w_1,b_1} = \hyper_{w_2,b_2} $
  \proves that for all $ i, j \in \{ 1, 2 \} $, $ x \in \hyper_{w_i,0} $ it holds that
  $ \scalprod{ w_j }{ x } = b_j + \scalprod{ w_j }{ u } + \scalprod{ w_j }{ x } = b_j + \scalprod{ w_j }{ u + x } = 0 $.
  \Hence that
  \begin{equation}\label{eq:lem:hyperplane:scalar:identity}
    \hyper_{w_1,0} = \hyper_{w_2,0}.
  \end{equation}
  This \proves that
  $ \ker( A ) = \{ x \in \R^d \colon \scalprod{ w_1 }{ x } = 0 \} \cap \{ x \in \R^d \colon \scalprod{ w_2 }{ x } = 0 \}
    = \{ x \in \R^d \colon \scalprod{ w_1 }{ x } = 0 \} $.
  The rank-nullity theorem \hence \proves that
  \begin{equation}
  \begin{split}
    \rank( A )
    &= d - \dim_{\R} ( \ker(A) )
    = d - \dim_{\R} ( \{ x \in \R^d \colon \scalprod{ w_1 }{ x } = 0 \} ) \\
    &= d - \bigl( d - \dim_{\R} ( \{ y \in \R \colon [ \Exists x \in \R^d \colon \scalprod{ w_1 }{ x } = y ] \} ) \bigr)
    = d - ( d - 1 )
    = 1.
  \end{split}
  \end{equation}
  \Hence that there exists $ \lambda \in \R \backslash \{ 0 \} $ which satisfies that
  \begin{equation}\label{eq:lem:hyperplane:scalar:def:lambda}
    w_1 = \lambda w_2.
  \end{equation}
  \Nobs that \cref{eq:lem:hyperplane:scalar:u}, \cref{eq:lem:hyperplane:scalar:def:lambda}, and
  the fact that $ z \notin \half_{w_1,b_1}^1 $ \prove that
  \begin{equation}
  \begin{split}
    0 &> b_1 + \scalprod{ w_1 }{ z }
    = \bigl[ b_1 + \scalprod{ w_1 }{ z } \bigr] - \bigl[ b_1 + \scalprod{ w_1 }{ u } \bigr]
    = \scalprod{ w_1 }{ z - u }
    = \lambda \scalprod{ w_2 }{ z - u } \\
    &= \lambda \bigl( \bigl[ b_2 + \scalprod{ w_2 }{ z } \bigr] - \bigl[ b_2 + \scalprod{ w_2 }{ u } \bigr] \bigr)
    = \lambda \bigl( b_2 + \scalprod{ w_2 }{ z } \bigr).
  \end{split}
  \end{equation}
  The fact that $ z \notin \half_{w_2,b_2}^1 $ \hence \proves that $ \lambda > 0 $.
  Combining this with \cref{eq:lem:hyperplane:scalar:def:lambda} \proves that
  $ \eucnorm{ w_1 } = \eucnorm{ \lambda w_2 } =  \lambda \eucnorm{ w_2 } $.
  This and the fact that $ \min\{ \lambda, \eucnorm{ w_1 } \} > 0 $ \prove that
  \begin{equation}\label{eq:lem:hyperplane:scalar:lambda:property}
    \lambda = \nicefrac{ \eucnorm{ w_1 } }{ \eucnorm{ w_2 } }.
  \end{equation}
  \Moreover \cref{eq:lem:hyperplane:scalar:u} and \cref{eq:lem:hyperplane:scalar:def:lambda} \prove that
  $ b_1 = - \scalprod{ w_1 }{ u } = - \lambda \scalprod{ w_2 }{ u } = \lambda b_2 $.
  Combining this with \cref{eq:lem:hyperplane:scalar:def:lambda} and \cref{eq:lem:hyperplane:scalar:lambda:property} \proves that
  $ \eucnorm{ w_1 } w_2 = \eucnorm{ w_2 } w_1 $ and $ \eucnorm{ w_1 } b_2 = \eucnorm{ w_2 } b_1 $.
\end{cproof}

\cfclear
\begin{lemma}\label{lem:hyperplanes:non-parallism}
  Let $ d \in \N $, $ w_1, w_2 \in \R^d $, $ b_1, b_2 \in \R $ satisfy
  $ \hyper_{w_1,b_1} \neq \hyper_{w_2,b_2} $ and $ \hyper_{w_1,b_1} \cap \hyper_{w_2,b_2} \neq \emptyset $ \cfload.
  Then for all $ \lambda \in \R \backslash \{ 0 \} $ it holds that
  \begin{equation}\label{eq:lem:hyperplanes:non-parallism}
    w_1 \neq \lambda w_2.
  \end{equation}
\end{lemma}

\begin{cproof}{lem:hyperplanes:non-parallism}
  We \prove \cref{eq:lem:hyperplanes:non-parallism} by contradiction. In the following, we thus assume that there exists
  $ \lambda \in \R \backslash \{ 0 \} $ which satisfies that
  \begin{equation}\label{eq:lem:hyperplanes:non-parallism:lambda}
    w_1 = \lambda w_2.
  \end{equation}
  \Nobs that the assumption that $ \hyper_{w_1,b_1} \cap \hyper_{w_2,b_2} \neq \emptyset $
  \proves that there exists $ z \in \R^d $ which satisfies for all $ i \in \{ 1, 2 \} $ that
  \begin{equation}\label{eq:lem:hyperplanes:non-parallism:z}
    b_i + \scalprod{ w_i }{ z } = 0
  \end{equation}
  \cfload.
  \Nobs that \cref{eq:lem:hyperplanes:non-parallism:lambda} and \cref{eq:lem:hyperplanes:non-parallism:z}
  \prove that for all $ x \in \R^d $ it holds that
  \begin{equation}
  \begin{split}
    b_1 + \scalprod{ w_1 }{ x }
    &= \bigl[ b_1 + \scalprod{ w_1 }{ x } \bigr] - \bigl[ b_1 + \scalprod{ w_1 }{ z } \bigr]
    = \scalprod{ w_1 }{ x - z }
    = \lambda \scalprod{ w_2 }{ x - z } \\
    &= \lambda \bigl( \bigl[ b_2 + \scalprod{ w_2 }{ x } \bigr] - \bigl[ b_2 + \scalprod{ w_2 }{ z } \bigr] \bigr)
    = \lambda \bigl( b_2 + \scalprod{ w_2 }{ x } \bigr).
  \end{split}
  \end{equation}
  The fact that $ \lambda \neq 0 $ \hence \proves that $ \hyper_{w_1,b_1} = \hyper_{w_2,b_2} $.
  This contradiction \proves \cref{eq:lem:hyperplanes:non-parallism}.
\end{cproof}

\cfclear
\begin{lemma}\label{lem:hyperplanes:isolation}
  Let $ d, N \in \N $, $ \ms a \in \R $, $ \ms b \in (\ms a, \infty) $, $ w_1, w_2, \ldots, w_N \in \R^d $, $ b_1, b_2, \ldots, b_N \in \R $,
  assume for all $ i \in \{ 1, 2, \ldots, N \} $ that 
  $ [\ms a, \ms b]^d \not\subseteq \half_{w_i,b_i}^1 $ and $ \half_{w_i,b_i}^1 \cap (\ms a, \ms b)^d \neq \emptyset $,
  and assume for all $ i, j \in \{ 1, 2, \ldots, N \} $ with $ i \neq j $ that
  $ \hyper_{w_i,b_i} \neq \hyper_{w_j,b_j} $ \cfload.
  Then there exist $ p_1, p_2, \ldots, p_N \in (\ms a, \ms b)^d $, $ \varepsilon \in (0,\infty) $ which satisfy for all
  $ i \in \{ 1, 2, \ldots, N \} $ that
  \begin{equation}\label{eq:lem:hyperplanes:isolation}
    p_i \in \hyper_{w_i,b_i}
    \qqandqq
    \{ x \in \R^d \colon \eucnorm{ x - p_i } \le \varepsilon \} \cap
    \Bigl( \smallbigcup_{ j \in \{ 1, 2, \ldots, N \} \backslash \{ i \} } \hyper_{w_j,b_j} \Bigr) = \emptyset
  \end{equation}
  \cfout.
\end{lemma}

\begin{cproof}{lem:hyperplanes:isolation}
  Throughout this proof let $ \varphi_i^{x,y} \colon [0,1] \to \R $, $ i \in \{ 1, 2, \ldots, N \} $, $ x, y \in \R^d $, satisfy for all
  $ i \in \{ 1, 2, \ldots, N \} $, $ x, y \in \R^d $, $ t \in [0,1] $ that $ \varphi_i^{x,y} (t) = b_i + \scalprod{ w_i }{ (1-t)x + ty } $
  and let $ A^{i,j} = ( A_1^{i,j}, A_2^{i,j} ) \in \R^{ 2 \times d } $ satisfy for all $ i, j \in \{ 1, 2, \ldots, N \} $ that
  \begin{equation}
    A_1^{i,j} = w_i
    \qqandqq
    A_2^{i,j} = w_j
  \end{equation}
  \cfload.
  \Nobs that the assumption that for all $ i \in \{ 1, 2, \ldots, N \} $ it holds that
  $ [\ms a, \ms b]^d \not\subseteq \half_{w_i,b_i}^1 $ and $ \half_{w_i,b_i}^1 \cap (\ms a, \ms b)^d \neq \emptyset $
  \proves that there exist $ u_1, u_2, \ldots, u_N \in [\ms a, \ms b]^d $,
  $ v_1, v_2, \ldots, v_N \in (\ms a, \ms b)^d $ which satisfy for all $ i \in \{ 1, 2, \ldots, N \} $ that
  \begin{equation}\label{eq:lem:hyperplanes:isolation:def:uv}
    b_i + \scalprod{ w_i }{ u_i } < 0
    \qqandqq
    b_i + \scalprod{ w_i }{ v_i } \ge 0.
  \end{equation}
  \Nobs that \cref{eq:lem:hyperplanes:isolation:def:uv} \proves that for all $ i \in \{ 1, 2, \ldots, N \} $ it holds that
  \begin{equation}
    \varphi_i^{u_i,v_i} (0) = b_i + \scalprod{ w_i }{ u_i } < 0
    \qqandqq
    \varphi_i^{u_i,v_i} (1) = b_i + \scalprod{ w_i }{ v_i } \ge 0.
  \end{equation}
  This and the fact that for all $ i \in \{ 1, 2, \ldots, N \} $ it holds that $ \varphi_i^{u_i,v_i} \in C( [0,1], \R) $ \prove that for all
  $ i \in \{ 1, 2, \ldots, N \} $ there exists $ t \in (0,1] $ such that $ \varphi_i^{u_i,v_i} (t) = 0 $.
  \Hence that there exist $ q_1, q_2, \ldots, q_N \in (\ms a, \ms b)^d $, $ \delta \in (0,\infty) $ which satisfy for all
  $ i \in \{ 1, 2, \ldots, N \} $ that
  \begin{equation}\label{eq:lem:hyperplanes:isolation:def:q}
    b_i + \scalprod{ w_i }{ q_i } = 0
    \qqandqq
    \bigl\{ x \in \R^d \colon \eucnorm{ x - q_i } \le \delta \} \subseteq (\ms a, \ms b)^d.
  \end{equation}
  Let $ M_i \subseteq \{ 1, 2, \ldots, N \} $, $ i \in \{ 1, 2, \ldots, N \} $, satisfy for all $ i \in \{ 1, 2, \ldots, N \} $ that
  \begin{equation}
    M_i = \bigl\{ j \in \{ 1, 2, \ldots, N \} \colon b_j + \scalprod{ w_j }{ q_i } = 0 \bigr\}.
  \end{equation}
  \Nobs that the fact that for all $ i \in \{ 1, 2, \ldots, N \} $ it holds that $ \R^d \ni x \mapsto b_i + \scalprod{ w_i }{ x } \in \R $
  is continuous \proves that for all $ i \in \{ 1, 2, \ldots, N \} $, $ j \in \{ 1, 2, \ldots, N \} \backslash M_i $ there exists
  $ \eta \in (0,\infty) $ such that for all $ x \in \{ y \in \R^d \colon \eucnorm{ x - q_i } \le \eta \} $ it holds that
  $ \abs{ b_j + \scalprod{ w_j }{ x } } > 0 $.
  \Hence that there exists $ \eta \in (0,\delta] $ which satisfies for all $ i \in \{ 1, 2, \ldots, N \} $,
  $ j \in \{ 1, 2, \ldots, N \} \backslash M_i $, $ x \in \{ y \in \R^d \colon \eucnorm{ x - q_i } \le \eta \} $ that
  \begin{equation}\label{eq:lem:hyperplanes:isolation:complement}
      \abs{ b_j + \scalprod{ w_j }{ x } } > 0.
  \end{equation}
  In the following, we distinguish between the case $ d = 1 $ and the case $ d > 1 $.
  We first \prove[p] \cref{eq:lem:hyperplanes:isolation} in the case
  \begin{equation}\label{eq:lem:hyperplanes:isolation:case1}
    d = 1.
  \end{equation}
  \Nobs that \cref{eq:lem:hyperplanes:isolation:case1}, the fact that for all $ i \in \{ 1, 2, \ldots, N \} $ it holds that $ w_i \neq 0 $, and
  the assumption that for all $ i, j \in \{ 1, 2, \ldots, N \} $ with $ i \neq j $ it holds that
  $ \hyper_{w_i,b_i} \neq \hyper_{w_j,b_j} $ \prove that for all $ i \in \{ 1, 2, \ldots, N \} $ it holds that $ M_i = \{ i \} $.
  Combining this with \cref{eq:lem:hyperplanes:isolation:def:q} and \cref{eq:lem:hyperplanes:isolation:complement} \proves that for all
  $ i \in \{ 1, 2, \ldots, N \} $ it holds that
  \begin{equation}
    q_i \in \hyper_{w_i,b_i}
    \qqandqq
    \bigl\{ x \in \R^d \colon \eucnorm{ x - q_i } \le \eta \bigr\} \cap
      \Bigl( \smallbigcup_{ j \in \{ 1, 2, \ldots, N \} \backslash \{ i \} } \hyper_{w_j,b_j} \Bigr) = \emptyset.
  \end{equation}
  This \proves \cref{eq:lem:hyperplanes:isolation} in the case $ d = 1 $.
  In the next step we \prove[p] \cref{eq:lem:hyperplanes:isolation} in the case
  \begin{equation}\label{eq:lem:hyperplanes:isolation:case2}
    d > 1.
  \end{equation}
  Let $ \mu \colon \mc B( \R^{d-1} ) \to [0,\infty] $ be the Lebesgue measure.
  \Nobs that \cref{lem:hyperplanes:non-parallism} (applied for every $ i \in \{ 1, 2, \ldots, N \} $, $ j \in M_i \backslash \{ i \} $ with
  $ d \is \inp $, $ w_1 \is w_i $, $ w_2 \is w_j $, $ b_1 \is b_i $, $ b_2 \is b_j $
  in the notation of \cref{lem:hyperplanes:non-parallism}) and
  the assumption that for all $ i, j \in \{ 1, 2, \ldots, N \} $ with $ i \neq j $ it holds that $ \hyper_{w_i,b_i} \neq \hyper_{w_j,b_j} $
  \prove that for all $ i \in \{ 1, 2, \ldots, N \} $, $ j \in M_i \backslash \{ i \} $, $ \lambda \in \R \backslash \{ 0 \} $ it holds that
  \begin{equation}\label{eq:lem:hyperplanes:isolation:non-parallel}
    w_i \neq \lambda w_j.
  \end{equation}
  \Moreover the rank-nullity theorem and the fact that for all $ i \in \{ 1, 2, \ldots, N \} $ it holds that $ w_i \neq 0 $ \prove that for all
  $ i \in \{ 1, 2, \ldots, N \} $ it holds that
  \begin{equation}
  \begin{split}
    \dim_{\R} ( \hyper_{w_i,0} )
    &= \dim_{\R} ( \{ x \in \R^d \colon \scalprod{ w_i }{ x } = 0 \} ) \\
    &= d - \dim_{\R} ( \{ y \in \R \colon [ \Exists x \in \R^d \colon \scalprod{ w_i }{ x } = y ] \} )
    = d - 1.
  \end{split}
  \end{equation}
  \Hence that there exist $ f_i \colon \hyper_{w_i,0} \to \R^{d-1} $, $ i \in \{ 1, 2, \ldots, N \} $, which satisfy for all
  $ i \in \{ 1, 2, \ldots, N \} $, $ x, y \in \hyper_{w_i,0} $, $ \lambda \in \R $ that
  \begin{equation}\label{eq:lem:hyperplanes:isolation:f}
    f_i( \lambda x ) = \lambda f_i (x), \quad
    f_i (x+y) = f_i (x) + f_i (y), \quad
    \ker( f_i ) = \{ 0 \}, \qandq
    f_i( \hyper_{w_i,0} ) = \R^{d-1}.
  \end{equation}
  \Nobs that \cref{eq:lem:hyperplanes:isolation:non-parallel}, \cref{eq:lem:hyperplanes:isolation:f}, the fact that for all
  $ i, j \in \{ 1, 2, \ldots, N \} $ it holds that
  $ \hyper_{w_i,0} \cap \hyper_{w_j,0}
    = \{ x \in \R^d \colon \scalprod{ w_i }{ x } = 0 \} \cap \{ x \in \R^d \colon \scalprod{ w_j }{ x } = 0 \}
    = \ker( A^{i,j} ) $,
  and the rank-nullity theorem \prove that for all $ i \in \{ 1, 2, \ldots, N \} $, $ j \in M_i \backslash \{ i \} $
  it holds that
  \begin{equation}
    \dim_{\R} \bigl( f_i( \hyper_{w_i,0} \cap \hyper_{w_j,0} ) \bigr)
    = \dim_{\R} ( \hyper_{w_i,0} \cap \hyper_{w_j,0} )
    = \dim_{\R} ( \ker( A^{i,j} ) )
    = d - 2.
  \end{equation}
  This \proves that for all $ i \in \{ 1, 2, \ldots, N \} $, $ j \in M_i \backslash \{ i \} $ it holds that
  $ \mu( f_i ( \hyper_{w_i,0} \cap \hyper_{w_j,0} ) ) = 0 $.
  \Hence that for all $ i \in \{ 1, 2, \ldots, N \} $ it holds that
  \begin{equation}\label{eq:lem:hyperplanes:isolation:Lebesgue:zero}
  \begin{split}
    0
    &\le \mu \bigl( f_i \bigl( \hyper_{w_i,0} \cap \bigl( \smallbigcup_{ j \in M_i \backslash \{ i \} } \hyper_{w_j,0} \bigr) \bigr) \bigr)
    = \mu \bigl( \smallbigcup_{ j \in M_i \backslash \{ i \} } f_i( \hyper_{w_i,0} \cap \hyper_{w_j,0} ) \bigr) \\
    &\le \smallsum_{ j \in M_i \backslash \{ i \} } \mu( f_i( \hyper_{w_i,0} \cap \hyper_{w_j,0} ) )
    = 0.
  \end{split}
  \end{equation}
  \Moreover \cref{eq:lem:hyperplanes:isolation:f} \proves that for all $ i \in \{ 1, 2, \ldots, N \} $ it holds that
  $ \mu( f_i ( \hyper_{w_i,0} ) ) = \mu( \R^{d-1} ) = \infty $.
  Combining this with \cref{eq:lem:hyperplanes:isolation:Lebesgue:zero} \proves that
  $ \hyper_{w_i,0} \not \subseteq \smallbigcup_{ j \in M_i \backslash \{ i \} } \hyper_{w_j,0} $.
  \Hence that there exist $ m_1, m_2, \ldots, m_N \in \R^d \backslash \{ 0 \} $ which satisfy for all
  $ i \in \{ 1, 2, \ldots, N \} $, $ j \in M_i \backslash \{ i \} $ that
  \begin{equation}\label{eq:lem:hyperplanes:isolation:m}
    \scalprod{ w_i }{ m_i } = 0, \qquad
    \abs{ \scalprod{ w_j }{ m_i } } > 0, \qqandqq
    \eucnorm{ m_i } \le \nicefrac{ \eta }{ 2 }.
  \end{equation}
  \Nobs that \cref{eq:lem:hyperplanes:isolation:m} and the fact that for all $ i \in \{ 1, 2, \ldots, N \} $ it holds that
  $ \R^d \ni x \mapsto \scalprod{ w_i }{ x } \in \R $ is continuous \prove that for all
  $ i \in \{ 1, 2, \ldots, N \} $, $ j \in M_i \backslash \{ i \} $ there exists $ \varepsilon \in (0,\infty) $ such that for all
  $ x \in \{ y \in \R^d \colon \eucnorm{ y - m_i } \le \varepsilon \} $ it holds that $ \abs{ \scalprod{ w_j }{ x } } > 0 $.
  \Hence that there exists $ \varepsilon \in (0,\nicefrac{\eta}{2}] $ which satisfies for all $ i \in \{ 1, 2, \ldots, N \} $,
  $ j \in M_i \backslash \{ i \} $, $ x \in \{ y \in \R^d \colon \eucnorm{ y - m_i } \le \varepsilon \} $ that
  \begin{equation}\label{eq:lem:hyperplanes:isolation:varepsilon}
    \abs{ \scalprod{ w_j }{ x } } > 0.
  \end{equation}
  \Nobs that \cref{eq:lem:hyperplanes:isolation:def:q} and \cref{eq:lem:hyperplanes:isolation:varepsilon} \prove that for all
  $ i \in \{ 1, 2, \ldots, N \} $, $ j \in M_i \backslash \{ i \} $,
  $ x \in \{ y \in \R^d \colon \eucnorm{ x - ( q_i + m_i ) } \le \varepsilon \} $ it holds that
  \begin{equation}\label{eq:lem:hyperplanes:isolation:M:property}
    \abs{ b_j + \scalprod{ w_j }{ x } }
    = \babs{ \bigl[ b_j + \scalprod{ w_j }{ x } \bigr] - \bigl[ b_j + \scalprod{ w_j }{ q_i } \bigr] }
    = \abs{ \scalprod{ w_j }{ x - q_i } }
    > 0.
  \end{equation}
  \Moreover \cref{eq:lem:hyperplanes:isolation:m} \proves that for $ i \in \{ 1, 2, \ldots, N \} $,
  $ x \in \{ y \in \R^d \colon \eucnorm{ x - ( q_i + m_i ) } \le \varepsilon \} $ it holds that
  \begin{equation}
    \eucnorm{ x - q_i }
    = \eucnorm{ x - ( q_i + m_i ) + m_i }
    \le \eucnorm{ x - ( q_i + m_i ) } + \eucnorm{ m_i }
    \le \varepsilon + \nicefrac{ \eta }{ 2 }
    \le \eta.
  \end{equation}
  Combining this with \cref{eq:lem:hyperplanes:isolation:complement} and \cref{eq:lem:hyperplanes:isolation:M:property} \proves that for all
  $ i \in \{ 1, 2, \ldots, N \} $ it holds that
  \begin{equation}\label{eq:lem:hyperplanes:isolation:separation}
    \bigl\{ x \in \R^d \colon \eucnorm{ x - ( q_i + m_i ) } \le \varepsilon \bigr\} \cap
      \Bigl( \smallbigcup_{ j \in \{ 1, 2, \ldots, N \} \backslash \{ i \} } \hyper_{w_j,b_j} \Bigr) = \emptyset.
  \end{equation}
  \Moreover \cref{eq:lem:hyperplanes:isolation:def:q} and \cref{eq:lem:hyperplanes:isolation:m} \prove that for all
  $ i \in \{ 1, 2, \ldots, N \} $ it holds that
  \begin{equation}
    b_i + \scalprod{ w_i }{ q_i + m_i } = b_i + \scalprod{ w_i }{ q_i } + \scalprod{ w_i }{ m_i } = 0
    \qqandqq
    q_i + m_i \in (\ms a, \ms b)^d.
  \end{equation}
  This and \cref{eq:lem:hyperplanes:isolation:separation} \prove \cref{eq:lem:hyperplanes:isolation} in the case $ d > 1 $.
\end{cproof}

\subsection{Upper bounds for norms of reparameterized ANNs using Lipschitz norms}
\label{subsec:upper_bounds_lipschitz_norms}

\cfclear
\begin{setting}\label{setting:main}
  Let $ \inp, \width, \dimension \in \N $, $ \ms a \in \R $, $ \ms b \in (\ms a, \infty) $ satisfy $ \dimension = \inp \width + 2 \width + 1 $
  and for every $ \theta = ( \theta_1, \ldots, \theta_{\dimension} ) \in \R^{\dimension} $ let
  $ \mc N^{\theta} \in C( [\ms a, \ms b]^{\inp}, \R ) $ satisfy for all $ x = ( x_1, \ldots, x_{\inp} ) \in [\ms a, \ms b]^{\inp} $ that
  $ \mc N^{ \theta } (x) = \theta_{\dimension} + \sum_{i=1}^{\width} \theta_{ \inp \width + \width + i }
    \max \{ \theta_{ \inp \width + i } + \sum_{j=1}^{\inp} \theta_{ (i-1) \inp + j } x_j, 0 \} $.
\end{setting}

\cfclear
\begin{theorem}\label{thm:bound}
  Assume \cref{setting:main} and let $ \theta \in \R^{\dimension} $.
  Then there exists $ \vartheta = ( \vartheta_1, \ldots, \vartheta_{\dimension} ) \in \R^{\dimension} $ such that
  $ \mc N^{\vartheta} = \mc N^{\theta} $ and
  \begin{equation}\label{eq:thm:bound}
  \begin{split}
    \max\nolimits_{ i \in \{ 1, 2, \ldots, \dimension \} } \abs{ \vartheta_i }
    &\le \max \Bigl\{ \max \bigl\{ 2, \abs{ \ms a } \sqrt{\inp}, \abs{ \ms b } \sqrt{\inp} \bigr\}
      \Bigl[ \sup\nolimits_{ x, y \in [\ms a, \ms b]^{\inp}, \, x \neq y }
      \tfrac{ \abs{ \mc N^{\theta} (x) - \mc N^{\theta} (y) } }{ \eucnorm{ x - y } } \Bigr]^{\nicefrac{1}{2}}, \\
    &\quad \bigl[ \inf\nolimits_{ x \in [\ms a, \ms b]^{\inp} } \abs{ \mc N^{\theta} (x) } \bigr] + 2 \width ( \ms b - \ms a ) \sqrt{\inp}
      \Bigl[ \sup\nolimits_{ x, y \in [\ms a, \ms b]^{\inp}, \, x \neq y }
      \tfrac{ \abs{ \mc N^{\theta} (x) - \mc N^{\theta} (y) } }{ \eucnorm{ x - y } } \Bigr] \Bigr\}
  \end{split}
  \end{equation}
  \cfout.
\end{theorem}

\begin{cproof}{thm:bound}
  Throughout this proof let $ \theta_1, \theta_2, \ldots, \theta_{\dimension}, L \in \R $ satisfy
  $ \theta = ( \theta_1, \ldots, \theta_{\dimension} ) $ and
  \begin{equation}\label{eq:thm:bound:def:L}
    L = \sup\nolimits_{ x, y \in [\ms a, \ms b]^{\inp}, \, x \neq y }
      \tfrac{ \abs{ \mc N^{\theta} (x) - \mc N^{\theta} (y) } }{ \eucnorm{ x - y } },
  \end{equation}
  let $ w = ( w_1, \ldots, w_{\width} ) = ( w_{i,j} )_{ (i,j) \in \{ 1, 2, \ldots, \width \} \times \{ 1, 2, \ldots \inp \} }
    \in \R^{ \width \times \inp } $, $ b = ( b_1, \ldots, b_{\width} ) $, $ v = ( v_1, \ldots, v_{\width} ) \in \R^{ \width } $
  satisfy for all $ i \in \{ 1, 2, \ldots, \width \} $, $ j \in \{ 1, 2, \ldots, \inp \} $ that
  \begin{equation}\label{eq:thm:bound:def:wbv}
    w_{i,j} = \theta_{ (i-1) \inp + j }, \qquad
    b_i = \theta_{ \inp \width + i }, \qqandqq
    v_i = \theta_{ \inp \width + \width + i },
  \end{equation}
  let $ A_k \subseteq \N $, $ k \in \{ 1, 2, 3 \} $, satisfy
  \begin{equation}\label{eq:thm:bound:def:ABC}
  \begin{gathered}
    A_1 = \bigl\{ i \in \{ 1, 2, \ldots, \width \} \colon \bigl( [\ms a, \ms b]^{\inp} \subseteq \half_{w_i,b_i}^1 \bigr) \bigr\}, \\
    A_2 = \bigl\{ i \in \{ 1, 2, \ldots, \width \} \colon \bigl[ \bigl( [\ms a, \ms b]^{\inp} \not\subseteq \half_{w_i,b_i}^1 \bigr) \wedge
      \bigl( \half_{w_i,b_i}^1 \cap (\ms a, \ms b)^{\inp} \neq \emptyset \bigr) \bigr] \bigr\}, \\ \text{and} \qquad
    A_3 = \bigl\{ i \in \{ 1, 2, \ldots, \width \} \colon \bigl( \half_{w_i,b_i}^1 \cap (\ms a, \ms b)^{\inp} = \emptyset \bigr) \bigr\},
  \end{gathered}
  \end{equation}
  and let $ N \in \N $ satisfy $ N = \# ( \smallbigcup_{ i \in A_2 } \{ \hyper_{w_i,b_i} \} ) $ \cfload.
  \Nobs that the fact that $ \mc N^{\theta} \in C( [\ms a, \ms b]^{\inp}, \R ) $ \proves that there exists
  $ z = ( z_1, \ldots, z_{\inp} ) \in [\ms a, \ms b]^{\inp} $ which satisfies
  \begin{equation}\label{eq:thm:bound:def:inf}
    \abs{ \mc N^{\theta} (z) } = \inf\nolimits_{ x \in [\ms a, \ms b]^{\inp} } \abs{ \mc N^{\theta} (x) }.
  \end{equation}
  \Nobs that \cref{lem:tesselations} (applied with
  $ d \is \inp $, $ N \is \width $, $ \ms a \is \ms a $, $ \ms b \is \ms b $,
  $ ( w_i )_{ i \in \{ 1, 2, \ldots, N \} } \is ( w_i )_{ i \in \{ 1, 2, \ldots, \width \} } $,
  $ ( b_i )_{ i \in \{ 1, 2, \ldots, N \} } \is ( b_i )_{ i \in \{ 1, 2, \ldots, \width \} } $, $ x \is z $
  in the notation of \cref{lem:tesselations})
  \proves that there exist $ \ms z = ( \ms z_1, \ldots, \ms z_{\inp} ) \in (\ms a, \ms b)^{\inp} $,
  $ \mf l_1, \mf l_2, \ldots, \mf l_{\width} \in \{ 0, 1 \} $, $ \varepsilon \in (0,\infty) $ which satisfy that
  \begin{equation}\label{eq:thm:bound:def:polygon}
    z \in \Bigl( \smallbigcap_{i=1}^{\width} \half_{w_i,b_i}^{\mf l_i} \Bigr)
    \qqandqq
    \{ x \in \R^{\inp} \colon \eucnorm{ x - \ms z } \le \varepsilon \}
      \subseteq \Bigl( \smallbigcap_{i=1}^{\width} \half_{w_i,b_i}^{\mf l_i} \Bigr) \cap [\ms a, \ms b]^{\inp}.
  \end{equation}
  \Moreover \cref{eq:thm:bound:def:ABC} \proves that for all $ i, j \in \{ 1, 2, 3 \} $ with $ i \neq j $ it holds that
  \begin{equation}\label{eq:thm:bound:ABC:properties}
    \{ 1, 2, \ldots, \width \} = A_1 \cup A_2 \cup A_3
    \qqandqq
    A_i \cap A_j = \emptyset.
  \end{equation}
  In the following, we distinguish between the case $ L = 0 $, the case $ [ L \in (0,\infty) ] \wedge [ N = 0 ] $,
  the case $ [ L \in (0,\infty) ] \wedge [ 0 < N < \width ] $, and the case $ [ L \in (0,\infty) ] \wedge [ N = \width ] $.
  We first \prove[p] \cref{eq:thm:bound} in the case
  \begin{equation}\label{eq:thm:bound:case1}
    L = 0.
  \end{equation}
  Let $ \vartheta = ( \vartheta_1, \ldots, \vartheta_{\dimension} ) \in \R^{\dimension} $ satisfy for all
  $ i \in \{ 1, 2, \ldots, \dimension - 1 \} $ that
  \begin{equation}\label{eq:thm:bound:vartheta1}
    \vartheta_i = 0
    \qqandqq
    \vartheta_{\dimension} = \mc N^{\theta} (z).
  \end{equation}
  \Nobs that \cref{eq:thm:bound:case1} \proves that for all $ x \in [\ms a, \ms b]^{\inp} $ it holds that
  $ \mc N^{\theta} (x) = \mc N^{\theta} (z) $.
  This and \cref{eq:thm:bound:vartheta1} \prove that for all $ x \in [\ms a, \ms b]^{\inp} $ it holds that
  \begin{equation}\label{eq:thm:bound:realization1}
    \mc N^{\vartheta} (x)
    = \vartheta_{\dimension} + \smallsum_{i=1}^{\width} \vartheta_{ \inp \width + \width + i } \max \{ \vartheta_{ \inp \width + i }
      + \smallsum_{j=1}^{\inp} \vartheta_{ (i-1) \inp + j } x_j, 0 \}
    = \vartheta_{\dimension} = \mc N^{\theta} (z) = \mc N^{\theta} (x).
  \end{equation}
  \Moreover \cref{eq:thm:bound:def:inf} and \cref{eq:thm:bound:vartheta1} \prove that
  \begin{equation}\label{eq:thm:bound:c1}
    \abs{ \vartheta_{\dimension} }
    = \abs{ \mc N^{\theta} (z) }
    = \inf\nolimits_{ x \in [\ms a, \ms b]^{\inp} } \abs{ \mc N^{\theta} (x) }.
  \end{equation}
  The fact that for all $ i \in \{ 1, 2, \ldots, \dimension - 1 \} $ it holds that $ \vartheta_i = 0 $ \hence \proves that
  \begin{equation}
    \max\nolimits_{ i \in \{ 1, 2, \ldots, \dimension \} } \abs{ \vartheta_i }
    \le \max \bigl\{ \max\{ 2, \abs{ \ms a } \sqrt{\inp}, \abs{ \ms b } \sqrt{\inp} \} \sqrt{L},
      \bigl[ \inf\nolimits_{ x \in [\ms a, \ms b]^{\inp} } \abs{ \mc N^{\theta} (x) } \bigr]
      + 2 \width L ( \ms b - \ms a ) \sqrt{\inp} \bigr\}.
  \end{equation}
  Combining this with \cref{eq:thm:bound:realization1} \proves \cref{eq:thm:bound} in the case $ L = 0 $.
  In the next step we \prove[p] \cref{eq:thm:bound} in the case
  \begin{equation}\label{eq:thm:bound:case2}
    [ L \in (0,\infty) ] \wedge [ N = 0 ].
  \end{equation}
  Let $ u = ( u_1, \ldots, u_{\inp} ) $, $ \mf u = ( \mf u_1, \ldots, \mf u_{\inp} ) \in \R^{\inp} $ , $ \delta \in (0,\infty) $
  satisfy for all $ j \in \{ 1, 2, \ldots, \inp \} $ that
  \begin{equation}\label{eq:thm:bound:def:u2}
    u_j = \smallsum_{ i \in A_1 } v_i w_{i,j}, \qquad
    \delta \le \nicefrac{ \varepsilon }{ \max\{ 1, \eucnorm{ u } \} }, \qqandqq
    \mf u = \begin{cases}
      \tfrac{ \sqrt{L} }{ \eucnorm{ u } } u &\colon \eucnorm{ u } > 0 \\
      0 &\colon \eucnorm{ u } = 0.
    \end{cases}
  \end{equation}
  \Nobs that the fact that $ [\ms a, \ms b]^{\inp} \ni x \mapsto \scalprod{ \mf u }{ x } \in \R $ is continuous \proves that there exists
  $ q \in [\ms a, \ms b]^{\inp} $ which satisfies that
  \begin{equation}\label{eq:thm:bound:def:q2}
    \scalprod{ \mf u }{ q } = \inf\nolimits_{ x \in [\ms a, \ms b]^{\inp} } \scalprod{ \mf u }{ x }.
  \end{equation}
  Let $ \vartheta = ( \vartheta_1, \ldots, \vartheta_{\dimension} ) \in \R^{\dimension} $ satisfy for all
  $ i \in \{ 2, 3, \ldots, \width \} $, $ j \in \{ 1, 2, \ldots, \inp \} $ that
  \begin{equation}\label{eq:thm:bound:vartheta2}
  \begin{gathered}
    \vartheta_j = \mf u_j, \qquad
    \vartheta_{ \inp \width + 1 } = - \scalprod{ \mf u }{ q }, \qquad
    \vartheta_{ \inp \width + \width + 1 } = \nicefrac{ \eucnorm{ u } }{ \sqrt{L} }, \\
    \vartheta_{\dimension} = \mc N^{\theta} (z) + \scalprod{ u }{ q - z }, \qqandqq
    \vartheta_{ (i-1) \inp + j } = \vartheta_{ \inp \width + i } = \vartheta_{ \inp \width + \width + i } = 0.
  \end{gathered}
  \end{equation}
  \Nobs that \cref{eq:thm:bound:vartheta2} \proves that for all $ x = ( x_1, \ldots, x_{\inp} ) \in [\ms a, \ms b]^{\inp} $ it holds that
  \begin{equation}\label{eq:thm:bound:realization2:1}
  \begin{split}
    \mc N^{\vartheta} (x)
    &= \vartheta_{\dimension} + \smallsum_{i=1}^{\width} \vartheta_{ \inp \width + \width + i }
      \max \{ \vartheta_{ \inp \width + i } + \smallsum_{j=1}^{\inp} \vartheta_{ (i-1) \inp + j } x_j, 0 \} \\
    &= \vartheta_{\dimension} + \vartheta_{ \inp \width + \width + 1 } \max \bigl\{ \vartheta_{ \inp \width + 1 }
      + \smallsum_{j=1}^{\inp} \vartheta_j x_j, 0 \bigr\} \\
    &= \vartheta_{\dimension} + \tfrac{ \eucnorm{ u } }{ \sqrt{L} } \max \{ \scalprod{ \mf u }{ x } - \scalprod{ \mf u }{ q }, 0 \}.
  \end{split}
  \end{equation}
  \Moreover \cref{eq:thm:bound:def:q2} \proves that for all $ x \in [\ms a, \ms b]^{\inp} $ it holds that
  \begin{equation}
    \scalprod{ \mf u }{ x }
    \ge \inf\nolimits_{ y \in [\ms a, \ms b]^{\inp} } \scalprod{ \mf u }{ y }
    = \scalprod{ \mf u }{ q }.
  \end{equation}
  Combining this, \cref{eq:thm:bound:def:u2}, \cref{eq:thm:bound:vartheta2}, and the fact that $ \eucnorm{ u } \mf u = \sqrt{L} u $
  \proves that for all $ x \in [\ms a, \ms b]^{\inp} $ it holds that
  \begin{equation}\label{eq:thm:bound:realization2:2}
  \begin{split}
    \vartheta_{\dimension} + \tfrac{ \eucnorm{ u } }{ \sqrt{L} } \max \{ \scalprod{ \mf u }{ x } - \scalprod{ \mf u }{ q }, 0 \}
    &= \mc N^{\theta} (z) + \scalprod{ u }{ q - z } + \tfrac{ \eucnorm{ u } }{ \sqrt{L} }
      \bigl( \scalprod{ \mf u }{ x } - \scalprod{ \mf u }{ q } \bigr) \\
    &= \mc N^{\theta} (z) + \scalprod{ u }{ q } - \scalprod{ u }{ z } + \scalprod{ u }{ x } - \scalprod{ u }{ q } \\
    &= \mc N^{\theta} (z) - \scalprod{ u }{ z } + \scalprod{ u }{ x } \\
    &= \mc N^{\theta} (z) - \smallsum_{ i \in A_1 } v_i \scalprod{ w_i }{ z } + \smallsum_{ i \in A_1 } v_i \scalprod{ w_i }{ x }.
  \end{split}
  \end{equation}
  \Moreover \cref{eq:thm:bound:case2} \proves that $ A_2 = \emptyset $.
  The fact that for all $ x \in [\ms a, \ms b]^{\inp} $, $ i \in A_1 $,$ j \in A_3 $ it holds that $ b_i + \scalprod{ w_i }{ x } \ge 0 $ and
  $ b_j + \scalprod{ w_j }{ x } \le 0 $, \cref{eq:thm:bound:def:wbv}, and \cref{eq:thm:bound:ABC:properties} \hence \prove that for all
  $ x \in [\ms a, \ms b]^{\inp} $ it holds that
  \begin{equation}\label{eq:thm:bound:realization2:3}
  \begin{split}
    \mc N^{\theta} (x)
    &= \theta_{\dimension} + \smallsum_{i=1}^{\width} \theta_{ \inp \width + \width + i } \max \{ \theta_{ \inp \width + i }
      + \smallsum_{j=1}^{\inp} \theta_{ (i-1) \inp + j } x_j, 0 \} \\
    &= \theta_{\dimension} + \smallsum_{i=1}^{\width} v_i \max\{ b_i + \scalprod{ w_i }{ x }, 0 \}
    = \theta_{\dimension} + \smallsum_{ i \in A_1 } v_i \bigl( b_i + \scalprod{ w_i }{ x } \bigr) \\
    &= \theta_{\dimension} + \smallsum_{ i \in A_1 } v_i b_i + \smallsum_{ i \in A_1 } v_i \scalprod{ w_i }{ x }.
  \end{split}
  \end{equation}
  Combining this, \cref{eq:thm:bound:realization2:1}, and \cref{eq:thm:bound:realization2:2} \proves that for all
  $ x \in [\ms a, \ms b]^{\inp} $ it holds that
  \begin{equation}\label{eq:thm:bound:realization2}
  \begin{split}
    \mc N^{\vartheta} (x)
    &= \vartheta_{\dimension} + \tfrac{ \eucnorm{ u } }{ \sqrt{L} } \max \{ \scalprod{ \mf u }{ x } - \scalprod{ \mf u }{ q }, 0 \} \\
    &= \mc N^{\theta} (z) - \smallsum_{ i \in A_1 } v_i \scalprod{ w_i }{ z } + \smallsum_{ i \in A_1 } v_i \scalprod{ w_i }{ x } \\
    &= \theta_{\dimension} + \smallsum_{ i \in A_1 } v_i b_i + \smallsum_{ i \in A_1 } v_i \scalprod{ w_i }{ x }
    = \mc N^{\theta} (x).
  \end{split}
  \end{equation}
  Next \nobs that \cref{eq:thm:bound:vartheta2}, the fact that $ \eucnorm{ \mf u } \le \sqrt{L} $, and the Cauchy Schwarz inequality \prove
  that for all $ j \in \{ 1, 2, \ldots, \inp \} $ it holds that
  \begin{equation}\label{eq:thm:bound:wb2}
    \abs{ \vartheta_j } = \abs{ \mf u_j } \le \eucnorm{ \mf u } \le \sqrt{L}
    \qqandqq
    \abs{ \vartheta_{ \inp \width + 1 } } = \abs{ \scalprod{ \mf u }{ q } } \le \eucnorm{ \mf u } \eucnorm{ q }
      \le \sqrt{ \inp L } \max\{ \abs{ \ms a }, \abs{ \ms b } \}.
  \end{equation}
  \Moreover \cref{eq:thm:bound:def:u2}, \cref{eq:thm:bound:realization2:3}, and the fact that $ \ms z + \delta u \in [\ms a, \ms b]^{\inp} $
  \prove that
  \begin{equation}
  \begin{split}
    \abs{ \mc N^{\theta} ( \ms z + \delta u ) - \mc N^{\theta} ( \ms z ) }
    &= \babs{ \smallsum_{ i \in A_1 } v_i \bscalprod{ w_i }{ \ms z + \delta u } - \smallsum_{ i \in A_1 } v_i \scalprod{ w_i }{ \ms z } }
    = \delta \babs{ \smallsum_{ i \in A_1 } v_i \scalprod{ w_i }{ u } } \\
    &= \delta \abs{ \scalprod{ u }{ u } }
    = \delta \eucnorm{ u }^2
    = \eucnorm{ u } \eucnorm{ ( \ms z + \delta u ) - \ms z }.
  \end{split}
  \end{equation}
  \Hence that
  \begin{equation}
    \eucnorm{ u }
    \le \sup\nolimits_{ x, y \in [\ms a, \ms b]^{\inp}, \, x \neq y }
      \tfrac{ \abs{ \mc N^{\theta} (x) - \mc N^{\theta} (y) } }{ \eucnorm{ x - y } }
    = L.
  \end{equation}
  Combining this with \cref{eq:thm:bound:def:inf}, \cref{eq:thm:bound:vartheta2}, and the Cauchy Schwarz inequality \proves that
  \begin{equation}
  \begin{split}
    \abs{ \vartheta_{\dimension} }
    &= \abs{ \mc N^{\theta} (z) + \scalprod{ u }{ q - z } }
    \le \abs{ \mc N^{\theta} (z) } + \eucnorm{ u } \eucnorm{ q - z }
    \le \bigl[ \inf\nolimits_{ x \in [\ms a, \ms b]^{\inp} } \abs{ \mc N^{\theta} (x) } \bigr] + L (\ms b - \ms a) \sqrt{\inp}
  \end{split}
  \end{equation}
  and $ \abs{ \vartheta_{ \inp \width + \width + 1 } } = \nicefrac{ \eucnorm{ u } }{ \sqrt{L} } \le \sqrt{L} $.
  The fact that for all $ i \in \{ 2, 3, \ldots, \width \} $, $ j \in \{ 1, 2, \ldots, \inp \} $ it holds that
  $ \vartheta_{ (i-1) \inp + j } = \vartheta_{ \inp \width + i } = \vartheta_{ \inp \width + \width + i } = 0 $
  and \cref{eq:thm:bound:wb2} \hence \prove that
  \begin{equation}
    \max\nolimits_{ i \in \{ 1, 2, \ldots, \dimension \} } \abs{ \vartheta_i }
    \le \max \bigl\{ \max\{ 2, \abs{ \ms a } \sqrt{\inp}, \abs{ \ms b } \sqrt{\inp} \} \sqrt{L},
      \bigl[ \inf\nolimits_{ x \in [\ms a, \ms b]^{\inp} } \abs{ \mc N^{\theta} (x) } \bigr]
      + 2 \width L ( \ms b - \ms a ) \sqrt{\inp} \bigr\}.
  \end{equation}
  Combining this with \cref{eq:thm:bound:realization2} \proves \cref{eq:thm:bound} in the case $ [ L \in (0,\infty) ] \wedge [ N = 0 ] $.
  Next we \prove[p] \cref{eq:thm:bound} in the case
  \begin{equation}\label{eq:thm:bound:case3}
    [ L \in (0,\infty) ] \wedge [ 0 < N < \width ].
  \end{equation}
  Let $ m_1, m_2, \ldots, m_N \in A_2 $ satisfy for all $ s, t \in \{ 1, 2, \ldots, N \} $ with $ s \neq t $ that
  \begin{equation}\label{eq:thm:bound:def:m}
    \hyper_{w_{m_s},b_{m_s}} \neq \hyper_{w_{m_t},b_{m_t}},
  \end{equation}
  let $ D_s^{\ell} \subseteq \N $, $ s \in \{ 1, 2, \ldots, N \} $, $ \ell \in \{ 0, 1 \} $, satisfy for all $ s \in \{ 1, 2, \ldots, N \} $,
  $ \ell \in \{ 0, 1 \} $ that
  \begin{equation}\label{eq:thm:bound:def:D}
    D_s^{\ell} = \bigl \{ i \in A_2 \colon \bigl[ \hyper_{w_i,b_i} = \hyper_{w_{m_s},b_{m_s}}, \ms z \in \half_{w_i,b_i}^{\ell} \bigr] \bigr\},
  \end{equation}
  and let $ u = ( u_1, \ldots, u_{\inp} ) $, $ \mf u = ( \mf u_1, \ldots, \mf u_{\inp} ) \in \R^{\inp} $ satisfy for all
  $ j \in \{ 1, 2, \ldots, \inp \} $ that
  \begin{equation}\label{eq:thm:bound:def:u3}
    u_j = \smallsum_{ i \in A_1 } v_i w_{i,j} + \smallsum_{s=1}^N \smallsum_{i \in D_s^1} v_i w_{i,j},
    \qqandqq
    \mf u = \begin{cases}
      \tfrac{ \sqrt{L} }{ \eucnorm{ u } } u &\colon \eucnorm{ u } > 0 \\
      0 &\colon \eucnorm{ u } = 0.
    \end{cases}
  \end{equation}
  \Nobs that the fact that $ [\ms a, \ms b]^{\inp} \ni x \mapsto \scalprod{ \mf u }{ x } \in \R $ is continuous \proves that there exists
  $ q \in [\ms a, \ms b]^{\inp} $ which satisfies that
  \begin{equation}\label{eq:thm:bound:def:q3}
    \scalprod{ \mf u }{ q } = \inf\nolimits_{ x \in [\ms a, \ms b]^{\inp} } \scalprod{ \mf u }{ x }.
  \end{equation}
  \Moreover \cref{eq:thm:bound:def:ABC} \proves that for all $ s \in \{ 1, 2, \ldots, N \} $ it holds that $ \eucnorm{ w_{m_s} } > 0 $.
  This and the fact that for all $ s \in \{ 1, 2, \ldots, N \} $ it holds that
  $ \ms z \in \half_{w_{m_s},b_{m_s}}^0 \Delta \half_{w_{m_s},b_{m_s}}^1 $ \prove that there exist
  $ \mf w = ( \mf w_1, \ldots, \mf w_N ) = ( \mf w_{s,j} )_{ (s,j) \in \{ 1, 2, \ldots N \} \times \{ 1, 2, \ldots, \inp \} }
    \in \R^{ N \times \inp } $,
  $ \mf b = ( \mf b_1, \ldots, \mf b_N ) \in \R^N $ which satisfy for all $ s \in \{ 1, 2, \ldots, N \} $ that
  \begin{equation}\label{eq:thm:bound:def:mfwb}
    \mf w_s = \begin{cases}
      \tfrac{ \sqrt{L} }{ \eucnorm{w_{m_s}} } w_{m_s} &\colon \ms z \in \half_{w_{m_s},b_{m_s}}^0 \vspace{0.3cm} \\
      \tfrac{ - \sqrt{L} }{ \eucnorm{w_{m_s}} } w_{m_s} &\colon \ms z \in \half_{w_{m_s},b_{m_s}}^1
    \end{cases}
    \qqandqq
    \mf b_s = \begin{cases}
      \tfrac{ \sqrt{L} }{ \eucnorm{w_{m_s}} } b_{m_s} &\colon \ms z \in \half_{w_{m_s},b_{m_s}}^0 \vspace{0.3cm} \\ 
      \tfrac{ - \sqrt{L} }{ \eucnorm{w_{m_s}} } b_{m_s} &\colon \ms z \in \half_{w_{m_s},b_{m_s}}^1.
    \end{cases}
  \end{equation}
  \Nobs that \cref{eq:thm:bound:case3} \proves that there exists
  $ \vartheta = ( \vartheta_1, \ldots, \vartheta_{\dimension} ) \in \R^{\dimension} $ which satisfies for all
  $ s \in \{ 1, 2, \ldots, N \} $, $ j \in \{ 1, 2, \ldots, d \} $, $ t \in \{ N + 2, N + 3, \ldots, \width \} $ that
  \begin{equation}\label{eq:thm:bound:vartheta3}
  \begin{gathered}
    \vartheta_{ (s-1) \inp + j } = \mf w_{s,j}, \qquad 
    \vartheta_{ \inp \width + s } = \mf b_s, \qquad
    \vartheta_{ \inp \width + \width + s } = \smallsum_{ i \in D_s^0 \cup D_s^1 } \nicefrac{ v_i \eucnorm{ w_i } }{ \sqrt{L} }, \\
    \vartheta_{ N \inp + j } = \mf u_j, \qquad
    \vartheta_{ \inp \width + N + 1 } = - \scalprod{ \mf u }{ q }, \qquad
    \vartheta_{ \inp \width + \width + N + 1 } = \nicefrac{ \eucnorm{ u } }{ \sqrt{L} }, \\
    \vartheta_{\dimension} = \mc N^{\theta} (z) + \scalprod{ u }{ q - z }, \qqandqq
    \vartheta_{ (t-1) \inp + j } = \vartheta_{ \inp \width + t } = \vartheta_{ \inp \width + \width + t } = 0.
  \end{gathered}
  \end{equation}
  \Nobs that \cref{eq:thm:bound:vartheta3} \proves that for all $ x = ( x_1, \ldots, x_{\inp} ) \in [\ms a, \ms b]^{\inp} $ it holds that
  \begin{equation}\label{eq:thm:bound:realization3:1}
  \begin{split}
    \mc N^{\vartheta} (x)
    &= \vartheta_{\dimension} + \smallsum_{i=1}^{\width} \vartheta_{ \inp \width + \width + i }
      \max \{ \vartheta_{ \inp \width + i } + \smallsum_{j=1}^{\inp} \vartheta_{ (i-1) \inp + j } x_j, 0 \} \\
    &= \vartheta_{\dimension} + \smallsum_{s=1}^{N+1} \vartheta_{ \inp \width + \width + s } \max \bigl\{ \vartheta_{ \inp \width + s }
      + \smallsum_{j=1}^{\inp} \vartheta_{ (s-1) \inp + j } x_j, 0 \bigr\} \\
    &= \vartheta_{\dimension}
      + \tfrac{ \eucnorm{ u } }{ \sqrt{L} } \max \{ \scalprod{ \mf u }{ x } - \scalprod{ \mf u }{ q }, 0 \}
      + \smallsum_{s=1}^N \vartheta_{ \inp \width + \width + s } \max\{ \mf b_s + \scalprod{ \mf w_s }{ x }, 0 \}.
  \end{split}
  \end{equation}
  \Moreover \cref{eq:thm:bound:def:q3} \proves that for all $ x \in [\ms a, \ms b]^{\inp} $ it holds that
  \begin{equation}\label{eq:thm:bound:mfu:properties:3}
    \scalprod{ \mf u }{ x }
    \ge \inf\nolimits_{ y \in [\ms a, \ms b]^{\inp} } \scalprod{ \mf u }{ y }
    = \scalprod{ \mf u }{ q }.
  \end{equation}
  Combining this with \cref{eq:thm:bound:realization3:1} and the fact that $ \eucnorm{ u } \mf u = \sqrt{L} u $
  \proves that for all $ x \in [\ms a, \ms b]^{\inp} $ it holds that
  \begin{equation}\label{eq:thm:bound:realization3:2}
  \begin{split}
    \mc N^{\vartheta} (x)
    &= \vartheta_{\dimension}
      + \tfrac{ \eucnorm{ u } }{ \sqrt{L} } \max \{ \scalprod{ \mf u }{ x } - \scalprod{ \mf u }{ q }, 0 \}
      + \smallsum_{s=1}^N \vartheta_{ \inp \width + \width + s } \max\{ \mf b_s + \scalprod{ \mf w_s }{ x }, 0 \} \\
    &= \vartheta_{\dimension} + \tfrac{ \eucnorm{ u } }{ \sqrt{L} } \bigl( \scalprod{ \mf u }{ x } - \scalprod{ \mf u }{ q } \bigr)
      + \smallsum_{s=1}^N \vartheta_{ \inp \width + \width + s } \max\{ \mf b_s + \scalprod{ \mf w_s }{ x }, 0 \} \\
    &= \vartheta_{\dimension} + \scalprod{ u }{ x } - \scalprod{ u }{ q }
      + \smallsum_{s=1}^N \vartheta_{ \inp \width + \width + s } \max\{ \mf b_s + \scalprod{ \mf w_s }{ x }, 0 \}.
  \end{split}
  \end{equation}
  \Moreover \cref{eq:thm:bound:def:u3} and \cref{eq:thm:bound:vartheta3} \prove that for all $ x \in [\ms a, \ms b]^{\inp} $ it holds that
  \begin{equation}\label{eq:thm:bound:realization3:3}
  \begin{split}
    \vartheta_{\dimension} + \scalprod{ u }{ x } - \scalprod{ u }{ q }
    &= \mc N^{\theta} (z) + \scalprod{ u }{ q - z } + \scalprod{ u }{ x } - \scalprod{ u }{ q } \\
    &= \mc N^{\theta} (z) - \scalprod{ u }{ z } + \scalprod{ u }{ x } \\
    &= \mc N^{\theta} (z) - \smallsum_{ i \in A_1 } v_i \scalprod{ w_i }{ z }
      - \smallsum_{s=1}^N \smallsum_{ i \in D_s^1 } v_i \scalprod{ w_i }{ z } \\
    &\quad + \smallsum_{ i \in A_1 } v_i \scalprod{ w_i }{ x }
      + \smallsum_{s=1}^N \smallsum_{ i \in D_s^1 } v_i \scalprod{ w_i }{ x }.
  \end{split}
  \end{equation}
  \Moreover \cref{lem:hyperplane:scalar} (applied for every $ s \in \{ 1, 2, \ldots, N \} $, $ i \in D_s^0 $ with
  $ d \is \inp $, $ z \is \ms z $, $ w_1 \is w_i $, $ w_2 \is \mf w_s $, $ b_1 \is b_i $, $ b_2 \is \mf b_s $
  in the notation of \cref{lem:hyperplane:scalar})
  and the fact that for all $ s \in \{ 1, 2, \ldots, N \} $ it holds that $ \eucnorm{ \mf w_s } = \sqrt{L} $
  \prove that for all $ s \in \{ 1, 2, \ldots, N \} $, $ i \in D_s^0 $ it holds that
  \begin{equation}\label{eq:thm:bound:w_relation_0}
    \eucnorm{ w_i } \mf w_s = \eucnorm{ \mf w_s } w_i = \sqrt{L} w_i
    \qqandqq
    \eucnorm{ w_i } \mf b_s = \eucnorm{ \mf w_s } b_i = \sqrt{L} b_i.
  \end{equation}
  \Moreover \cref{lem:hyperplane:scalar} (applied for every $ s \in \{ 1, 2, \ldots, N \} $, $ i \in D_s^1 $ with
  $ d \is \inp $, $ z \is \ms z $, $ w_1 \is - w_i $, $ w_2 \is \mf w_s $, $ b_1 \is - b_i $, $ b_2 \is \mf b_s $
  in the notation of \cref{lem:hyperplane:scalar})
  and the fact that for all $ s \in \{ 1, 2, \ldots, N \} $ it holds that $ \eucnorm{ \mf w_s } = \sqrt{L} $
  \prove that for all $ s \in \{ 1, 2, \ldots, N \} $, $ i \in D_s^1 $ it holds that
  \begin{equation}\label{eq:thm:bound:w_relation_1}
    \eucnorm{ w_i } \mf w_s = - \eucnorm{ \mf w_s } w_i = - \sqrt{L} w_i
    \qqandqq
    \eucnorm{ w_i } \mf b_s = - \eucnorm{ \mf w_s } b_i = - \sqrt{L} b_i.
  \end{equation}
  Combining this and \cref{eq:thm:bound:w_relation_0} \proves that for all
  $ s \in \{ 1, 2, \ldots, N \} $, $ x \in \half_{\mf w_s, \mf b_s}^0 $, $ y \in \half_{\mf w_s, \mf b_s}^1 $,
  $ i \in D_s^0 $, $ j \in D_s^1 $ it holds that
  \begin{equation}\label{eq:thm:bound:B:properties}
  \begin{split}
    \scalprod{ w_i }{ x } + b_i &= \tfrac{ \eucnorm{ w_i } }{ \sqrt{L} } \bigl( \mf b_s + \scalprod{ \mf w_s }{ x } \bigr) \le 0, \\
    \scalprod{ w_j }{ x } + b_j &= - \tfrac{ \eucnorm{ w_i } }{ \sqrt{L} } \bigl( \mf b_s + \scalprod{ \mf w_s }{ x } \bigr) \ge 0, \\
    \scalprod{ w_i }{ y } + b_i &= \tfrac{ \eucnorm{ w_i } }{ \sqrt{L} } \bigl( \mf b_s + \scalprod{ \mf w_s }{ y } \bigr) \ge 0,
      \qquad\text{and} \\
    \scalprod{ w_j }{ y } + b_j &= - \tfrac{ \eucnorm{ w_i } }{ \sqrt{L} } \bigl( \mf b_s + \scalprod{ \mf w_s }{ y } \bigr) \le 0.
  \end{split}
  \end{equation}
  The fact that for all $ x \in [\ms a, \ms b]^{\inp} $, $ i \in A_1 $, $ j \in A_3 $ it holds that
  $ b_i + \scalprod{ w_i }{ x } \ge 0 $ and $ b_j + \scalprod{ w_j }{ x } \le 0 $,
  the fact that $ A_2 = \bigcup_{s=1}^N ( D_s^0 \cup D_s^1 ) $,
  the fact that for all $ s, t \in \{ 1, 2, \ldots, N \} $ with $ s \neq t $ it holds that
  $ D_s^0 \cap D_s^1 = \emptyset $, $ D_s^0 \cap D_t^0 = \emptyset $, and $ D_s^1 \cap D_t^1 = \emptyset $,
  \cref{eq:thm:bound:ABC:properties}, and \cref{eq:thm:bound:B:properties} \hence
  \prove that for all $ \ell_1, \ell_2, \ldots, \ell_N \in \{ 0, 1 \} $,
  $ x = ( x_1, \ldots, x_{\inp} ) \in ( \bigcap_{s=1}^N \half_{\mf w_s, \mf b_s}^{\ell_s} ) \cap [\ms a, \ms b]^{\inp} $
  it holds that
  \begin{equation}\label{eq:thm:bound:realization3:4}
  \begin{split}
    \mc N^{\theta} (x)
    &= \theta_{\dimension} + \smallsum_{i=1}^{\width} \theta_{ \inp \width + \width + i } \max \{ \theta_{ \inp \width + i }
      + \sum_{j=1}^{\inp} \theta_{ (i-1) \inp + j } x_j, 0 \} \\
    &= \theta_{\dimension} + \smallsum_{i=1}^{\width} v_i \max \{ b_i + \scalprod{ w_i }{ x }, 0 \} \\
    &= \theta_{\dimension} + \smallsum_{ i \in A_1 } v_i \bigl( b_i + \scalprod{ w_i }{ x } \bigr)
      + \smallsum_{ s \in \{ 1, 2, \ldots, N \}, \, \ell_s = 1 } \smallsum_{ i \in D_s^0 } v_i
      \bigl( b_i + \scalprod{ w_i }{ x } \bigr) \\
    &\quad + \smallsum_{ s \in \{ 1, 2, \ldots, N \}, \, \ell_s = 0 } \smallsum_{ i \in D_s^1 } v_i
        \bigl( b_i + \scalprod{ w_i }{ x } \bigr).
  \end{split}
  \end{equation}
  This and the fact that $ z \in ( \bigcap_{s=1}^N \half_{\mf w_s, \mf b_s}^0 ) \cap [\ms a, \ms b]^{\inp} $
  \prove that for all $ x \in \R^{\inp} $ it holds that
  \begin{equation}\label{eq:thm:bound:realization3:5}
  \begin{split}
    &\mc N^{\theta} (z) - \smallsum_{ i \in A_1 } v_i \scalprod{ w_i }{ z }
      - \smallsum_{s=1}^N \smallsum_{ i \in D_s^1 } v_i \scalprod{ w_i }{ z }
      + \smallsum_{ i \in A_1 } v_i \scalprod{ w_i }{ x }
      + \smallsum_{s=1}^N \smallsum_{ i \in D_s^1 } v_i \scalprod{ w_i }{ x } \\
    &= \theta_{\dimension} + \smallsum_{ i \in A_1 } v_i \bigl( b_i + \scalprod{ w_i }{ z } \bigr)
      + \smallsum_{s=1}^N \smallsum_{ i \in D_s^1 } v_i \bigl( b_i + \scalprod{ w_i }{ z } \bigr)
      - \smallsum_{ i \in A_1 } v_i \scalprod{ w_i }{ z } \\
    &\quad - \smallsum_{s=1}^N \smallsum_{ i \in D_s^1 } v_i \scalprod{ w_i }{ z }
      + \smallsum_{ i \in A_1 } v_i \scalprod{ w_i }{ x }
      + \smallsum_{s=1}^N \smallsum_{ i \in D_s^1 } v_i \scalprod{ w_i }{ x } \\
    &= \theta_{\dimension} + \smallsum_{ i \in A_1 } v_i b_i + \smallsum_{s=1}^N \smallsum_{ i \in D_s^1 } v_i b_i
      + \smallsum_{ i \in A_1 } v_i \scalprod{ w_i }{ x } + \smallsum_{s=1}^N \smallsum_{ i \in D_s^1 } v_i \scalprod{ w_i }{ x } \\
    &= \theta_{\dimension} + \smallsum_{ i \in A_1 } v_i \bigl( b_i + \scalprod{ w_i }{ x } \bigr)
      + \smallsum_{s=1}^N \smallsum_{ i \in D_s^1 } v_i \bigl( b_i + \scalprod{ w_i }{ x } \bigr).
  \end{split}
  \end{equation}
  \Moreover the fact that for all $ s \in \{ 1, 2, \ldots, N \} $,
  $ x \in \half_{\mf w_s, \mf b_s}^0 $, $ y \in \half_{\mf w_s, \mf b_s}^1 $ it holds that
  $ \mf b_s + \scalprod{ \mf w_s }{ x } \le 0 $ and $ \mf b_s + \scalprod{ \mf w_s }{ y } \ge 0 $,
  the fact that for all $ s \in \{ 1, 2, \ldots, N \} $ it holds that $ D_s^0 \cap D_s^1 = \emptyset $,
  \cref{eq:thm:bound:w_relation_0}, and \cref{eq:thm:bound:w_relation_1} \prove that for all
  $ \ell_1, \ell_2, \ldots, \ell_N \in \{ 0, 1 \} $, $ x \in ( \bigcap_{s=1}^N \half_{\mf w_s, \mf b_s}^{\ell_s} ) $
  it holds that
  \begin{equation}
  \begin{split}
    \smallsum_{s=1}^N \vartheta_{ \inp \width + \width + s } \max\{ \mf b_s + \scalprod{ \mf w_s }{ x }, 0 \}
    &= \smallsum_{ s \in \{ 1, \ldots, N \}, \, \ell_s = 1 } \vartheta_{ \inp \width + \width + s }
      \bigl( \mf b_s + \scalprod{ \mf w_s }{ x } \bigr) \\
    &= \smallsum_{ s \in \{ 1, 2, \ldots, N \}, \, \ell_s = 1 } \smallsum_{ i \in D_s^0 \cup D_s^1 }
      \tfrac{ v_i \eucnorm{ w_i } }{ \sqrt{L} } \bigl( \mf b_s + \scalprod{ \mf w_s }{ x  } \bigr) \\
    &= \smallsum_{ s \in \{ 1, 2, \ldots, N \}, \, \ell_s = 1 } \smallsum_{ i \in D_s^0 }
      \tfrac{ v_i \eucnorm{ w_i } }{ \sqrt{L} } \bigl( \mf b_s + \scalprod{ \mf w_s }{ x  } \bigr) \\
    &\quad + \smallsum_{ s \in \{ 1, 2, \ldots, N \}, \, \ell_s = 1 } \smallsum_{ i \in D_s^1 }
      \tfrac{ v_i \eucnorm{ w_i } }{ \sqrt{L} } \bigl( \mf b_s + \scalprod{ \mf w_s }{ x  } \bigr) \\
    &= \smallsum_{ s \in \{ 1, 2, \ldots, N \}, \, \ell_s = 1 } \smallsum_{ i \in D_s^0 }
      v_i \bigl( b_i + \scalprod{ w_i }{ x } \bigr) \\
    &\quad - \smallsum_{ s \in \{ 1, 2, \ldots, N \}, \, \ell_s = 1 } \smallsum_{ i \in D_s^1 }
      v_i \bigl( b_i + \scalprod{ w_i }{ x } \bigr).
  \end{split}
  \end{equation}
  Combining this, \cref{eq:thm:bound:realization3:2}, \cref{eq:thm:bound:realization3:3}, \cref{eq:thm:bound:realization3:4}, and
  \cref{eq:thm:bound:realization3:5} \proves that for all
  $ \ell_1, \ell_2, \ldots, \ell_N \in \{ 0, 1 \} $,
  $ x \in ( \bigcap_{s=1}^N \half_{\mf w_s, \mf b_s}^{\ell_s} ) \cap [\ms a, \ms b]^{\inp} $
  it holds that
  \begin{equation}\label{eq:thm:bound:realization:6}
  \begin{split}
    \mc N^{\vartheta} (x)
    &= \vartheta_{\dimension} + \scalprod{ u }{ x } - \scalprod{ u }{ q }
      + \smallsum_{s=1}^N \vartheta_{ \inp \width + \width + s } \max\{ \mf b_s + \scalprod{ \mf w_s }{ x }, 0 \} \\
    &= \mc N^{\theta} (z) - \smallsum_{ i \in A_1 } v_i \scalprod{ w_i }{ z }
      - \smallsum_{s=1}^N \smallsum_{ i \in D_s^1 } v_i \scalprod{ w_i }{ z }
      + \smallsum_{ i \in A_1 } v_i \scalprod{ w_i }{ x } \\
    &\quad + \smallsum_{s=1}^N \smallsum_{ i \in D_s^1 } v_i \scalprod{ w_i }{ x }
      + \smallsum_{s=1}^N \vartheta_{ \inp \width + \width + s } \max\{ \mf b_s + \scalprod{ \mf w_s }{ x }, 0 \} \\
    &= \theta_{\dimension} + \smallsum_{ i \in A_1 } v_i \bigl( b_i + \scalprod{ w_i }{ x } \bigr)
      + \smallsum_{s=1}^N \smallsum_{ i \in D_s^1 } v_i \bigl( b_i + \scalprod{ w_i }{ x } \bigr) \\
    &\quad + \smallsum_{ s \in \{ 1, 2, \ldots, N \}, \, \ell_s = 1 } \smallsum_{ i \in D_s^0 }
      v_i \bigl( b_i + \scalprod{ w_i }{ x } \bigr) \\
    &\quad - \smallsum_{ s \in \{ 1, 2, \ldots, N \}, \, \ell_s = 1 } \smallsum_{ i \in D_s^1 }
      v_i \bigl( b_i + \scalprod{ w_i }{ x } \bigr) \\
    &= \theta_{\dimension} + \smallsum_{ i \in A_1 } v_i \bigl( b_i + \scalprod{ w_i }{ x } \bigr)
      + \smallsum_{ s \in \{ 1, 2, \ldots, N \}, \, \ell_s = 1 } \smallsum_{ i \in D_s^0 } v_i \bigl( b_i + \scalprod{ w_i }{ x } \bigr) \\
    &\qquad + \smallsum_{ s \in \{ 1, 2, \ldots, N \}, \, \ell_s = 0 } \smallsum_{ i \in D_s^1 } v_i \bigl( b_i + \scalprod{ w_i }{ x } \bigr) \\
    &= \mc N^{\theta} (x).
  \end{split}
  \end{equation}
  The fact that
  $ [\ms a, \ms b]^{\inp} \subseteq \smallbigcup_{ \ell_1, \ell_2, \ldots, \ell_N \in \{ 0, 1 \} }
    ( \smallbigcap_{s=1}^N \half_{\mf w_s, \mf b_s}^{\ell_s} ) $
  \hence \proves that for all $ x \in [\ms a, \ms b]^{\inp} $ it holds that
  \begin{equation}\label{eq:thm:bound:realization3}
    \mc N^{\vartheta} (x) = \mc N^{\theta} (x).
  \end{equation}
  \Moreover the fact that for all $ s \in \{ 1, 2, \ldots, N \} $ it holds that $ \eucnorm{ \mf w_s } = \sqrt{L} $,
  the fact that $ \eucnorm{ \mf u } \le \sqrt{L} $, and \cref{eq:thm:bound:vartheta3} \prove that for all
  $ s \in \{ 1, 2, \ldots, N \} $, $ j \in \{ 1, 2, \ldots, \inp \} $ it holds that
  \begin{equation}\label{eq:thm:bound:w3}
    \abs{ \vartheta_{ (s-1) \inp + j } } = \abs{ \mf w_{s,j} } \le \eucnorm{ \mf w_s } = \sqrt{L}
    \qqandqq
    \abs{ \vartheta_{ N \inp + j } } = \abs{ \mf u_j } \le \eucnorm{ \mf u } \le \sqrt{L}.
  \end{equation}
  \Moreover \cref{lem:hyperplanes:isolation} (applied with
  $ d \is \inp $, $ N \is N $, $ \ms a \is \ms a $, $ \ms b \is \ms b $,
  $ ( w_i )_{ i \in \{ 1, 2, \ldots, \allowbreak N \} } \allowbreak \is ( \mf w_s )_{ s \in \{ 1, 2, \ldots, N \} } $,
  $ ( b_i )_{ i \in \{ 1, 2, \ldots, N \} } \is ( \mf b_s )_{ s \in \{ 1, 2, \ldots, N \} } $
  in the notation of \cref{lem:hyperplanes:isolation})
  \proves that there exist
  $ p_1, p_2, \ldots, p_N \in (\ms a, \ms b)^{\inp} $, $ \delta \in ( 0, \nicefrac{ \varepsilon }{ \max\{1, \eucnorm{u} \} } ) $
  which satisfy that
  \begin{enumerate}[(i)]
    \item \label{item:thm:bound:def:p3:1} it holds for all $ s \in \{ 1, 2, \ldots, N \} $ that
      $ p_s \in \hyper_{\mf w_s, \mf b_s} $,
    \item \label{item:thm:bound:def:p3:2} it holds for all $ s \in \{ 1, 2, \ldots, N \} $ that
      $ \{ x \in \R^{\inp} \colon \eucnorm{ x - p_s } \le \delta \} \subseteq [\ms a, \ms b]^{\inp} $, and
    \item \label{item:thm:bound:def:p3:3} it holds for all $ s \in \{ 1, 2, \ldots, N \} $ that
      $ \{ x \in \R^{\inp} \colon \eucnorm{ x - p_s } \le \delta \} \cap
        ( \smallbigcup_{ t \in \{ 1, \ldots, N \} \backslash\{ s \} } \hyper_{\mf w_t, \mf b_t} ) = \emptyset $.
  \end{enumerate}
  \Nobs that \cref{item:thm:bound:def:p3:1} \proves that for all $ s \in \{ 1, 2, \ldots, N \} $ it holds that
  $ \mf b_s + \scalprod{ \mf w_s }{ p_s } = 0 $.
  The fact that for all $ s \in \{ 1, 2, \ldots, N \} $ it holds that $ \eucnorm{ \mf w_s } = \sqrt{L} $,
  the fact that $ \eucnorm{ \mf u } \le \sqrt{L} $, the Cauchy Schwarz inequality, and \cref{eq:thm:bound:vartheta3}
  \hence \prove that for all $ s \in \{ 1, 2, \ldots, N \} $, $ j \in \{ 1, 2, \ldots, \inp \} $ it holds that
  \begin{equation}\label{eq:thm:bound:b3:1}
    \abs{ \vartheta_{ \inp \width + s } }
    = \abs{ \mf b_s }
    = \abs{ \scalprod{ \mf w_s }{ p_s } }
    \le \eucnorm{ \mf w_s } \eucnorm{ p_s }
    \le \sqrt{\inp L} \max\{ \abs{ \ms a }, \abs{ \ms b } \}
  \end{equation}
  and
  \begin{equation}\label{eq:thm:bound:b3:2}
    \abs{ \vartheta_{ \inp \width + N + 1 } }
    = \abs{ \scalprod{ \mf u }{ q } }
    \le \eucnorm{ \mf u } \eucnorm{ q }
    \le \sqrt{\inp L} \max\{ \abs{ \ms a }, \abs{ \ms b } \}.
  \end{equation}
  \Moreover \cref{eq:thm:bound:realization3:2} and the fact that for all
  $ s \in \{ 1, 2, \ldots, N \} $, $ x \in \half_{\mf w_s, \mf b_s}^0 $, $ y \in \half_{\mf w_s, \mf b_s}^1 $
  it holds that $ \mf b_s + \scalprod{ \mf w_s }{ x } \le 0 $ and $ \mf b_s + \scalprod{ \mf w_s }{ y } \ge 0 $
  \proves that for all
  $ \ell_1, \ell_2, \ldots, \ell_N \in \{ 0, 1 \} $,
  $ x, y \in ( \smallbigcap_{s=1}^N \half_{\mf w_s, \mf b_s}^{\ell_s} ) \cap [\ms a, \ms b]^{\inp} $ it holds that
  \begin{equation}\label{eq:thm:bound:diff3}
  \begin{split}
    \mc N^{\vartheta} (x) - \mc N^{\vartheta} (y)
    &= \Bigl[ \vartheta_{\dimension} + \scalprod{ u }{ x } - \scalprod{ u }{ q }
      + \smallsum_{s=1}^N \vartheta_{ \inp \width + \width + s } \max\{ \mf b_s + \scalprod{ \mf w_s }{ x }, 0 \} \Bigr] \\
    &\quad - \Bigl[ \vartheta_{\dimension} + \scalprod{ u }{ y } - \scalprod{ u }{ q }
      + \smallsum_{s=1}^N \vartheta_{ \inp \width + \width + s } \max\{ \mf b_s + \scalprod{ \mf w_s }{ y }, 0 \} \Bigr] \\
    &= \scalprod{ u }{ x - y }
      + \smallsum_{ s \in \{ 1, 2, \ldots, N \}, \, \ell_s = 1 } \vartheta_{ \inp \width + \width + s }
        \bigl( \mf b_s + \scalprod{ \mf w_s }{ x } \bigr) \\
    &\quad - \smallsum_{ s \in \{ 1, 2, \ldots, N \}, \, \ell_s = 1 } \vartheta_{ \inp \width + \width + s }
        \bigl( \mf b_s + \scalprod{ \mf w_s }{ y } \bigr) \\
    &= \scalprod{ u }{ x - y } + \smallsum_{ s \in \{ 1, 2, \ldots, N \}, \, \ell_s = 1 } \vartheta_{ \inp \width + \width + s }
      \scalprod{ \mf w_s }{ x - y }.
  \end{split}
  \end{equation}
  \Moreover the fact that
  $ [\ms a, \ms b]^{\inp} \subseteq \smallbigcup_{ \ell_1, \ell_2, \ldots, \ell_N \in \{ 0, 1 \} }
    ( \smallbigcap_{s=1}^N \half_{\mf w_s, \mf b_s}^{\ell_s} ) $
  and \cref{item:thm:bound:def:p3:1} \prove that for all $ s \in \{ 1, 2, \ldots, N \} $ there exist
  $ \ell_1, \ell_2, \ldots, \ell_N \in \{ 0, 1 \}  $ such that
  $ p_s \in ( \smallbigcap_{s=1}^N \half_{\mf w_s, \mf b_s}^{\ell_s} ) $ and
  $ \ell_s = 1 $.
  Combining this, \cref{item:thm:bound:def:p3:2,item:thm:bound:def:p3:3}, \cref{eq:thm:bound:diff3}, and the fact that for all
  $ s \in \{ 1, 2, \ldots, N \} $ it holds that $ \eucnorm{ \mf w_s } = \sqrt{L} $ \proves that for all
  $ s \in \{ 1, 2, \ldots, N \} $ there exists $ \ell_1, \ell_2, \ldots, \ell_N \in \{ 0, 1 \} $ such that
  \begin{equation}
  \begin{split}
    &\mc N^{\vartheta} \bigl( p_s + \tfrac{ \delta }{ \eucnorm{ \mf w_s } } \mf w_s \bigr) - \mc N^{\vartheta} ( p_s ) \\
    &= \bscalprod{ u }{ \tfrac{ \delta }{ \eucnorm{ \mf w_s } } \mf w_s } + \smallsum_{ t \in \{ 1, 2, \ldots, N \}, \, \ell_t = 1 }
      \vartheta_{ \inp \width + \width + t } \bscalprod{ \mf w_t }{ \tfrac{ \delta }{ \eucnorm{ \mf w_s } } \mf w_s } \\
    &= \bscalprod{ u }{ \tfrac{ \delta }{ \eucnorm{ \mf w_s } } \mf w_s }
      + \smallsum_{ t \in \{ 1, 2, \ldots, N \} \backslash \{ s \}, \, \ell_t = 1 }
      \vartheta_{ \inp \width + \width + t } \bscalprod{ \mf w_t }{ \tfrac{ \delta }{ \eucnorm{ \mf w_s } } \mf w_s }
      + \vartheta_{ \inp \width + \width + s } \tfrac{ \delta }{ \eucnorm{ \mf w_s } } \scalprod{ \mf w_s }{ \mf w_s } \\
    &= \mc N^{\vartheta} ( p_s ) - \mc N^{\vartheta} \bigl( p_s - \tfrac{ \delta }{ \eucnorm{ \mf w_s } } \mf w_s \bigr)
      + \delta \vartheta_{ \inp \width + \width + s } \sqrt{L}.
  \end{split}
  \end{equation}
  This and \cref{eq:thm:bound:realization3} \prove that for all $ s \in \{ 1, 2, \ldots, N \} $ it holds that
  \begin{equation}\label{eq:thm:bound:v3}
  \begin{split}
    \abs{ \vartheta_{ \inp \width + \width + s } }
    &\le \tfrac{1}{\sqrt{L}} \Bigl[ \tfrac{ 1 }{ \delta } \babs{
      \mc N^{\vartheta} \bigl( p_s + \tfrac{ \delta }{ \eucnorm{ \mf w_s } } \mf w_s \bigr) - \mc N^{\vartheta} ( p_s ) } +
      \tfrac{ 1 }{ \delta } \babs{
      \mc N^{\vartheta} ( p_s ) - \mc N^{\vartheta} \bigl( p_s - \tfrac{ \delta }{ \eucnorm{ \mf w_s } } \mf w_s \bigr) } \Bigr] \\
    &= \tfrac{1}{\sqrt{L}} \Bigl[ \tfrac{ 1 }{ \delta } \babs{
      \mc N^{\theta} \bigl( p_s + \tfrac{ \delta }{ \eucnorm{ \mf w_s } } \mf w_s \bigr) - \mc N^{\theta} ( p_s ) } +
      \tfrac{ 1 }{ \delta } \babs{
      \mc N^{\theta} ( p_s ) - \mc N^{\theta} \bigl( p_s - \tfrac{ \delta }{ \eucnorm{ \mf w_s } } \mf w_s \bigr) } \Bigr] \\
    &\le \tfrac{2}{\sqrt{L}} \Bigl( \sup\nolimits_{ x, y \in [\ms a, \ms b]^{\inp}, \, x \neq y }
      \tfrac{ \abs{ \mc N^{\theta} (x) - \mc N^{\theta} (y) } }{ \eucnorm{ x - y } } \Bigr)
    = 2 \sqrt{L}.
  \end{split}
  \end{equation}
  \Moreover \cref{eq:thm:bound:def:u3}, \cref{eq:thm:bound:realization3}, \cref{eq:thm:bound:diff3},
  the fact that $ \ms z \in ( \smallbigcap_{s=1}^N \half_{\mf w_s, \mf b_s}^0 ) $, and the fact that
  $ \ms z + \delta u \in ( \smallbigcap_{s=1}^N \half_{\mf w_s, \mf b_s}^0 ) $ \prove that
  \begin{equation}
    \abs{ \mc N^{\theta} ( \ms z + \delta u ) - \mc N^{\theta} (\ms z) }
    = \abs{ \mc N^{\vartheta} ( \ms z + \delta u ) - \mc N^{\vartheta} (\ms z) }
    = \abs{ \scalprod{ u }{ \delta u } }
    = \delta \eucnorm{ u }^2
    = \eucnorm{ u } \eucnorm{ ( \ms z + \delta u ) - \ms z }.
  \end{equation}
  \Hence that
  \begin{equation}\label{eq:thm:bound:u3}
      \eucnorm{ u } \le \sup\nolimits_{ x, y \in [\ms a, \ms b]^{\inp}, \, x \neq y }
        \tfrac{ \abs{ \mc N^{\theta} (x) - \mc N^{\theta} (y) } }{ \eucnorm{ x - y } } = L.
  \end{equation}
  Combining this with \cref{eq:thm:bound:def:inf}, \cref{eq:thm:bound:vartheta3}, and the Cauchy Schwarz inequality \proves that
  \begin{equation}
  \begin{split}
    \abs{ \vartheta_{\dimension} }
    &= \abs{ \mc N^{\theta} (z) + \scalprod{ u }{ q - z } }
    \le \abs{ \mc N^{\theta} (z) } + \abs{ \scalprod{ u }{ q - z } } \\
    &\le \bigl( \inf\nolimits_{ x \in [\ms a, \ms b]^{\inp} } \abs{ \mc N^{\theta} (x) } \bigr) + \eucnorm{ u } \eucnorm{ q - z }
    \le \bigl( \inf\nolimits_{ x \in [\ms a, \ms b]^{\inp} } \abs{ \mc N^{\theta} (x) } \bigr) + L ( \ms b - \ms a ) \sqrt{\inp}
  \end{split}
  \end{equation}
  and $ \abs{ \vartheta_{ \inp \width + \width + N + 1 } } = \nicefrac{ \eucnorm{ u } }{ \sqrt{L} } \le \sqrt{L} $.
  This, \cref{eq:thm:bound:w3}, \cref{eq:thm:bound:b3:1}, \cref{eq:thm:bound:b3:2}, \cref{eq:thm:bound:v3}, and
  the fact that for all $ t \in \{ N+2, N+3, \ldots, \width \} $, $ j \in \{ 1, 2, \ldots, \inp \} $ it holds that
  $ \vartheta_{ (t-1) \inp + j } = \vartheta_{ \inp \width + t } = \vartheta_{ \inp \width + \width + t } = 0 $ \prove that
  \begin{equation}
    \max\nolimits_{ i \in \{ 1, 2, \ldots, \dimension \} } \abs{ \vartheta_i }
    \le \max \bigl\{ \max\{ 2, \abs{ \ms a } \sqrt{\inp}, \abs{ \ms b } \sqrt{\inp} \} \sqrt{L},
      \bigl[ \inf\nolimits_{ x \in [\ms a, \ms b]^{\inp} } \abs{ \mc N^{\theta} (x) } \bigr]
      + 2 \width L ( \ms b - \ms a ) \sqrt{\inp} \bigr\}.
  \end{equation}
  Combining this with \cref{eq:thm:bound:realization3} \proves \cref{eq:thm:bound} in the case
  $ [ L \in (0,\infty) ] \wedge [ N < \width ] $.
  In the last step we \prove[p] \cref{eq:thm:bound} in the case
  \begin{equation}\label{eq:thm:bound:case4}
    [ L \in (0,\infty) ] \wedge [ N = \width ].
  \end{equation}
  \Nobs that \cref{eq:thm:bound:def:ABC} and \cref{eq:thm:bound:case4} \prove that for all $ i \in \{ 1, 2, \ldots, \width \} $ it holds that
  $ \eucnorm{ w_i } > 0 $.
  \Hence that there exists $ \vartheta = ( \vartheta_1, \ldots, \vartheta_{\dimension} ) \in \R^{\dimension} $ which satisfies for all
  $ i \in \{ 1, 2, \ldots, \width \} $, $ j \in \{ 1, 2, \ldots, \inp \} $ that
  \begin{equation}\label{eq:thm:bound:vartheta4}
    \vartheta_{ (i-1) \inp + j } = \tfrac{ \sqrt{L} w_{i,j} }{ \eucnorm{ w_i } }, \qquad
    \vartheta_{ \inp \width + i } = \tfrac{ \sqrt{L} b_i }{ \eucnorm{ w_i } }, \qquad
    \vartheta_{ \inp \width + \width + i } = \tfrac{ v_i \eucnorm{ w_i } }{ \sqrt{L} }, \qqandqq
    \vartheta_{\dimension} = \theta_{\dimension}.
  \end{equation}
  \Nobs that \cref{eq:thm:bound:def:wbv} and \cref{eq:thm:bound:vartheta4} \prove that for all
  $ x = ( x_1, \ldots, x_{\inp} ) \in [\ms a, \ms b]^{\inp} $ it holds that
  \begin{equation}\label{eq:thm:bound:realization4}
  \begin{split}
    \mc N^{\vartheta} (x)
    &= \vartheta_{ \mf d } + \smallsum_{i=1}^{\width} \vartheta_{ \inp \width + \width + i } \max \bigl\{ \vartheta_{ \inp \width + i }
      + \smallsum_{j=1}^{\inp} \vartheta_{ (i-1) \inp + j } x_j, 0 \bigr\} \\
    &= \theta_{\dimension} + \smallsum_{i=1}^{\width} \tfrac{ v_i \eucnorm{ w_i } }{ \sqrt{L} } \max \Bigl\{
      \tfrac{ \sqrt{L} b_i }{ \eucnorm{ w_i } } + \smallsum_{j=1}^{\inp} \tfrac{ \sqrt{L} w_{i,j} }{ \eucnorm{ w_i } } x_j, 0 \Bigr\} \\
    &= \theta_{\dimension} + \smallsum_{i=1}^{\width} v_i \max \bigl\{ b_i + \smallsum_{j=1}^{\inp} w_{i,j} x_j, 0 \bigr\} \\
    &= \theta_{\dimension} + \smallsum_{i=1}^{\width} \theta_{ \inp \width + \width + i } \max \bigl\{ \theta_{ \inp \width + i }
      + \smallsum_{j=1}^{\inp} \theta_{ (i-1) \inp + j } x_j, 0 \bigr\}
    = \mc N^{\theta} (x).
  \end{split}
  \end{equation}
  \Moreover \cref{eq:thm:bound:vartheta4} \proves that for all $ i \in \{ 1, 2, \ldots, \width \} $, $ j \in \{ 1, 2, \ldots, \inp \} $
  it holds that
  \begin{equation}\label{eq:thm:bound:w4}
    \abs{ \vartheta_{ (i-1) \inp + j } }
    = \tfrac{ \sqrt{L} \abs{ w_{i,j} } }{ \eucnorm{ w_i } }
    \le \sqrt{L}.
  \end{equation}
  \Moreover \cref{lem:hyperplanes:isolation} (applied with
  $ d \is \inp $, $ N \is \width $, $ \ms a \is \ms a $, $ \ms b \is \ms b $,
  $ ( w_i )_{ i \in \{ 1, 2, \ldots, N \} } \is ( w_i )_{ i \in \{ 1, 2, \ldots, \width \} } $,
  $ ( b_i )_{ i \in \{ 1, 2, \ldots, N \} } \is ( b_i )_{ i \in \{ 1, 2, \ldots, \width \} } $
  in the notation of \cref{lem:hyperplanes:isolation})
  \proves that there exist
  $ p_1, p_2, \ldots, p_{\width} \in (\ms a, \ms b)^{\inp} $, $ \delta \in (0,\infty) $ which satisfy that
  \begin{enumerate}[(i)]
    \item \label{item:thm:bound:def:p4:1} it holds for all $ i \in \{ 1, 2, \ldots, \width \} $ that
      $ p_i \in \hyper_{w_i,b_i} $,
    \item \label{item:thm:bound:def:p4:2} it holds for all $ i \in \{ 1, 2, \ldots, \width \} $ that
      $ \{ x \in \R^{\inp} \colon \eucnorm{ x - p_i } \le \delta \} \subseteq [\ms a, \ms b]^{\inp} $, and
    \item \label{item:thm:bound:def:p4:3} it holds for all $ i \in \{ 1, 2, \ldots, \width \} $ that
      $ \{ x \in \R^{\inp} \colon \eucnorm{ x - p_i } \le \delta \} \cap
        ( \smallbigcup_{ j \in \{ 1, \ldots, \width \} \backslash \{ i \} } \hyper_{w_i,b_i} ) = \emptyset $.
  \end{enumerate}
  \Nobs that \cref{item:thm:bound:def:p4:1} \proves that for all $ i \in \{ 1, 2, \ldots, \width \} $ it holds that
  $ b_i + \scalprod{ w_i }{ p_i } = 0 $.
  Combining this with \cref{eq:thm:bound:vartheta4} and the Cauchy Schwarz inequality \proves that for all
  $ i \in \{ 1, 2, \ldots, \width \} $ it holds that
  \begin{equation}\label{eq:thm:bound:b4}
    \abs{ \vartheta_{ \inp \width + i } }
    = \tfrac{ \sqrt{L} }{ \eucnorm{ w_i } } \abs{ b_i }
    = \tfrac{ \sqrt{L} }{ \eucnorm{ w_i } } \abs{ \scalprod{ w_i }{ p_i } }
    \le \tfrac{ \sqrt{L} }{ \eucnorm{ w_i } } \eucnorm{ w_i } \eucnorm{ p_i }
    \le \sqrt{\inp L} \max\{ \abs{ \ms a }, \abs{ \ms b } \}.
  \end{equation}
  Next \nobs that the fact that for all $ i \in \{ 1, 2, \ldots, \width \} $,
  $ x \in \half_{w_i,b_i}^0 $, $ y \in \half_{w_i,b_i}^1 $ it holds that
  $ b_i + \scalprod{ w_i }{ x } \le 0 $ and $ b_i + \scalprod{ w_i }{ y } \ge 0 $ \proves that for all
  $ \ell_1, \ell_2, \ldots, \ell_{\width} \in \{ 0, 1 \} $, $ x = ( x_1, \ldots, x_{\inp} ) $,
  $ y = ( y_1, \ldots, y_{\inp} ) \in ( \smallbigcap_{i=1}^{\width} \half_{w_i,b_i}^{\ell_i} ) \cap [\ms a, \ms b]^{\inp} $
  it holds that
  \begin{equation}\label{eq:thm:bound:diff4}
  \begin{split}
    \mc N^{\theta} (x) - \mc N^{\theta} (y)
    &= \Bigl[ \theta_{\dimension} + \smallsum_{i=1}^{\width} \theta_{ \inp \width + \width + i } \max \bigl\{ \theta_{ \inp \width }
      + \smallsum_{j=1}^{\inp} \theta_{ (i-1) \inp + j } x_j, 0 \bigr\} \Bigr] \\
    &\qquad - \Bigl[ \theta_{\dimension} + \smallsum_{i=1}^{\width} \theta_{ \inp \width + \width + i } \max \bigl\{ \theta_{ \inp \width }
      + \smallsum_{j=1}^{\inp} \theta_{ (i-1) \inp + j } y_j, 0 \bigr\} \Bigr] \\
    &= \smallsum_{i=1}^{\width} v_i \max\{ b_i + \scalprod{ w_i }{ x }, 0 \}
      - \smallsum_{i=1}^{\width} v_i \max\{ b_i + \scalprod{ w_i }{ y }, 0 \} \\
    &= \smallsum_{i=1}^{\width} v_i \bigl( \max\{ b_i + \scalprod{ w_i }{ x }, 0 \} - \max\{ b_i + \scalprod{ w_i }{ y }, 0 \} \bigr) \\
    &= \smallsum_{ i \in \{ 1, 2, \ldots, \width \}, \, \ell_i = 1 } v_i \bigl( \bigr[ b_i + \scalprod{ w_i }{ x } \bigr]
      - \bigl[ b_i + \scalprod{ w_i }{ y } \bigr] \bigr) \\
    &= \smallsum_{ i \in \{ 1, 2, \ldots, \width \}, \, \ell_i = 1 } v_i \scalprod{ w_i }{ x - y }.
  \end{split}
  \end{equation}
  \Moreover
  the fact that $ [\ms a, \ms b]^{\inp} \subseteq \smallbigcup_{ \ell_1, \ell_2, \ldots, \ell_{\width} \in \{ 0, 1 \} }
    ( \bigcap_{i=1}^{\width} \half_{w_i,b_i}^{\ell_i} ) $
  and \cref{item:thm:bound:def:p4:1} \prove that for all $ i \in \{ 1, 2, \ldots, \width \} $ there exist
  $ \ell_1, \ell_2, \ldots, \ell_{\width} \in \{ 0, 1 \} $ such that
  $ p_i \in ( \bigcap_{j=1}^{\width} \half_{w_j,b_j}^{\ell_j} ) $ and $ \ell_i = 1 $.
  Combining this, \cref{item:thm:bound:def:p4:2,item:thm:bound:def:p4:3}, and \cref{eq:thm:bound:diff4} \proves that for all
  $ i \in \{ 1, 2, \ldots, \width \} $ there exist $ \ell_1, \ell_2, \ldots, \ell_{\width} \in \{ 0, 1 \} $ such that for all
  $ i \in \{ 1, 2, \ldots, \width \} $ it holds that
  \begin{equation}
  \begin{split}
    \mc N^{\theta} \bigl( p_i + \tfrac{ \delta }{ \eucnorm{ w_i } } w_i \bigr) - \mc N^{\theta} ( p_i )
    &= \smallsum_{ j \in \{ 1, 2, \ldots, \width \}, \, \ell_j = 1 } v_j \bscalprod{ w_j }{ \tfrac{ \delta }{ \eucnorm{ w_i } } w_i } \\
    &= \smallsum_{ j \in \{ 1, 2, \ldots, \width \} \backslash \{ i \}, \, \ell_j = 1 } v_j
      \bscalprod{ w_j }{ \tfrac{ \delta }{ \eucnorm{ w_i } } w_i }
      + v_i \tfrac{ \delta }{ \eucnorm{ w_i } } \scalprod{ w_i }{ w_i } \\
    &= \smallsum_{ j \in \{ 1, 2, \ldots, \width \} \backslash \{ i \}, \, \ell_j = 1 } v_j
      \bscalprod{ w_j }{ \tfrac{ \delta }{ \eucnorm{ w_i } } w_i } + \delta v_i \eucnorm{ w_i } \\
    &= \mc N^{\theta} ( p_i ) - \mc N^{\theta} \bigl( p_i - \tfrac{ \delta }{ \eucnorm{ w_i } } w_i \bigr)
      + \delta v_i \eucnorm{ w_i }.
  \end{split}
  \end{equation}
  This and \cref{eq:thm:bound:vartheta4} \prove that for all $ i \in \{ 1, 2, \ldots, \width \} $ it holds that
  \begin{equation}\label{eq:thm:bound:v4}
  \begin{split}
    \abs{ \vartheta_{ \inp \width + \width + i } }
    &= \tfrac{ \abs{ v_i } \eucnorm{ w_i } }{ \sqrt{L} }
    \le \tfrac{ 1 }{ \sqrt{L} } \Bigl[ \tfrac{ 1 }{ \delta } \babs{
      \mc N^{\theta} \bigl( p_i + \tfrac{ \delta }{ \eucnorm{ w_i } } w_i \bigr) - \mc N^{\theta} ( p_i ) }
      + \tfrac{ 1 }{ \delta } \babs{ \mc N^{\theta} ( p_i )
      - \mc N^{\theta} \bigl( p_i - \tfrac{ \delta }{ \eucnorm{ w_i } } w_i \bigr) } \Bigr] \\
    &\le \tfrac{ 2 }{ \sqrt{L} } \Bigl( \sup\nolimits_{ x, y \in [\ms a, \ms b]^{\inp}, \, x \neq y }
      \tfrac{ \abs{ \mc N^{\theta} (x) - \mc N^{\theta} (y) } }{ \eucnorm{ x - y } } \Bigr)
    = 2 \sqrt{L}.
  \end{split}
  \end{equation}
  The fact that for all $ i \in \{ 1, 2, \ldots, \width \} $ it holds that $ b_i + \scalprod{ w_i }{ p_i } = 0 $,
  the fact that for all $ x \in \R $ it holds that $ \abs{ \max\{ x, 0 \} } \le \abs{ x } $, and
  the Cauchy Schwarz inequality \hence \prove that for all $ i \in \{ 1, 2, \ldots, \width \} $ it holds that
  \begin{equation}
  \begin{split}
    \abs{ v_i \max\{ b_i + \scalprod{ w_i }{ z } \} }
    &= \abs{ v_i \max\{ \scalprod{ w_i }{ z - p_i } \} }
    \le \abs{ v_i } \abs{ \scalprod{ w_i }{ z - p_i } }
    \le \abs{ v_i } \eucnorm{ w_i } \eucnorm{ z - p_i } \\
    &\le \tfrac{ \sqrt{L} }{ \eucnorm{ w_i } } \abs{ \vartheta_{ \inp \width + \width + i } } \eucnorm{ w_i } ( \ms b - \ms a ) \sqrt{\inp}
    \le 2 L ( \ms b - \ms a ) \sqrt{\inp}.
  \end{split}
  \end{equation}
  This, \cref{eq:thm:bound:def:wbv}, \cref{eq:thm:bound:def:inf}, and \cref{eq:thm:bound:vartheta4} \prove that
  \begin{equation}\label{eq:thm:bound:c4}
  \begin{split}
    \abs{ \vartheta_{\dimension} }
    &= \abs{ \theta_{\dimension} }
    = \abs{ \mc N^{\theta} (z) - \smallsum_{i=1}^{\width} \theta_{ \inp \width + \width + i } \max \bigl\{ \theta_{ \inp \width }
      + \smallsum_{j=1}^{\inp} \theta_{ (i-1) \inp + j } z_j, 0 \bigr\} } \\
    &= \abs{ \mc N^{\theta} (z) - \smallsum_{i=1}^{\width} v_i \max \bigl\{ b_i + \scalprod{ w_i }{ z }, 0 \} } \\
    &\le \abs{ \mc N^{\theta} (z) } + \smallsum_{i=1}^{\width} \abs{ v_i \max\{ b_i + \scalprod{ w_i }{ z }, 0 \} } \\
    &\le \bigl( \inf\nolimits_{ x \in [\ms a, \ms b]^{\inp} } \abs{ \mc N^{\theta} (x) } \bigr) + 2 \width L ( \ms b - \ms a ) \sqrt{\inp}.
  \end{split}
  \end{equation}
  Combining this, \cref{eq:thm:bound:w4}, \cref{eq:thm:bound:b4}, and \cref{eq:thm:bound:v4} \proves that
  \begin{equation}
    \max\nolimits_{ i \in \{ 1, 2, \ldots, \dimension \} } \abs{ \vartheta_i }
    \le \max \bigl\{ \max\{ 2, \abs{ \ms a } \sqrt{\inp}, \abs{ \ms b } \sqrt{\inp} \} \sqrt{L},
      \bigl[ \inf\nolimits_{ x \in [\ms a, \ms b]^{\inp} } \abs{ \mc N^{\theta} (x) } \bigr]
      + 2 \width L ( \ms b - \ms a ) \sqrt{\inp} \bigr\}.
  \end{equation}
  This and \cref{eq:thm:bound:realization4} \prove \cref{eq:thm:bound} in the case $ [ L \in (0,\infty) ] \wedge [ N = \width ] $.
\end{cproof}

\cfclear
\begin{definition}\label{def:lipschitz_norm}
  Let $ d \in \N $, $ \ms a \in \R $, $ \ms b \in (\ms a, \infty) $, $ A \subseteq [\ms a, \ms b]^d $ satisfy $ A \neq \emptyset $ and
  let $ f \colon [\ms a, \ms b]^{\inp} \to \R $ be a function.
  Then we denote by $ \lipnorm{f}{A} \in [0,\infty] $ the extended real number given by
  \begin{equation}
    \lipnorm{f}{A} = \inf\nolimits_{ x \in A } \abs{ f(x) }
      + \sup\nolimits_{ x, y \in [\ms a, \ms b]^d, \, x \neq y } \tfrac{ \abs{ f(x) - f(y) } }{ \eucnorm{ x - y } }
  \end{equation}
  (cf.~\cref{def:scalar_product_norm}).
\end{definition}

\cfclear
\begin{cor}\label{cor:bound:explicit}
  Assume \cref{setting:main} and let $ \theta \in \R^{\dimension} $, $ A \subseteq [\ms a, \ms b]^{\inp} $ satisfy $ A \neq \emptyset $.
  Then there exists $ \vartheta = ( \vartheta_1, \ldots, \vartheta_{\dimension} ) \in \R^{\dimension} $ such that
  $ \mc N^{\vartheta} = \mc N^{\theta} $ and
  \begin{equation}\label{eq:cor:bound:explicit}
    \max\nolimits_{ i \in \{ 1, 2, \ldots, \dimension \} } \abs{ \vartheta_i }
      \le \max \bigl\{ 2, \abs{ \ms a } \sqrt{\inp}, \abs{ \ms b } \sqrt{\inp}, 2 \width ( \ms b - \ms a ) \sqrt{\inp} \bigr\}
        \max\{ \lipnorm{ \mc N^{\theta} }{ A }^{\nicefrac{1}{2}} , \lipnorm{ \mc N^{\theta} }{ A } \}.
  \end{equation}
  \cfout.
\end{cor}

\begin{cproof}{cor:bound:explicit}
  Throughout this proof let $ L \in [0,\infty) $ satisfy 
  \begin{equation}
    L = \sup\nolimits_{ x, y \in [\ms a, \ms b]^{\inp}, \, x \neq y }
      \tfrac{ \abs{ \mc N^{\theta} (x) - \mc N^{\theta} (y) } }{ \eucnorm{ x - y } }
  \end{equation}
  \cfload.
  \Nobs that \cref{thm:bound} \proves that there exists
  $ \vartheta = ( \vartheta_1, \ldots, \allowbreak \vartheta_{\dimension} ) \in \R^{\dimension} $ which satisfies that
  $ \mc N^{\vartheta} = \mc N^{\theta} $ and
  \begin{equation}\label{cor:bound:explicit:vartheta}
    \max\nolimits_{ i \in \{ 1, 2, \ldots, \dimension \} } \abs{ \vartheta_i }
    \le \max \bigl\{ \max\{ 2, \abs{ \ms a } \sqrt{\inp}, \abs{ \ms b } \sqrt{\inp} \} \sqrt{L},
      \bigl[ \inf\nolimits_{ x \in [\ms a, \ms b]^{\inp} } \abs{ \mc N^{\theta} (x) } \bigr]
      + 2 \width L ( \ms b - \ms a ) \sqrt{\inp} \bigr\}.
  \end{equation}
  \Moreover the fact that $ L \le \lipnorm{ \mc N^{\theta} }{ A } $ \proves that
  \begin{equation}\label{cor:bound:explicit:1}
  \begin{split}
    \max \bigl\{ 2, \abs{ \ms a } \sqrt{\inp}, \abs{ \ms b } \sqrt{\inp} \bigr\} \sqrt{L}
    &\le \max \bigl\{ 2, \abs{ \ms a } \sqrt{d}, \abs{ \ms b } \sqrt{d} \bigr\} \lipnorm{ \mc N^{\theta} }{ A }^{\nicefrac{1}{2}} \\
    &\le \max \bigl\{ 2, \abs{ \ms a } \sqrt{\inp}, \abs{ \ms b } \sqrt{\inp}, 2 \width ( \ms b - \ms a ) \sqrt{\inp} \bigr\}
      \lipnorm{ \mc N^{\theta} }{ A }^{\nicefrac{1}{2}}
  \end{split}
  \end{equation}
  \cfload.
  \Moreover the fact that
  $ \inf_{ x \in [\ms a, \ms b]^{\inp} } \abs{ \mc N^{\theta} (x) } \le \inf_{ x \in A } \abs{ \mc N^{\theta} (x) } $
  \proves that
  \begin{equation}\label{cor:bound:explicit:2}
  \begin{split}
    \bigl[ \inf\nolimits_{ x \in [\ms a, \ms b]^{\inp} } \abs{ \mc N^{\theta} (x) } \bigr] + 2 \width L ( \ms b - \ms a ) \sqrt{\inp}
    &\le \inf\nolimits_{ x \in A } \abs{ \mc N^{\theta} (x) } + 2 \width L ( \ms b - \ms a ) \sqrt{\inp} \\
    &\le \max \bigl\{ 1, 2 \width ( \ms b - \ms a ) \sqrt{\inp} \bigr\}
      \bigl[ \inf\nolimits_{ x \in A } \abs{ \mc N^{\theta} (x) } + L \bigr] \\
    &= \max\{ 1, 2 \width ( \ms b - \ms a ) \sqrt{\inp} \} \lipnorm{ \mc N^{\theta} }{ A } \\
    &\le \max \bigl\{ 2, \abs{ \ms a } \sqrt{\inp}, \abs{ \ms b } \sqrt{\inp}, 2 \width ( \ms b - \ms a ) \sqrt{\inp} \bigr\}
      \lipnorm{ \mc N^{\theta} }{ A }.
  \end{split}
  \end{equation}
  Combining this with \cref{cor:bound:explicit:vartheta} and \cref{cor:bound:explicit:1} \proves that
  \begin{equation}
  \begin{split}
    \max\nolimits_{ i \in \{ 1, 2, \ldots, \dimension \} } \abs{ \vartheta_i }
    &\le \max \bigl\{ \max\{ 2, \abs{ \ms a } \sqrt{\inp}, \abs{ \ms b } \sqrt{\inp} \} \sqrt{L},
      \bigl[ \inf\nolimits_{ x \in [\ms a, \ms b]^{\inp} } \abs{ \mc N^{\theta} (x) } \bigr]
      + 2 \width L ( \ms b - \ms a ) \sqrt{\inp} \bigr\} \\
    &\le \max \bigl\{ 2, \abs{ \ms a } \sqrt{\inp}, \abs{ \ms b } \sqrt{\inp}, 2 \width ( \ms b - \ms a ) \sqrt{\inp} \bigr\}
        \max\{ \lipnorm{ \mc N^{\theta} }{ A }^{\nicefrac{1}{2}} , \lipnorm{ \mc N^{\theta} }{ A } \}.
  \end{split}
  \end{equation}
\end{cproof}

\cfclear
\begin{cor}\label{cor:lipnorm:range:positive}
  Assume \cref{setting:main} and let $ n \in \N $, $ \delta_1, \delta_2, \ldots, \delta_n \in [0,\infty) $,
  $ A \subseteq [\ms a, \ms b]^{\inp} $ satisfy $ \min\{ \delta_1, \delta_2, \ldots, \delta_n \} \le \nicefrac{1}{2} $,
  $ \max\{ \delta_1, \delta_2, \ldots, \delta_n \} \ge 1 $, and $ A \neq \emptyset $.
  Then there exists $ \ms c \in \R $ such that for all $ \theta \in \R^{\dimension} $ there exists $ \vartheta \in \R^{\dimension} $ such that
  $ \mc N^{\vartheta} = \mc N^{\theta} $ and
  \begin{equation}
    \eucnorm{ \vartheta } \le \ms c \Bigl( \smallsum\nolimits_{i=1}^n \lipnorm{ \mc N^{\theta} }{ A }^{ \delta_i } \Bigr)
  \end{equation}
  \cfout.
\end{cor}

\begin{cproof}{cor:lipnorm:range:positive}
  \Nobs that the assumption that $ \min\{ \delta_1, \delta_2, \ldots, \delta_n \} \le \nicefrac{1}{2} $ and the assumption that
  $ \max\{ \delta_1, \delta_2, \allowbreak \ldots, \delta_n \} \ge 1 $ \prove that there exist $ i, j \in \{ 1, 2, \ldots, n \} $
  which satisfy that
  \begin{equation}\label{eq:cor:lipnorm:range:positive:ij}
    \delta_i \le \nicefrac{1}{2}
    \qqandqq
    \delta_j \ge 1.
  \end{equation}
  \Nobs that \cref{cor:bound:explicit} and \cref{eq:cor:lipnorm:range:positive:ij} \prove that there exists $ \ms c \in \R $ such that for all
  $ \theta \in \R^{\dimension} $ there exists $ \vartheta = ( \vartheta_1, \ldots, \vartheta_{\dimension} ) \in \R^{\dimension} $ such that
  $ \mc N^{\theta} = \mc N^{\vartheta} $ and
  \begin{equation}
  \begin{split}
    \eucnorm{ \vartheta }
    &\le \sqrt{ \dimension } \max\nolimits_{ i \in \{ 1, 2, \ldots, \dimension \} } \abs{ \vartheta_i }
    \le \sqrt{ \dimension } \ms c \max \bigl\{ \lipnorm{ \mc N^{\theta} }{ A }^{\nicefrac{1}{2}}, \lipnorm{ \mc N^{\theta} }{ A } \bigr\} \\
    &\le \sqrt{ \dimension } \ms c \max \bigl\{ \lipnorm{ \mc N^{\theta} }{ A }^{\delta_i}, \lipnorm{ \mc N^{\theta} }{ A }^{\delta_j} \bigr\}
    \le 2 \sqrt{ \dimension } \ms c \Bigl( \lipnorm{ \mc N^{\theta} }{ A }^{\delta_i} + \lipnorm{ \mc N^{\theta} }{ A }^{\delta_j} \Bigr) \\
    &\le 2 \sqrt{ \dimension } \ms c \Bigl( \smallsum\nolimits_{k=1}^n \lipnorm{ \mc N^{\theta} }{ A }^{\delta_k} \Bigr)
  \end{split}
  \end{equation}
  \cfload.
\end{cproof}

\subsection{Equivalence of norms of reparameterized ANNs and Lipschitz norms}
\label{subsec:equivalence}

\cfclear
\begin{lemma}\label{lem:lipschitzconstant:bound}
  Assume \cref{setting:main} and let $ \theta = ( \theta_1, \ldots, \theta_{\dimension} ) \in \R^{\dimension} $,
  $ w \in \R^{\inp \width} $, $ v \in \R^{\width} $ satisfy
  $ w = ( \theta_1, \ldots, \theta_{\inp \width} ) $ and
  $ v = ( \theta_{\inp \width + \width + 1}, \ldots, \theta_{\inp \width + 2 \width} ) $.
  Then
  \begin{equation}\label{eq:lem:lipschitzconstant:bound}
    \sup\nolimits_{ x, y \in [\ms a, \ms b]^{\inp}, \, x \neq y } \tfrac{ \abs{ \mc N^{\theta}(x) - \mc N^{\theta}(y) } }
      { \eucnorm{ x - y } }
    \le \eucnorm{ v } \eucnorm{ w }
    \le \tfrac{1}{2} \eucnorm{ \theta }^2.
  \end{equation}
  \cfout.
\end{lemma}

\begin{cproof}{lem:lipschitzconstant:bound}
  \Nobs that the fact that for all $ x, y \in \R $ it holds that $ \abs{ \max\{ x, 0 \} - \max\{ y, 0 \} } \le \abs{ x - y } $ \proves
  that for all $ x = ( x_1, \ldots, x_{\inp} ) $, $ y = ( y_1, \ldots, y_{\inp} ) \in [\ms a, \ms b]^{\inp} $ it holds that
  \begin{equation}\label{eq:lem:lipschitzconstant:bound:difference}
  \begin{split}
    \abs{ \mc N^{\theta} ( x ) - \mc N^{\theta} ( y ) }
    &= \babs{ \theta_{\dimension} + \smallsum_{i=1}^{\width} \theta_{ \inp \width + \width + i } \max\{ \theta_{ \inp \width + i }
      + \smallsum_{j=1}^{\inp} \theta_{ (i-1) \inp + j } x_j, 0 \} \\
    &\quad - \bigl[ \theta_{\dimension} + \smallsum_{i=1}^{\width} \theta_{ \inp \width + \width + i } \max\{ \theta_{ \inp \width + i }
      + \smallsum_{j=1}^{\inp} \theta_{ (i-1) \inp + j } y_j, 0 \} \bigr] } \\
    &= \babs{ \smallsum_{i=1}^{\width} \theta_{ \inp \width + \width + i }
      \bigl( \max\{ \theta_{ \inp \width + i } + \smallsum_{j=1}^{\inp} \theta_{ (i-1) \inp + j } x_j, 0 \} \\
    &\quad - \max\{ \theta_{ \inp \width + i } + \smallsum_{j=1}^{\inp} \theta_{ (i-1) \inp + j } y_j, 0 \} \bigr) } \\
    &\le \smallsum_{i=1}^{\width} \abs{ \theta_{ \inp \width + \width + i } }
      \babs{ \max\{ \theta_{ \inp \width + i } + \smallsum_{j=1}^{\inp} \theta_{ (i-1) \inp + j } x_j, 0 \} \\
    &\quad - \max\{ \theta_{ \inp \width + i }
      + \smallsum_{j=1}^{\inp} \theta_{ (i-1) \inp + j } y_j, 0 \} } \\
    &\le \smallsum_{i=1}^{\width} \abs{ \theta_{ \inp \width + \width + i } }
      \babs{ \bigl[ \theta_{ \inp \width + i } + \smallsum_{j=1}^{\inp} \theta_{ (i-1) \inp + j } x_j \bigr] \\
    &\quad - \bigl[ \theta_{ \inp \width + i } + \smallsum_{j=1}^{\inp} \theta_{ (i-1) \inp + j } y_j \bigr] } \\
    &= \smallsum_{i=1}^{\width} \abs{ \theta_{ \inp \width + \width + i } }
      \abs{ \smallsum_{j=1}^{\inp} \theta_{ (i-1) \inp + j } ( x_j - y_j ) }.
  \end{split}
  \end{equation}
  \Moreover the Cauchy Schwarz inequality \proves that for all
  $ x = ( x_1, \ldots, \allowbreak x_{\inp} ) $, $ y = ( y_1, \ldots, y_{\inp} ) \in [\ms a, \ms b]^{\inp} $ it holds that
  \begin{equation}
  \begin{split}
    &\smallsum_{i=1}^{\width} \abs{ \theta_{ \inp \width + \width + i } }
      \abs{ \smallsum_{j=1}^{\inp} \theta_{ (i-1) \inp + j } ( x_j - y_j ) } \\
    &\le \eucnorm{ x - y } \smallsum_{i=1}^{\width} \abs{ \theta_{ \inp \width + \width + i } }
      \Bigl[ \smallsum_{j=1}^{\inp} \abs{ \theta_{ (i-1) \inp + j } }^2 \Bigr]^{\nicefrac{1}{2}} \\
    &\le \eucnorm{ x - y }
      \Bigl[ \smallsum_{i=1}^{\width} \abs{ \theta_{ \inp \width + \width + i } }^2 \Bigr]^{\nicefrac{1}{2}}
      \Bigl[ \smallsum_{i=1}^{\width} \smallsum_{j=1}^{\inp} \abs{ \theta_{ (i-1) \inp + j } }^2 \Bigr]^{\nicefrac{1}{2}}
    = \eucnorm{ x - y } \eucnorm{ v } \eucnorm{ w }
  \end{split}
  \end{equation}
  \cfload.
  Combining this with \cref{eq:lem:lipschitzconstant:bound:difference} \proves that for all
  $ x = ( x_1, \ldots, x_{\inp} ) $, $ y = ( y_1, \ldots, y_{\inp} ) \in [\ms a, \ms b]^{\inp} $ it holds that
  \begin{equation}
    \abs{ \mc N^{\theta} ( x ) - \mc N^{\theta} ( y ) }
    \le \smallsum_{i=1}^{\width} \abs{ \theta_{ \inp \width + \width + i } }
      \abs{ \smallsum_{j=1}^{\inp} \theta_{ (i-1) \inp + j } ( x_j - y_j ) }
    \le \eucnorm{ x - y } \eucnorm{ v } \eucnorm{ w }.
  \end{equation}
  The fact that for all $ x, y \in \R $ it holds that $ 2 x y \le x^2 + y^2 $ \hence \proves that
  \begin{equation}
    \sup\nolimits_{ x, y \in [\ms a, \ms b]^{\inp}, \, x \neq y } \tfrac{ \abs{ \mc N^{\theta}(x) - \mc N^{\theta}(y) } }
      { \eucnorm{ x - y } }
    \le \eucnorm{ v } \eucnorm{ w }
    \le \tfrac{1}{2} \bigl( \eucnorm{ v }^2 + \eucnorm{ w }^2 \bigr)
    \le \tfrac{1}{2} \eucnorm{ \theta }^2.
  \end{equation}
\end{cproof}

\cfclear
\begin{lemma}\label{lem:inverse:bound}
  Assume \cref{setting:main}, let $ A \subseteq [\ms a, \ms b]^{\inp } $ satisfy $ A \neq \emptyset $, and
  let $ \theta = ( \theta_1, \ldots, \theta_{\dimension} ) \in \R^{\dimension} $,
  $ w \in \R^{\inp \width} $, $ b, v \in \R^{\width} $ satisfy
  $ w = ( \theta_1, \ldots, \theta_{\inp \width} ) $, $ b = ( \theta_{\inp \width + 1}, \ldots, \theta_{\inp \width + \width} ) $, and
  $ v = ( \theta_{\inp \width + \width + 1}, \ldots, \allowbreak \theta_{\inp \width + 2 \width} ) $.
  Then
  \begin{equation}\label{eq:lem:inverse:bound}
    \lipnorm{ \mc N^{\theta} }{ A }
    \le \abs{ \theta_{\dimension} } + \eucnorm{ v } \Bigl[ \eucnorm{ b }
      + \bigl( 1 + \inf\nolimits_{ x \in A } \eucnorm{ x } \bigr) \eucnorm{ w } \Bigr]
    \le \eucnorm{ \theta } + \bigl( 1 + \max\{ \abs{ \ms a }, \abs{ \ms b } \} \nicefrac{\sqrt{\inp}}{2} \bigr) \eucnorm{ \theta }^2
  \end{equation}
  \cfout.
\end{lemma}

\begin{cproof}{lem:inverse:bound}
  \Nobs that the fact that for all $ x \in \R $ it holds that $ \abs{ \max \{ x, 0 \} } \le \abs{ x } $ \proves that
  for all $ y = ( y_1, \ldots, y_{\inp} ) \in A $ it holds that
  \begin{equation}\label{eq:lem:inverse:bound:1}
  \begin{split}
    \inf\nolimits_{ x \in A } \abs{ \mc N^{\theta} (x) }
    &\le \abs{ \mc N^{\theta} ( y ) }
    = \babs{ \theta_{\dimension} + \smallsum_{i=1}^{\width} \theta_{ \inp \width + \width + i } \max \{ \theta_{ \inp \width + i }
      + \smallsum_{j=1}^{\inp} \theta_{ (i-1) \inp + j } y_j, 0 \} } \\
    &\le \abs{ \theta_{\dimension} } + \smallsum_{i=1}^{\width} \babs{ \theta_{ \inp \width + \width + i } \max \{ \theta_{ \inp \width + i }
      + \smallsum_{j=1}^{\inp} \theta_{ (i-1) \inp + j } y_j, 0 \} } \\
    &\le \abs{ \theta_{\dimension} } + \smallsum_{i=1}^{\width} \abs{ \theta_{ \inp \width + \width + i } }
      \babs{ \theta_{ \inp \width + i } + \smallsum_{j=1}^{\inp} \theta_{ (i-1) \inp + j } y_j } \\
    &\le \abs{ \theta_{\dimension} } + \smallsum_{i=1}^{\width} \abs{ \theta_{ \inp \width + \width + i } \theta_{ \inp \width + i } }
      + \smallsum_{i=1}^{\width} \abs{ \theta_{ \inp \width + \width + i } } \smallsum_{j=1}^{\inp} \abs{ \theta_{ (i-1) \inp + j } y_j }.
  \end{split}
  \end{equation}
  \Moreover the Cauchy Schwarz inequality \proves that for all $ y = ( y_1, \ldots, y_{\inp} ) \in A $ it holds that
  \begin{equation}\label{eq:lem:inverse:bound:2}
  \begin{split}
    &\smallsum_{i=1}^{\width} \abs{ \theta_{ \inp \width + \width + i } \theta_{ \inp \width + i } }
      + \smallsum_{i=1}^{\width} \abs{ \theta_{ \inp \width + \width + i } } \smallsum_{j=1}^{\inp} \abs{ \theta_{ (i-1) \inp + j } y_j } \\
    &\le \Bigl[ \smallsum_{i=1}^{\width} \abs{ \theta_{ \inp \width + \width + i } }^2 \Bigr]^{\nicefrac{1}{2}}
      \Bigl[ \smallsum_{i=1}^{\width} \abs{ \theta_{ \inp \width + i } }^2 \Bigr]^{\nicefrac{1}{2}}
      + \eucnorm{ y } \smallsum_{i=1}^{\width} \abs{ \theta_{ \inp \width + \width + i } }
      \Bigl[ \smallsum_{j=1}^{\inp} \abs{ \theta_{ (i-1) \inp + j } }^2 \Bigr]^{\nicefrac{1}{2}} \\
    &\le \eucnorm{ v } \eucnorm{ b } + \eucnorm{ y }
      \Bigl[ \smallsum_{i=1}^{\width} \abs{ \theta_{ \inp \width + \width + i } }^2 \Bigr]^{\nicefrac{1}{2}}
      \Bigl[ \smallsum_{i=1}^{\width} \smallsum_{j=1}^{\inp} \abs{ \theta_{ (i-1) \inp + j } }^2 \Bigr]^{\nicefrac{1}{2}} \\
    &= \eucnorm{ v } \eucnorm{ b } + \eucnorm{ y } \eucnorm{ v } \eucnorm{ w }
    = \eucnorm{ v } \bigl( \eucnorm{ b } + \eucnorm{ y } \eucnorm{ w } \bigr)
  \end{split}
  \end{equation}
  \cfload.
  Combining this with \cref{eq:lem:inverse:bound:1} \proves that for all $ y = ( y_1, \ldots, y_{\inp} ) \in A $ it holds that
  \begin{equation}\label{eq:lem:inverse:bound:3}
  \begin{split}
    \inf\nolimits_{ x \in A } \abs{ \mc N^{\theta} (x) }
    &\le \abs{ \theta_{\dimension} } + \smallsum_{i=1}^{\width} \abs{ \theta_{ \inp \width + \width + i } \theta_{ \inp \width + i } }
      + \smallsum_{i=1}^{\width} \abs{ \theta_{ \inp \width + \width + i } } \smallsum_{j=1}^{\inp} \abs{ \theta_{ (i-1) \inp + j } y_j } \\
    &\le \abs{ \theta_{\dimension} } + \eucnorm{ v } \bigl( \eucnorm{ b } + \eucnorm{ y } \eucnorm{ w } \bigr).
  \end{split}
  \end{equation}
  \Hence that
  \begin{equation}\label{eq:lem:inverse:bound:4}
    \inf\nolimits_{ x \in A } \abs{ \mc N^{\theta} (x) }
    \le \abs{ \theta_{\dimension} } + \eucnorm{ v } \Bigl[ \eucnorm{ b }
      + \bigl( \inf\nolimits_{ x \in A } \eucnorm{ x } \bigr) \eucnorm{ w } \Bigr].
  \end{equation}
  Combining this and \cref{lem:lipschitzconstant:bound} \proves that
  \begin{equation}
  \begin{split}
    \lipnorm{ \mc N^{\theta} }{ A }
    &= \inf\nolimits_{ x \in A } \abs{ \mc N^{\theta} (x) }
      + \sup\nolimits_{ x, y \in [\ms a, \ms b]^d, \, x \neq y }
      \tfrac{ \abs{ \mc N^{\theta} (x) - \mc N^{\theta} (y) } }{ \eucnorm{ x - y } } \\
    &\le \abs{ \theta_{\dimension} } + \eucnorm{ v } \Bigl[ \eucnorm{ b } + \bigl( \inf\nolimits_{ x \in A } \eucnorm{ x } \bigr) \eucnorm{ w } \Bigr]
      + \eucnorm{ v } \eucnorm{ w } \\
    &\le \abs{ \theta_{\dimension} } + \eucnorm{ v } \Bigl[ \eucnorm{ b }
      + \bigl( 1 + \inf\nolimits_{ x \in A } \eucnorm{ x } \bigr) \eucnorm{ w } \Bigr]
  \end{split}
  \end{equation}
  \cfload.
  \Moreover the fact that for all $ x \in A $ it holds that $ \eucnorm{ x } \le \max\{ \abs{ \ms a }, \abs{ \ms b } \} \sqrt{\inp} $ and
  the fact that for all $ x, y \in \R $ it holds that $ 2 x y \le x^2 + y^2 $ \prove that
  \begin{equation}
  \begin{split}
    &\abs{ \theta_{\dimension} } + \eucnorm{ v } \Bigl[ \eucnorm{ b }
      + \bigl( 1 + \inf\nolimits_{ x \in A } \eucnorm{ x } \bigr) \eucnorm{ w } \Bigr] \\
    &\le \eucnorm{ \theta } + \eucnorm{ v } \eucnorm{ b }
      + \bigl( 1 + \max\{ \abs{ \ms a }, \abs{ \ms b } \} \sqrt{\inp} \bigr) \eucnorm{ v } \eucnorm{ w } \\
    &\le \eucnorm{ \theta } + \tfrac{1}{2} \bigl( \eucnorm{ v }^2 + \eucnorm{ b }^2 \bigr)
      + \tfrac{1}{2} \bigl( 1 + \max\{ \abs{ \ms a }, \abs{ \ms b } \} \sqrt{\inp} \bigr) \bigl( \eucnorm{ v }^2 + \eucnorm{ w }^2 \bigr) \\
    &\le \eucnorm{ \theta } + \tfrac{1}{2} \eucnorm{ \theta }^2
      + \tfrac{1}{2} \bigl( 1 + \max\{ \abs{ \ms a }, \abs{ \ms b } \} \sqrt{\inp} \bigr) \eucnorm{ \theta }^2 \\
    &= \eucnorm{ \theta } + \bigl( 1 + \max\{ \abs{ \ms a }, \abs{ \ms b } \} \nicefrac{\sqrt{\inp}}{2} \bigr) \eucnorm{ \theta }^2.
  \end{split}
  \end{equation}
\end{cproof}

\cfclear
\begin{cor}\label{cor:bound:combined}
  Assume \cref{setting:main} and let $ A \subseteq [\ms a, \ms b]^{\inp} $ satisfy $ A \neq \emptyset $.
  Then for all $ \theta \in \R^{\dimension} $ there exists $ \vartheta \in \R^{\dimension} $ such that $ \mc N^{\vartheta} = \mc N^{\theta} $
  and
  \begin{equation}\label{eq:cor:bound:combined}
  \begin{split}
    \eucnorm{ \vartheta }
    &\le \sqrt{\dimension} \max \bigl\{ 2, \abs{ \ms a } \sqrt{\inp}, \abs{ \ms b } \sqrt{\inp}, 2 \width ( \ms b - \ms a ) \sqrt{\inp} \bigr\}
       \max\{ \lipnorm{ \mc N^{\theta} }{ A }^{\nicefrac{1}{2}} , \lipnorm{ \mc N^{\theta} }{ A } \} \\
    &\le 2 \sqrt{\dimension} \bigl[ \max \bigl\{ 2, \abs{ \ms a } \sqrt{\inp}, \abs{ \ms b } \sqrt{\inp},
      2 \width ( \ms b - \ms a ) \sqrt{\inp} \bigr\} \bigr]^2 \max\{ \eucnorm{ \vartheta }^{\nicefrac{1}{2}}, \eucnorm{ \vartheta }^2 \}
  \end{split}
  \end{equation}
  \cfout.
\end{cor}

\begin{cproof}{cor:bound:combined}
  \Nobs that \cref{lem:inverse:bound} \proves that for all $ \theta \in \R^{\dimension} $ it holds that
  \begin{equation}
  \begin{split}
    \lipnorm{ \mc N^{\theta} }{ A }
    &\le \eucnorm{ \theta } + \bigl( 1 + \max\{ \abs{ \ms a }, \abs{ \ms b } \} \nicefrac{\sqrt{\inp}}{2} \bigr) \eucnorm{ \theta }^2 \\
    &\le \bigl( 1 + \max\{ \abs{ \ms a }, \abs{ \ms b } \} \nicefrac{\sqrt{\inp}}{2} \bigr)
      \bigl( \eucnorm{ \theta } + \eucnorm{ \theta }^2 \bigr) \\
    &\le \bigl( 2 + \max\{ \abs{ \ms a }, \abs{ \ms b } \} \sqrt{\inp} \bigr) \max\{ \eucnorm{ \theta }, \eucnorm{ \theta }^2 \}
  \end{split}
  \end{equation}
  and
  \begin{equation}
    \lipnorm{ \mc N^{\theta} }{ A }^{\nicefrac{1}{2}}
    \le \Bigl( 2 + \max\{ \abs{ \ms a }, \abs{ \ms b } \} \sqrt{\inp} \Bigr)^{\nicefrac{1}{2}}
      \max\{ \eucnorm{ \theta }^{\nicefrac{1}{2}}, \eucnorm{ \theta } \}
  \end{equation}
  \cfload.
  \Hence that for all $ \theta \in \R^{\dimension} $ it holds that
  \begin{equation}\label{eq:cor:bound:combined:upper}
  \begin{split}
    \max\{ \lipnorm{ \mc N^{\theta} }{ A }^{\nicefrac{1}{2}} , \lipnorm{ \mc N^{\theta} }{ A } \}
    &\le \bigl( 2 + \max\{ \abs{ \ms a }, \abs{ \ms b } \} \sqrt{\inp} \bigr)
      \max\{ \eucnorm{ \theta }^{\nicefrac{1}{2}}, \eucnorm{ \theta }, \eucnorm{ \theta }^2 \} \\
    &\le 2 \max\{ 2, \abs{ \ms a } \sqrt{\inp}, \abs{ \ms b } \sqrt{\inp} \}
      \max\{ \eucnorm{ \theta }^{\nicefrac{1}{2}}, \eucnorm{ \theta }^2 \} \\
    &\le 2 \max\{ 2, \abs{ \ms a } \sqrt{\inp}, \abs{ \ms b } \sqrt{\inp}, 2 \width ( \ms b - \ms a ) \sqrt{\inp} \}
      \max\{ \eucnorm{ \theta }^{\nicefrac{1}{2}}, \eucnorm{ \theta }^2 \}.
  \end{split}
  \end{equation}
  \Moreover \cref{cor:bound:explicit} \proves that for all $ \theta \in \R^{\dimension} $ there exists
  $ \vartheta = ( \vartheta_1, \ldots, \vartheta_{\dimension} ) \in \R^{\dimension} $ such that $ \mc N^{\vartheta} = \mc N^{\theta} $ and
  \begin{equation}
    \sqrt{\dimension} \max\nolimits_{ i \in \{ 1, 2, \ldots, \dimension \} } \abs{ \vartheta_i }
    \le \sqrt{\dimension} \max \bigl\{ 2, \abs{ \ms a } \sqrt{\inp}, \abs{ \ms b } \sqrt{\inp}, 2 \width ( \ms b - \ms a ) \sqrt{\inp} \bigr\}
      \max\{ \lipnorm{ \mc N^{\theta} }{ A }^{\nicefrac{1}{2}} , \lipnorm{ \mc N^{\theta} }{ A } \}.
  \end{equation}
  Combining this with \cref{eq:cor:bound:combined:upper} \proves that for all $ \theta \in \R^{\dimension} $ there exists
  $ \vartheta = ( \vartheta_1, \ldots, \vartheta_{\dimension} ) \in \R^{\dimension} $ such that $ \mc N^{\vartheta} = \mc N^{\theta} $ and
  \begin{equation}
  \begin{split}
    \eucnorm{ \vartheta }
    &\le \sqrt{\dimension} \max\nolimits_{ i \in \{ 1, 2, \ldots, \dimension \} } \abs{ \vartheta_i } \\
    &\le \sqrt{\dimension} \max \bigl\{ 2, \abs{ \ms a } \sqrt{\inp}, \abs{ \ms b } \sqrt{\inp}, 2 \width ( \ms b - \ms a ) \sqrt{\inp} \bigr\}
      \max\{ \lipnorm{ \mc N^{\theta} }{ A }^{\nicefrac{1}{2}} , \lipnorm{ \mc N^{\theta} }{ A } \} \\
    &= \sqrt{\dimension} \max \bigl\{ 2, \abs{ \ms a } \sqrt{\inp}, \abs{ \ms b } \sqrt{\inp}, 2 \width ( \ms b - \ms a ) \sqrt{\inp} \bigr\}
      \max\{ \lipnorm{ \mc N^{\vartheta} }{ A }^{\nicefrac{1}{2}} , \lipnorm{ \mc N^{\vartheta} }{ A } \} \\
    &\le 2 \sqrt{\dimension} \bigl[ \max \bigl\{ 2, \abs{ \ms a } \sqrt{\inp}, \abs{ \ms b } \sqrt{\inp},
      2 \width ( \ms b - \ms a ) \sqrt{\inp} \bigr\} \bigr]^2 \max\{ \eucnorm{ \vartheta }^{\nicefrac{1}{2}}, \eucnorm{ \vartheta }^2 \}.
  \end{split}
  \end{equation}
\end{cproof}

\cfclear
\begin{cor}\label{cor:bound:weak}
  Assume \cref{setting:main} and let $ A \subseteq [\ms a, \ms b]^{\inp} $ satisfy $ A \neq \emptyset $.
  Then there exist $ \ms c, \ms C \in \R $ such that for all $ \theta \in \R^{\dimension} $ there exists $ \vartheta \in \R^{\dimension} $
  such that $ \mc N^{\vartheta} = \mc N^{\theta} $ and
  \begin{equation}\label{eq:cor:bound:weak}
    \max\{ 1, \eucnorm{ \vartheta } \}
    \le \ms c \max\{ 1, \lipnorm{ \mc N^{\theta} }{ A } \}
    \le \ms C \max\{ 1, \eucnorm{ \vartheta }^2 \}.
  \end{equation}
  \cfout.
\end{cor}

\begin{cproof}{cor:bound:weak}
  \Nobs that \cref{cor:bound:combined} \proves that there exist $ \ms c, \ms C \in [2,\infty) $ which satisfy that for all
  $ \theta \in \R^{\dimension} $ there exists $ \vartheta \in \R^{\dimension} $ such that $ \mc N^{\vartheta} = \mc N^{\theta} $ and
  \begin{equation}\label{eq:cor:bound:weak:c}
    \eucnorm{ \vartheta }
    \le \ms c \max\{ \lipnorm{ \mc N^{\theta} }{ A }^{\nicefrac{1}{2}}, \lipnorm{ \mc N^{\theta} }{ A } \}
    \le \ms C \max\{ \eucnorm{ \vartheta }^{\nicefrac{1}{2}}, \eucnorm{ \vartheta }^2 \}
  \end{equation}
  \cfload.
  \Nobs that \cref{eq:cor:bound:weak:c} \proves that for all $ \theta \in \R^{\dimension} $ there exists $ \vartheta \in \R^{\dimension} $
  such that $ \mc N^{\vartheta} = \mc N^{\theta} $ and
  \begin{equation}
  \begin{split}
    \max\{ 1, \eucnorm{ \vartheta } \}
    &\le \max \bigl\{ 1, \ms c \max\{ \lipnorm{ \mc N^{\theta} }{ A }^{\nicefrac{1}{2}}, \lipnorm{ \mc N^{\theta} }{ A } \} \bigr\}
    \le \max \bigl\{ 1, \ms c \max\{ 1, \lipnorm{ \mc N^{\theta} }{ A } \} \bigr\} \\
    &= \ms c \max\{ 1, \lipnorm{ \mc N^{\theta} }{ A } \}
    = \max\{ \ms c, \ms c \lipnorm{ \mc N^{\theta} }{ A } \} \\
    &\le \max \bigl\{ \ms c, \ms C \max\{ \eucnorm{ \vartheta }^{\nicefrac{1}{2}}, \eucnorm{ \vartheta }^2 \} \bigr\}
    \le \max \bigl\{ \ms c, \ms C \max\{ 1, \eucnorm{ \vartheta }^2 \} \bigr\} \\
    &\le \max\{ \ms c, \ms C \} \max\{ 1, \eucnorm{ \vartheta }^2 \}.
  \end{split}
  \end{equation}
\end{cproof}

\section{Lower bounds for norms of reparameterized ANNs using Lipschitz norms}
\label{sec:lower_bounds_lipschitz_norms}

This section addresses the optimality of the upper bounds from \cref{sec:upper_bounds_lipschitz_norms} with regard to the exponents $ \nicefrac{1}{2} $ and $ 1 $ of the powers of the Lipschitz norm of the realization function and is devoted to establishing lower bounds for norms of reparameterized \ANN\ parameter vectors using Lipschitz norms. In \cref{cor:lipnorm:range:negative} in Subsection~\ref{subsec:lower_bounds_lipschitz_norms} below, we show that it is not possible to bound reparameterized \ANN\ parameter vectors from above by sums of powers of the Lipschitz norm of the realization function if the range of the exponents does not extend from $ \nicefrac{1}{2} $ to $ 1 $. Our proof of \cref{cor:lipnorm:range:negative} uses the lower bounds for reparameterized \ANNs\ established in \cref{thm:bound:range} in Subsection~\ref{subsec:lower_bounds_lipschitz_norms}, which, in turn, is based on the result for output biases of \ANNs\ with a maximum number of different kinks shown in \cref{lem:output:bias} in Subsection~\ref{subsec:output_biases} below.

\subsection{Output biases of ANNs with a maximum number of different kinks}
\label{subsec:output_biases}

\cfclear
\begin{lemma}\label{lem:output:bias}
  Assume \cref{setting:main} and let $ c \in \R $, $ \theta = ( \theta_1, \ldots, \theta_{\dimension} ) \in \R^{\dimension} $ satisfy for all
  $ x = ( x_1, \ldots, x_{\inp} ) \in [\ms a, \ms b]^{\inp} $ that
  $ \mc N^{\theta} (x) = c + \smallsum_{i=1}^{\width} \max\{ x_1 - \ms a - \tfrac{ i ( \ms b - \ms a ) }{ \width + 1 }, 0 \} $.
  Then $ \theta_{\dimension} = c $.
\end{lemma}

\begin{cproof}{lem:output:bias}
  Throughout this proof let $ u = ( 1, 0, 0, \ldots, 0 ) \in \R^{\inp} $,
  $ w = ( w_1, \ldots, w_{\width} ) = ( w_{i,j} )_{ (i,j) \in \{ 1, 2, \ldots, \width \} \times \{ 1, 2, \ldots \inp \} }
    \in \R^{ \width \times \inp } $, $ b = ( b_1, \ldots, b_{\width} ) $, $ v = ( v_1, \ldots, v_{\width} ) \in \R^{ \width } $
  satisfy for all $ i \in \{ 1, 2, \ldots, \width \} $, $ j \in \{ 1, 2, \ldots, \inp \} $ that
  \begin{equation}\label{eq:lem:output:bias:def:wbv}
    w_{i,j} = \theta_{ (i-1) \inp + j }, \qquad
    b_i = \theta_{ \inp \width + i }, \qqandqq
    v_i = \theta_{ \inp \width + \width + i },
  \end{equation}
  let $ A_k \subseteq \N $, $ k \in \{ 1, 2, 3 \} $, satisfy
  \begin{equation}\label{eq:lem:output:bias:def:ABC}
  \begin{gathered}
    A_1 = \bigl\{ i \in \{ 1, 2, \ldots, \width \} \colon \bigl( [\ms a, \ms b]^{\inp} \subseteq \half_{w_i,b_i}^1 \bigr) \bigr\}, \\
    A_2 = \bigl\{ i \in \{ 1, 2, \ldots, \width \} \colon \bigl[ \bigl( [\ms a, \ms b]^{\inp} \not\subseteq \half_{w_i,b_i}^1 \bigr) \wedge
      \bigl( \half_{w_i,b_i}^1 \cap (\ms a, \ms b)^{\inp} \neq \emptyset \bigr) \bigr] \bigr\}, \\ \text{and} \qquad
    A_3 = \bigl\{ i \in \{ 1, 2, \ldots, \width \} \colon \bigl( \half_{w_i,b_i}^1 \cap (\ms a, \ms b)^{\inp} = \emptyset \bigr) \bigr\},
  \end{gathered}
  \end{equation}
  let $ N \in \N $ satisfy $ N = \# ( \smallbigcup_{ i \in A_2 } \{ \hyper_{w_i,b_i} \} ) $,
  let $ A_4 \subseteq A_2 $ satisfy for all $ i, j \in A_4 $ with $ i \neq j $ that
  $ \hyper_{w_i,b_i} \neq \hyper_{w_j,b_j} $ and $ \# A_4 = N $,
  and let $ q_1, q_2, \ldots, q_{\width} \in [\ms a, \ms b]^{\inp} $, $ \varepsilon \in (0, \infty) $
  satisfy for all $ i \in \{ 1, 2, \ldots, \width \} $ that
  \begin{equation}\label{eq:lem:output:bias:def:q}
    q_i = \Bigl( \ms a + \tfrac{ i (\ms b - \ms a) }{ \width + 1 }, \ms a, \ms a, \ldots, \ms a \Bigr)
    \qqandqq
    \varepsilon < \tfrac{ \ms b - \ms a }{ \width + 1 }.
  \end{equation}
  \cfload.
  \Nobs that \cref{eq:lem:output:bias:def:ABC} \proves that for all $ i, j \in \{ 1, 2, 3 \} $ with $ i \neq j $ it holds that
  \begin{equation}\label{eq:lem:output:bias:ABC:properties}
    \{ 1, 2, \ldots, \width \} = A_1 \cup A_2 \cup A_3
    \qqandqq
    A_i \cap A_j = \emptyset.
  \end{equation}
  We now \prove by contradiction that for all $ i \in \{ 1, 2, \ldots, \width \} $ there exists $ j \in \{ 1, 2, \ldots, \width \} $ such that
  \begin{equation}
    \hyper_{w_j,b_j}
    = \Bigl\{ x \in \R^{\inp} \colon \scalprod{ u }{ x } - \ms a - \tfrac{ i (\ms b - \ms a) }{ \width + 1 } = 0 \Bigr\}
  \end{equation}
  \cfload.
  In the following, we thus assume that there exists $ i \in \{ 1, 2, \ldots, \width \} $ which satisfies that for all
  $ j \in \{ 1, 2, \ldots, \width \} $ it holds that
  \begin{equation}\label{eq:lem:output:bias:assumption}
    \hyper_{w_j,b_j}
    \neq \Bigl\{ x \in \R^{\inp} \colon \scalprod{ u }{ x } - \ms a - \tfrac{ i (\ms b - \ms a) }{ \width + 1 } = 0 \Bigr\}.
  \end{equation}
  \Nobs that \cref{lem:hyperplanes:isolation} (applied with $ \inp \is \inp $, $ N \is N + 1 $, $ \ms a \is \ms a $, $ \ms b \is \ms b $,
  $ ( w_j )_{ j \in \{ 1, 2, \ldots, N - 1 \} } \is ( w_j )_{ j \in A_4 } $, $ w_N \is u $,
  $ ( b_j )_{ j \in \{ 1, 2, \ldots, N - 1 \} } \is ( b_j )_{ j \in A_4 } $, $ b_N \is - \ms a - i (\ms b - \ms a) ( \width + 1 )^{-1} $
  in the notation of \cref{lem:hyperplanes:isolation})
  and \cref{eq:lem:output:bias:assumption} \prove that there exist $ p \in (\ms a, \ms b)^{\inp} $,
  $ \delta \in (0, \tfrac{ \ms b - \ms a }{ \width + 1 } ) $ which satisfy that
  \begin{enumerate}[(i)]
    \item \label{item:lem:output:bias:nonlinearity:1} it holds that
      $ \scalprod{ u }{ p } - \ms a - \tfrac{ i (\ms b - \ms a) }{ \width + 1 } = 0 $,
    \item \label{item:lem:output:bias:nonlinearity:2} it holds that
      $ \{ x \in \R^{\inp} \colon \eucnorm{ x - p } \le \delta \} \subseteq [\ms a, \ms b]^{\inp} $, and
    \item \label{item:lem:output:bias:nonlinearity:3} it holds that
      $ \{ x \in \R^{\inp} \colon \eucnorm{ x - p } \le \delta \}
        \cap ( \smallbigcup_{ j \in A_2 } \hyper_{w_j,b_j} ) = \emptyset $.
  \end{enumerate}
  \Nobs that \cref{item:lem:output:bias:nonlinearity:1,item:lem:output:bias:nonlinearity:2} \prove that for all
  $ x \in \{ y \in \R^{\inp} \colon \eucnorm{ y } \le \delta \} $ it holds that
  \begin{equation}
  \begin{split}
    \mc N^{\theta} ( p + x )
    &= c + \smallsum_{j=1}^{\width} \max \bigl\{ \scalprod{ u }{ p + x } - \ms a - \tfrac{ j (\ms b - \ms a) }{ \width + 1 }, 0 \bigr\} \\
    &= c + \smallsum_{j=1}^{\width} \max \bigl\{ \scalprod{ u }{ p } - \ms a - \tfrac{ i (\ms b - \ms a ) }{ \width + 1 }
      + \tfrac{ (i-j) (\ms b - \ms a) }{ \width + 1 } + \scalprod{ u }{ x }, 0 \bigr\} \\
    &= c + \smallsum_{j=1}^{\width} \max \bigl\{ \tfrac{ (i-j) (\ms b - \ms a) }{ \width + 1 } + \scalprod{ u }{ x }, 0 \bigr\}.
  \end{split}
  \end{equation}
  \Hence that
  \begin{equation}\label{eq:lem:output:bias:nonlinearity}
  \begin{split}
    &\mc N^{\theta} ( p + \delta u ) - 2 \mc N^{\theta} ( p ) + \mc N^{\theta} ( p - \delta u ) \\
    &= \smallsum_{j=1}^{\width} \max \bigl\{ \tfrac{ (i-j) (\ms b - \ms a) }{ \width + 1 } + \delta, 0 \bigr\}
      - 2 \smallsum_{j=1}^{\width} \max \bigl\{ \tfrac{ (i-j) (\ms b - \ms a) }{ \width + 1 }, 0 \bigr\} \\
    &\quad + \smallsum_{j=1}^{\width} \max \bigl\{ \tfrac{ (i-j) (\ms b - \ms a) }{ \width + 1 } - \delta, 0 \bigr\} \\
    &= \smallsum_{j=1}^i \Bigl[ \tfrac{ (i-j) (\ms b - \ms a) }{ \width + 1 } + \delta \Bigr]
      - 2 \smallsum_{j=1}^i \tfrac{ (i-j) (\ms b - \ms a) }{ \width + 1 }
      + \smallsum_{j=1}^{i-1} \Bigl[ \tfrac{ (i-j) (\ms b - \ms a) }{ \width + 1 } - \delta \Bigr]
    = \delta.
  \end{split}
  \end{equation}
  \Moreover \cref{item:lem:output:bias:nonlinearity:2,item:lem:output:bias:nonlinearity:3} \prove that for all
  $ x \in \{ y \in \R^{\inp} \colon \eucnorm{ y - p } \le \delta \} $ it holds that
  \begin{equation}
    \bigl\{ j \in A_2 \colon p \in \half_{w_j,b_j}^1 \bigr\}
    = \bigl\{ j \in A_2 \colon p + x \in \half_{w_j,b_j}^1 \bigr\}.
  \end{equation}
  Combining this, \cref{eq:lem:output:bias:ABC:properties}, and the fact that for all $ j \in A_3 $, $ x \in \R^{\inp} $
  it holds that $ b_j + \scalprod{ w_j }{ x } \le 0 $ \proves that for all
  $ x \in \{ y \in \R^{\inp} \colon \eucnorm{ y - p } \le \delta \} $ it holds that
  \begin{equation}
  \begin{split}
    \mc N^{\theta} (x)
    &= \theta_{\dimension} + \smallsum_{j=1}^{\width} v_j \max\{ b_j + \scalprod{ w_j }{ x }, 0 \} \\
    &= \theta_{\dimension} + \smallsum_{ j \in A_1 } v_j \bigl( b_j + \scalprod{ w_j }{ x } \bigr)
      + \smallsum_{ j \in A_2 } v_j \max\{ b_j + \scalprod{ w_j }{ x }, 0 \} \\
    &= \theta_{\dimension} + \smallsum_{ j \in A_1 } v_j \bigl( b_j + \scalprod{ w_j }{ x } \bigr)
      + \smallsum_{ j \in A_2, \, p \in \half_{\raisebox{0pt}{$\smash{\scriptscriptstyle w_j,b_j}$}}^1 }
      v_j \bigl( b_j + \scalprod{ w_j }{ x } \bigr) \\
    &= \theta_{\dimension} + \smallsum_{ j \in A_1 \cup A_2, \, p \in \half_{\raisebox{0pt}{$\smash{\scriptscriptstyle w_j,b_j}$}}^1 }
      v_j \bigl( b_j + \scalprod{ w_j }{ x } \bigr) \\
    &= \theta_{\dimension} + \smallsum_{ j \in A_1 \cup A_2, \, p \in \half_{\raisebox{0pt}{$\smash{\scriptscriptstyle w_j,b_j}$}}^1 } v_j b_j
      + \smallsum_{ j \in A_1 \cup A_2, \, p \in \half_{\raisebox{0pt}{$\smash{\scriptscriptstyle w_j,b_j}$}}^1 } v_j \scalprod{ w_j }{ x }.
  \end{split}
  \end{equation}
  This and \cref{eq:lem:output:bias:nonlinearity} \prove that
  \begin{equation}
  \begin{split}
    \delta
    &= \mc N^{\theta} ( p + \delta u ) - 2 \mc N^{\theta} ( p ) + \mc N^{\theta} ( p - \delta u ) \\
    &= \smallsum_{ j \in A_1 \cup A_2, \, p \in \half_{\raisebox{0pt}{$\smash{\scriptscriptstyle w_j,b_j}$}}^1 } v_j \scalprod{ w_j }{ p + \delta u }
      - 2 \smallsum_{ j \in A_1 \cup A_2, \, p \in \half_{\raisebox{0pt}{$\smash{\scriptscriptstyle w_j,b_j}$}}^1 } v_j \scalprod{ w_j }{ p } \\
    &\quad + \smallsum_{ j \in A_1 \cup A_2, \, p \in \half_{\raisebox{0pt}{$\smash{\scriptscriptstyle w_j,b_j}$}}^1 } v_j \scalprod{ w_j }{ p - \delta u } \\
    &= \smallsum_{ j \in A_1 \cup A_2, \, p \in \half_{\raisebox{0pt}{$\smash{\scriptscriptstyle w_j,b_j}$}}^1 } v_j \scalprod{ w_j }{ \delta u }
      - \smallsum_{ j \in A_1 \cup A_2, \, p \in \half_{\raisebox{0pt}{$\smash{\scriptscriptstyle w_j,b_j}$}}^1 } v_j \scalprod{ w_j }{ \delta u }
    = 0.
  \end{split}
  \end{equation}
  This is a contradiction to the fact that $ \delta > 0 $.
  \Hence that for all $ i \in \{ 1, 2, \ldots, \width \} $ there exists $ j \in \{ 1, 2, \ldots, \width \} $ such that
  \begin{equation}
    \hyper_{w_j,b_j}
    = \Bigl\{ x \in \R^{\inp} \colon \scalprod{ u }{ x } - \ms a - \tfrac{ i (\ms b - \ms a) }{ \width + 1 } = 0 \Bigr\}.
  \end{equation}
  This \proves that there exists a bijective function $ \varphi \colon \{ 1, 2, \ldots, \width \} \to \{ 1, 2, \ldots, \width \} $
  which satisfies that for all $ i \in \{ 1, 2, \ldots, \width \} $ it holds that
  \begin{equation}\label{eq:lem:output:bias:def:varphi}
    \hyper_{w_{\varphi(i)},b_{\varphi(i)}}
    = \Bigl\{ x \in \R^{\inp} \colon \scalprod{ u }{ x } - \ms a - \tfrac{ i (\ms b - \ms a) }{ \width + 1 } = 0 \Bigr\}.
  \end{equation}
  \Nobs that \cref{eq:lem:output:bias:def:q} \proves that for all $ i \in \{ 1, 2, \ldots, \width \} $, $ t \in [ - \varepsilon, \varepsilon ] $
  it holds that $ q_i + t u \in [\ms a, \ms b]^{\inp} $ and
  \begin{equation}\label{eq:lem:output:bias:realization:1}
  \begin{split}
    \mc N^{\theta} ( q_i + t u )
    &= c + \smallsum_{j=1}^{\width} \max \Bigl\{ \scalprod{ u }{ q_i + t u } - \ms a
      - \tfrac{ j (\ms b - \ms a) }{ \width + 1 }, 0 \Bigr\} \\
    &= c + \smallsum_{j=1}^{\width} \max \Bigl\{ \scalprod{ u }{ q_i } + t \scalprod{ u }{ u } - \ms a
      - \tfrac{ j (\ms b - \ms a) }{ \width + 1 }, 0 \Bigr\} \\
    &= c + \smallsum_{j=1}^{\width} \max \Bigl\{ \tfrac{ ( i - j ) (\ms b - \ms a) }{ \width + 1 } + t, 0 \Bigr\}.
  \end{split}
  \end{equation}
  \Hence that for all $ i \in \{ 1, 2, \ldots, \width \} $ it holds that
  \begin{equation}\label{eq:lem:output:bias:nonlinearity:i}
  \begin{split}
    &\mc N^{\theta} ( q_i + \varepsilon u ) - 2 \mc N^{\theta} ( q_i ) + \mc N^{\theta} ( q_i - \varepsilon u ) \\
    &= \smallsum_{j=1}^{\width} \max \Bigl\{ \tfrac{ (i-j) (\ms b - \ms a) }{ \width + 1 } + \varepsilon, 0 \Bigr\}
      - 2 \smallsum_{j=1}^{\width} \max \Bigl\{ \tfrac{ (i-j) (\ms b - \ms a) }{ \width + 1 }, 0 \Bigr\} \\
    &\quad + \smallsum_{j=1}^{\width} \max \Bigl\{ \tfrac{ (i-j) (\ms b - \ms a) }{ \width + 1 } - \varepsilon, 0 \Bigr\} \\
    &= \smallsum_{j=1}^i \Bigl[ \tfrac{ (i-j) (\ms b - \ms a) }{ \width + 1 } + \varepsilon \Bigr]
      - 2 \smallsum_{j=1}^i \tfrac{ (i-j) (\ms b - \ms a) }{ \width + 1 }
      + \smallsum_{j=1}^{i-1} \Bigl[ \tfrac{ (i-j) (\ms b - \ms a) }{ \width + 1 } - \varepsilon \Bigr]
    = \varepsilon.
  \end{split}
  \end{equation}
  \Moreover \cref{eq:lem:output:bias:def:varphi} and the fact that for all $ i \in \{ 1, 2, \ldots, \width \} $ it holds that
  $ q_i \in \hyper_{w_{\varphi(i)},b_{\varphi(i)}} $ \prove that for all
  $ i \in \{ 1, 2, \ldots, \width \} $, $ t \in [ - \varepsilon, \varepsilon ] $ it holds that
  \begin{equation}
    \bigl\{ j \in \{ 1, 2, \ldots, \width \} \backslash \{ \varphi(i) \} \colon q_i \in \half_{w_j,b_j}^1 \bigr\}
    = \bigl\{ j \in \{ 1, 2, \ldots, \width \} \backslash \{ \varphi(i) \} \colon q_i + t u \in \half_{w_j,b_j}^1 \bigr\}.
  \end{equation}
  This and the fact that $ A_1 \cup A_3 = \emptyset $ \prove that for all
  $ i \in \{ 1, 2, \ldots, \width \} $, $ t \in [ - \varepsilon, \varepsilon ] $ it holds that
  \begin{equation}\label{eq:lem:output:bias:realization:2}
  \begin{split}
    \mc N^{\theta} ( q_i + t u )
    &= \theta_{\dimension} + \smallsum_{j=1}^{\width} v_j \max\{ b_j + \scalprod{ w_i }{ q_i + t u }, 0 \} \\
    &= \theta_{\dimension} + \smallsum_{ j \in A_2 \backslash \{ \varphi(i) \} } v_j \max\{ b_j + \scalprod{ w_j }{ q_i + t u }, 0 \} \\
    &\quad + v_{ \varphi(i) } \max\{ b_{ \varphi(i) } + \scalprod{ w_{ \varphi(i) } }{ q_i + t u }, 0 \} \\
    &= \theta_{\dimension} + \smallsum_{ j \in A_2 \backslash \{ \varphi(i) \}, \, q_i \in \half_{\raisebox{0pt}{$\smash{\scriptscriptstyle w_j,b_j}$}}^1 }
      v_j \bigl( b_j + \scalprod{ w_j }{ q_i + t u } \bigr)
      + v_{ \varphi(i) } \max\{ t \scalprod{ w_{ \varphi(i) } }{ u }, 0 \} \\
    &= \theta_{\dimension} + \smallsum_{ j \in A_2 \backslash \{ \varphi(i) \}, \, q_i \in \half_{\raisebox{0pt}{$\smash{\scriptscriptstyle w_j,b_j}$}}^1 }
      v_j \bigl( b_j + \scalprod{ w_j }{ q_i } \bigr)
      + \smallsum_{ j \in A_2 \backslash \{ \varphi(i) \}, \, q_i \in \half_{\raisebox{0pt}{$\smash{\scriptscriptstyle w_j,b_j}$}}^1 }
      t v_j \scalprod{ w_j }{ u } \\
    &\quad + v_{ \varphi(i) } \max\{ t \scalprod{ w_{ \varphi(i) } }{ u }, 0 \}
  \end{split}
  \end{equation}
  The fact that for all $ x \in \R $ it holds that $ \max\{ x, 0 \} + \max\{ - x, 0 \} = \abs{ x } $ and
  \cref{eq:lem:output:bias:nonlinearity:i} \hence \prove that for all $ i \in \{ 1, 2, \ldots, \width \} $ it holds that
  \begin{equation}
  \begin{split}
    \varepsilon
    &= \mc N^{\theta} ( q_i + \varepsilon u ) - 2 \mc N^{\theta} ( q_i ) + \mc N^{\theta} ( q_i - \varepsilon u ) \\
    &= \smallsum_{ j \in A_2 \backslash \{ \varphi(i) \}, \, q_i \in \half_{\raisebox{0pt}{$\smash{\scriptscriptstyle w_j,b_j}$}}^1 }
      \varepsilon v_j \scalprod{ w_j }{ u }
      + v_{ \varphi(i) } \max\{ \varepsilon \scalprod{ w_{ \varphi(i) } }{ u }, 0 \} \\
    &\quad - \smallsum_{ j \in A_2 \backslash \{ \varphi(i) \}, \, q_i \in \half_{\raisebox{0pt}{$\smash{\scriptscriptstyle w_j,b_j}$}}^1 }
      \varepsilon v_j \scalprod{ w_j }{ u }
      + v_{ \varphi(i) } \max\{ - \varepsilon \scalprod{ w_{ \varphi(i) } }{ u }, 0 \} \\
    &= \varepsilon v_{ \varphi(i) } \bigl( \max\{ \scalprod{ w_{ \varphi(i) } }{ u }, 0 \}
      + \max\{ - \scalprod{ w_{ \varphi(i) } }{ u }, 0 \} \bigr)
    = \varepsilon v_{ \varphi(i) } \abs{ \scalprod{ w_{ \varphi(i) } }{ u } }.
  \end{split}
  \end{equation}
  Combining this and the fact that $ \varphi $ is bijective \proves that for all $ i \in \{ 1, 2, \ldots, \width \} $ it holds that
  \begin{equation}\label{eq:lem:output:bias:slope}
    v_i \abs{ \scalprod{ w_i }{ u } } = 1.
  \end{equation}
  \Moreover \cref{eq:lem:output:bias:realization:1} \proves that it holds that
  \begin{equation}\label{eq:lem:output:bias:difference}
  \begin{split}
    &\mc N^{\theta} \bigl( q_{\width} + \varepsilon u \bigr) - \mc N^{\theta} \bigl( q_{\width} + \tfrac{1}{2} \varepsilon u \bigr) \\
    &= \smallsum_{i=1}^{\width} \max \Bigl\{ \tfrac{ (\width-i) (\ms b - \ms a) }{ \width + 1 } + \varepsilon, 0 \Bigr\}
      - \smallsum_{i=1}^{\width} \max \Bigl\{ \tfrac{ (\width-i) (\ms b - \ms a) }{ \width + 1 } + \tfrac{1}{2} \varepsilon, 0 \Bigr\} \\
    &= \smallsum_{i=1}^{\width} \Bigl[ \tfrac{ (\width-i) (\ms b - \ms a) }{ \width + 1 } + \varepsilon \Bigr]
      - \smallsum_{i=1}^{\width} \Bigl[ \tfrac{ (\width-i) (\ms b - \ms a) }{ \width + 1 } + \tfrac{1}{2} \varepsilon \Bigr]
    = \varepsilon \width - \tfrac{1}{2} \varepsilon \width
    = \tfrac{1}{2} \varepsilon \width.
  \end{split}
  \end{equation}
  \Moreover \cref{eq:lem:output:bias:def:varphi} \proves that
  \begin{equation}
    \bigl\{ i \in \{ 1, 2, \ldots, \width \} \colon q_{\width} + \varepsilon u \in \half_{w_i,b_i}^1 \bigr\}
    = \bigl\{ i \in \{ 1, 2, \ldots, \width \} \colon q_{\width} + \tfrac{1}{2} \varepsilon u \in \half_{w_i,b_i}^1 \bigr\}.
  \end{equation}
  Combining this, \cref{eq:lem:output:bias:difference}, and the fact that $ A_1 \cup A_3 = \emptyset $ \proves that
  \begin{equation}\label{eq:lem:output:bias:edge}
  \begin{split}
    \tfrac{1}{2} \varepsilon \width
    &= \mc N^{\theta} ( q_{\width} + \varepsilon u ) - \mc N^{\theta} ( q_{\width} + \tfrac{1}{2} \varepsilon u ) \\
    &= \smallsum_{i=1}^{\width} v_i \max \bigl\{ b_i + \scalprod{ w_i }{ q_{\width} + \varepsilon u }, 0 \bigr\}
      - \smallsum_{i=1}^{\width} v_i \max \bigl\{ b_i + \bscalprod{ w_i }{ q_{\width} + \tfrac{1}{2} \varepsilon u }, 0 \bigr\} \\
    &= \smallsum_{ i \in A_2, \, q_{\width} + \varepsilon u \in \half_{\raisebox{0pt}{$\smash{\scriptscriptstyle w_i,b_i}$}}^1 }
      v_i \bigl( b_i + \scalprod{ w_i }{ q_{\width} + \varepsilon u } \bigr) \\
    &\quad - \smallsum_{ i \in A_2, \, q_{\width} + \varepsilon u \in \half_{\raisebox{0pt}{$\smash{\scriptscriptstyle w_i,b_i}$}}^1 }
      v_i \bigl( b_i + \scalprod{ w_i }{ q_{\width} + \tfrac{1}{2} \varepsilon u } \bigr) \\
    &= \smallsum_{ i \in A_2, \, q_{\width} + \varepsilon u \in \half_{\raisebox{0pt}{$\smash{\scriptscriptstyle w_i,b_i}$}}^1 }
      v_i \scalprod{ w_i }{ \varepsilon u }
      - \smallsum_{ i \in A_2, \, q_{\width} + \varepsilon u \in \half_{\raisebox{0pt}{$\smash{\scriptscriptstyle w_i,b_i}$}}^1 }
      v_i \scalprod{ w_i }{ \tfrac{1}{2} \varepsilon u } \\
    &= \tfrac{1}{2} \varepsilon \smallsum_{ i \in A_2, \, q_{\width} + \varepsilon u \in \half_{\raisebox{0pt}{$\smash{\scriptscriptstyle w_i,b_i}$}}^1 }
      v_i \scalprod{ w_i }{ u }.
  \end{split}
  \end{equation}
  \Moreover \cref{eq:lem:output:bias:slope} \proves that for all $ i \in \{ 1, 2, \ldots, \width \} $ it holds that $ v_i > 0 $.
  This, \cref{eq:lem:output:bias:slope}, and \cref{eq:lem:output:bias:edge} \prove that
  \begin{equation}
  \begin{split}
    \width
    &= \smallsum_{ i \in A_2, \, q_{\width} + \varepsilon u \in \half_{\raisebox{0pt}{$\smash{\scriptscriptstyle w_i,b_i}$}}^1 }
      v_i \scalprod{ w_i }{ u }
    \le \smallsum_{ i \in A_2, \, q_{\width} + \varepsilon u \in \half_{\raisebox{0pt}{$\smash{\scriptscriptstyle w_i,b_i}$}}^1 }
      v_i \abs{ \scalprod{ w_i }{ u } } \\
    &= \# \bigl\{ i \in A_2 \colon q_{\width} + \varepsilon u \in \half_{w_i,b_i}^1 \bigr\}
    \le \width.
  \end{split}
  \end{equation}
  \Hence that for all $ i \in \{ 1, 2, \ldots, \width \} $ it holds that $ q_{\width} + \varepsilon u \in \half_{w_i,b_i}^1 $.
  This \proves that for all $ i \in \{ 1, 2, \ldots, \width \} $ it holds that $ q_1 - \varepsilon u \in \half_{w_i,b_i}^0 $.
  Combining this with \cref{eq:lem:output:bias:def:q} \proves that
  \begin{equation}
  \begin{split}
    \theta_{\dimension}
    &= \theta_{\dimension} + \smallsum_{i=1}^{\width} v_i \max\{ b_i + \scalprod{ w_i }{ q_1 - \varepsilon u }, 0 \}
    = \mc N^{\theta} ( q_1 - \varepsilon u ) \\
    &= c + \smallsum_{i=1}^{\width} \max \Bigl\{ \scalprod{ u }{ q_1 - \varepsilon u }
      - \ms a - \tfrac{ i (\ms b - \ms a) }{ \width + 1 }, 0 \Bigr\} \\
    &= c + \smallsum_{i=1}^{\width} \max \Bigl\{ \tfrac{ (1-i) (\ms b - \ms a) }{ \width + 1 } - \varepsilon, 0 \Bigr\}
    = c
  \end{split}
  \end{equation}
\end{cproof}

\subsection{Lower bounds for norms of reparameterized ANNs using Lipschitz norms}
\label{subsec:lower_bounds_lipschitz_norms}

\cfclear
\begin{theorem}\label{thm:bound:range}
  Assume \cref{setting:main} and let $ \varepsilon \in [0,\infty) $, $ \delta \in [\varepsilon, \infty) $, $ A \subseteq [\ms a, \ms b]^{\inp} $
  satisfy $ [ \varepsilon > \nicefrac{1}{2} ] \vee [ \delta < 1 ] $ and $ A \neq \emptyset $.
  Then for all $ \ms c \in \R $ there exists $ \theta \in \R^{\dimension} $ such that for all
  $ \vartheta \in \{ \eta \in \R^{\dimension} \colon \mc N^{\eta} = \mc N^{\theta} \} $ it holds that
  \begin{equation}\label{eq:thm:bound:range}
    \eucnorm{ \vartheta }
    > \ms c \max \bigl\{ \lipnorm{ \mc N^{\theta} }{ A }^{\varepsilon}, \lipnorm{ \mc N^{\theta} }{ A }^{\delta} \bigr\}.
  \end{equation}
  \cfout.
\end{theorem}

\begin{cproof}{thm:bound:range}
  In the following, we distinguish between the case $ \varepsilon > \nicefrac{1}{2} $ and the case $ \delta < 1 $.
  We first \prove[p] \cref{eq:thm:bound:range} in the case
  \begin{equation}\label{eq:thm:bound:range:case1}
    \varepsilon > \nicefrac{1}{2}.
  \end{equation}
  Let $ \theta = ( \theta_1, \ldots, \theta_{\dimension} ) \colon \N \to \R^{\dimension} $ satisfy for all $ n \in \N $,
  $ i \in \{ 2, 3, \ldots, \width \} $, $ j \in \{ 1, 2, \ldots, \inp \} $ that
  \begin{equation}\label{eq:thm:bound:range:theta1}
    \theta_1 (n) = 1, \qquad
    \theta_{ \inp \width + 1 } (n) = - \ms a, \qquad
    \theta_{ \inp \width + \width + 1 } (n) = n^{-1},
  \end{equation}
  and $ \theta_i (n) = \theta_{ (i-1) \inp + j } (n) = \theta_{ \inp \width + i } (n) = \theta_{ \inp \width + \width + i } (n)
    = \theta_{\dimension} (n) = 0 $.
  \Nobs that \cref{eq:thm:bound:range:theta1} \proves that for all $ n \in \N $, $ x = ( x_1, \ldots, x_{\inp} ) \in [\ms a, \ms b]^{\inp} $
  it holds that
  \begin{equation}\label{eq:thm:bound:range:realization1}
  \begin{split}
    \mc N^{\theta(n)} (x)
    &= \theta_{\dimension}(n) + \smallsum_{i=1}^{\width} \theta_{ \inp \width + \width + i }(n)
      \max \bigl\{ \theta_{ \inp \width + i }(n) + \smallsum_{j=1}^{\inp} \theta_{ (i-1) \inp + j }(n) x_j, 0 \bigr\} \\
    &= n^{-1} \max\{ - \ms a + x_1, 0 \}
    = n^{-1} ( x_1 - \ms a ).
  \end{split}
  \end{equation}
  \Hence that for all $ n \in \N $, $ x = ( x_1, \ldots, x_{\inp} ) \in [\ms a, \ms b]^{\inp} $ it holds that
  \begin{equation}\label{eq:thm:bound:range:absolute1}
    \abs{ \mc N^{\theta(n)} (x) }
    = n^{-1} ( x_1 - \ms a )
    \le n^{-1} ( \ms b - \ms a ).
  \end{equation}
  \Moreover \cref{eq:thm:bound:range:realization1} \proves that for all
  $ n \in \N $, $ x = ( x_1, \ldots, x_d ) $, $ y = ( y_1, \ldots, y_d ) \in [\ms a, \ms b]^{\inp} $ it holds that
  \begin{equation}\label{eq:thm:bound:range:lipschitz1}
    \abs{ \mc N^{\theta(n)} (x) - \mc N^{\theta(n)} (y) }
    = \abs{ n^{-1} ( x_1 - \ms a ) - n^{-1} ( y_1 - \ms a ) }
    = n^{-1} \abs{ x_1 - y_1 }
    \le n^{-1} \eucnorm{ x - y }
  \end{equation}
  \cfload.
  Combining this and \cref{eq:thm:bound:range:absolute1} \proves that for all $ n \in \N $ it holds that
  \begin{equation}\label{eq:thm:bound:range:lipnorm1}
  \begin{split}
    \lipnorm{ \mc N^{\theta(n)} }{ A }
    &= \inf\nolimits_{ x \in A } \abs{ \mc N^{\theta(n)} (x) }
      + \sup\nolimits_{ x, y \in [\ms a, \ms b]^{\inp}, \, x \neq y } \tfrac{ \abs{ \mc N^{\theta(n)} (x) - \mc N^{\theta(n)} (y) } }
        { \eucnorm{ x - y } } \\
    &\le n^{-1} ( \ms b - \ms a ) + n^{-1}
    = ( \ms b - \ms a + 1 ) n^{-1}
  \end{split}
  \end{equation}
  \cfload.
  \Moreover \cref{lem:lipschitzconstant:bound} and \cref{eq:thm:bound:range:realization1} \prove that for all
  $ n \in \N $, $ \vartheta \in \{ \eta \in \R^{\dimension} \colon \mc N^{\eta} = \mc N^{\theta(n)} \} $ it holds that
  \begin{equation}
  \begin{split}
    \tfrac{1}{2} \eucnorm{ \vartheta }^2
    &\ge \sup\nolimits_{ x, y \in [\ms a, \ms b]^{\inp}, \, x \neq y } \tfrac{ \abs{ \mc N^{\vartheta}(x) - \mc N^{\vartheta}(y) } }
      { \eucnorm{ x - y } }
    \ge \tfrac{ 1 }{ \ms b - \ms a } \abs{ \mc N^{\vartheta} ( \ms b, \ms b, \ldots, \ms b )
      - \mc N^{\vartheta} ( \ms a, \ms b, \ldots, \ms b ) } \\
    &= \tfrac{ 1 }{ \ms b - \ms a } \abs{ \mc N^{\theta(n)} ( \ms b, \ms b, \ldots, \ms b )
      - \mc N^{\theta(n)} ( \ms a, \ms b, \ldots, \ms b ) }
    = \tfrac{ 1 }{ \ms b - \ms a } n^{-1} ( \ms b - \ms a )
    = n^{-1}.
  \end{split}
  \end{equation}
  The fact that $ \lim_{ n \to \infty } n^{ \varepsilon - \nicefrac{1}{2} } = \infty $ and \cref{eq:thm:bound:range:lipnorm1} \hence
  \prove that for all $ \ms c \in [0,\infty) $ there exists $ n \in \N $ such that for all
  $ \vartheta \in \{ \eta \in \R^{\dimension} \colon \mc N^{\eta} = \mc N^{\theta(n)} \} $ it holds that
  $ \lipnorm{ \mc N^{\theta(n)} }{ A } \le 1 $ and
  \begin{equation}
  \begin{split}
    \eucnorm{ \vartheta }
    &\ge \sqrt{2} n^{ - \nicefrac{1}{2} }
    = \sqrt{2} n^{ \varepsilon - \nicefrac{1}{2} } n^{ - \varepsilon }
    > \ms c ( \ms b - \ms a + 1 )^{\varepsilon} n^{ - \varepsilon }
    = \ms c \bigl[ ( \ms b - \ms a + 1 ) n^{-1} \bigr]^{\varepsilon} \\
    &\ge \ms c \lipnorm{ \mc N^{\theta(n)} }{ A }^{\varepsilon}
    = \ms c \max \bigl\{ \lipnorm{ \mc N^{\theta(n)} }{ A }^{\varepsilon}, \lipnorm{ \mc N^{\theta(n)} }{ A }^{\delta} \bigr\}.
  \end{split}
  \end{equation}
  \Hence that for all $ \ms c \in \R $ there exists $ \theta \in \R^{\dimension} $ such that for all
  $ \vartheta \in \{ \eta \in \R^{\dimension} \colon \mc N^{\eta} = \mc N^{\theta} \} $ it holds that
  \begin{equation}
    \eucnorm{ \vartheta }
    > \ms c \max \bigl\{ \lipnorm{ \mc N^{\theta} }{ A }^{\varepsilon}, \lipnorm{ \mc N^{\theta} }{ A }^{\delta} \bigr\}.
  \end{equation}
  This \proves \cref{eq:thm:bound:range} in the case $ \varepsilon > \nicefrac{1}{2} $.
  In the next step we will \prove[p] \cref{eq:thm:bound:range} in the case
  \begin{equation}\label{eq:thm:bound:range:case2}
    \delta < 1.
  \end{equation}
  Let $ \theta = ( \theta_1, \ldots, \theta_{\dimension} ) \colon \N \to \R^{\dimension} $ satisfy for all $ n \in \N $,
  $ i \in \{ 1, 2, \ldots, \width \} $, $ j \in \{ 2, 3, \ldots, \inp \} $ that
  \begin{equation}\label{eq:thm:bound:range:theta2}
  \begin{gathered}
    \theta_{ (i-1) \inp + 1 } (n) = 1, \qquad
    \theta_{ (i-1) \inp + j } (n) = 0, \qquad
    \theta_{ \inp \width + i } (n) = - \ms a - \tfrac{ i ( \ms b - \ms a ) }{ \width + 1 }, \\
    \theta_{ \inp \width + \width + i } (n) = 1, \qqandqq
    \theta_{\dimension} (n) = n.
  \end{gathered}
  \end{equation}
  \Nobs that \cref{eq:thm:bound:range:theta2} \proves that for all $ n \in \N $, $ x = ( x_1, \ldots, x_{\inp} ) \in [\ms a, \ms b]^{\inp} $
  it holds that
  \begin{equation}\label{eq:thm:bound:range:realization2}
  \begin{split}
    \mc N^{\theta(n)} (x)
    &= \theta_{\dimension}(n) + \smallsum_{i=1}^{\width} \theta_{ \inp \width + \width + i }(n)
      \max \bigl\{ \theta_{ \inp \width + i }(n) + \smallsum_{j=1}^{\inp} \theta_{ (i-1) \inp + j }(n) x_j, 0 \bigr\} \\
    &= n + \smallsum_{i=1}^{\width} \max \bigl\{ x_1 - \ms a - \tfrac{ i ( \ms b - \ms a ) }{ \width + 1 }, 0 \bigr\}.
  \end{split}
  \end{equation}
  \Hence that for all $ n \in \N $, $ x = ( x_1, \ldots, x_{\inp} ) \in [\ms a, \ms b]^{\inp} $ it holds that
  \begin{equation}\label{eq:thm:bound:range:absolute2}
  \begin{split}
    \abs{ \mc N^{\theta(n)} (x) }
    &= n + \smallsum_{i=1}^{\width} \max \bigl\{ x_1 - \ms a - \tfrac{ i ( \ms b - \ms a ) }{ \width + 1 }, 0 \bigr\} \\
    &\le n + \smallsum_{i=1}^{\width} \bigl( \ms b - \ms a - \tfrac{ i ( \ms b - \ms a ) }{ \width + 1 } \bigr)
    \le n + \width ( \ms b - \ms a ).
  \end{split}
  \end{equation}
  \Moreover \cref{eq:thm:bound:range:realization2} and
  the fact that for all $ x, y \in \R $ it holds that $ \abs{ \max\{ x, 0 \} - \max\{ y, \allowbreak 0 \} } \le \abs{ x - y } $
  \prove that for all $ n \in \N $, $ x = ( x_1, \ldots, x_d ) $, $ y = ( y_1, \ldots, y_d) \in [\ms a, \ms b]^{\inp} $ it holds that
  \begin{equation}\label{eq:thm:bound:range:lipschitz2}
  \begin{split}
    \abs{ \mc N^{\theta(n)} (x) - \mc N^{\theta(n)} (y) }
    &= \bbabs{ \smallsum_{i=1}^{\width} \bigl( \max \bigl\{ x_1 - \ms a - \tfrac{ i ( \ms b - \ms a ) }{ \width + 1 }, 0 \bigr\}
      - \max \bigl\{ y_1 - \ms a - \tfrac{ i ( \ms b - \ms a ) }{ \width + 1 }, 0 \bigr\} \bigr) } \\
    &\le \smallsum_{i=1}^{\width} \babs{ \max \bigl\{ x_1 - \ms a - \tfrac{ i ( \ms b - \ms a ) }{ \width + 1 }, 0 \bigr\}
      - \max \bigl\{ y_1 - \ms a - \tfrac{ i ( \ms b - \ms a ) }{ \width + 1 }, 0 \bigr\} } \\
    &\le \smallsum_{i=1}^{\width} \babs{ \bigl[ x_1 - \ms a - \tfrac{ i ( \ms b - \ms a ) }{ \width + 1 } \bigr]
      - \bigl[ y_1 - \ms a - \tfrac{ i ( \ms b - \ms a ) }{ \width + 1 } \bigr] } \\
    &= \smallsum_{i=1}^{\width} \abs{ x_1 - y_1 }
    = \width \abs{ x_1 - y_1 }
    \le \width \eucnorm{ x - y }.
  \end{split}
  \end{equation}
  This and \cref{eq:thm:bound:range:absolute2} \prove that for all $ n \in \N $ it holds that
  \begin{equation}\label{eq:thm:bound:range:lipnorm2}
  \begin{split}
    \lipnorm{ \mc N^{\theta(n)} }{ A }
    &= \inf\nolimits_{ x \in A } \abs{ \mc N^{\theta(n)} (x) }
      + \sup\nolimits_{ x, y \in [\ms a, \ms b]^{\inp}, \, x \neq y } \tfrac{ \abs{ \mc N^{\theta(n)} (x) - \mc N^{\theta(n)} (y) } }
        { \eucnorm{ x - y } } \\
    &\le n + \width ( \ms b - \ms a ) + \width
    = n + \width ( \ms b - \ms a + 1 ).
  \end{split}
  \end{equation}
  \Moreover \cref{eq:thm:bound:range:realization2} and \cref{lem:output:bias} (applied for every $ n \in \N $,
  $ \vartheta \in \{ \eta \in \R^{\dimension} \colon \mc N^{\eta} = \mc N^{\theta(n)} \} $ with
  $ c \is n $, $ \theta \is \vartheta $
  in the notation of \cref{lem:output:bias})
  \prove that for all $ n \in \N $,
  $ \vartheta = ( \vartheta_1, \ldots, \vartheta_{\dimension} ) \in \{ \eta \in \R^{\dimension} \colon \mc N^{\eta} = \mc N^{\theta(n)} \} $
  it holds that
  \begin{equation}
    \eucnorm{ \vartheta }
    \ge \abs{ \vartheta_{\dimension} }
    = n.
  \end{equation}
  The fact that for all $ \ms c \in \R $ it holds that $ \lim_{ n \to \infty } n ( 1 - \ms c n^{ \delta - 1 } ) = \infty $,
  the fact that for all $ x, y \in [0,\infty) $ it holds that $ ( x + y )^{\delta} \le x^{\delta} + y^{\delta} $, and
  \cref{eq:thm:bound:range:lipnorm2} \hence \prove that for all $ \ms c \in [0,\infty) $ there exists $ n \in \N $ such that for all
  $ \vartheta \in \{ \eta \in \R^{\dimension} \colon \mc N^{\eta} = \mc N^{\theta(n)} \} $ it holds that
  $ \lipnorm{ \mc N^{\theta(n)} }{ A } \ge 1 $ and
  \begin{equation}
  \begin{split}
    \eucnorm{ \vartheta }
    &\ge n
    = \ms c n^{\delta} + n ( 1 - \ms c n^{ \delta - 1 } )
    \ge \ms c n^{\delta} + \ms c \bigl[ \width ( \ms b - \ms a + 1 ) \bigr]^{\delta}
    = \ms c \bigl( n^{\delta} + \bigl[ \width ( \ms b - \ms a + 1 ) \bigr]^{\delta} \bigr) \\
    &\ge \ms c \bigl[ n + \width ( \ms b - \ms a + 1 ) \bigr]^{\delta}
    \ge \ms c \lipnorm{ \mc N^{\theta(n)} }{ A }^{\delta}
    = \ms c \max \bigl\{ \lipnorm{ \mc N^{\theta(n)} }{ A }^{\varepsilon}, \lipnorm{ \mc N^{\theta(n)} }{ A }^{\delta} \bigr\}.
  \end{split}
  \end{equation}
  \Hence that for all $ \ms c \in \R $ there exists $ \theta \in \R^{\dimension} $ such that for all
  $ \vartheta \in \{ \eta \in \R^{\dimension} \colon \mc N^{\eta} = \mc N^{\theta} \} $ it holds that
  \begin{equation}
    \eucnorm{ \vartheta }
    > \ms c \max \bigl\{ \lipnorm{ \mc N^{\theta} }{ A }^{\varepsilon}, \lipnorm{ \mc N^{\theta} }{ A }^{\delta} \bigr\}.
  \end{equation}
  This \proves \cref{eq:thm:bound:range} in the case $ \delta < 1 $.
\end{cproof}

\cfclear
\begin{cor}\label{cor:lipnorm:range:negative}
  Assume \cref{setting:main} and let $ n \in \N $, $ \delta_1, \delta_2, \ldots, \delta_n \in [0,\infty) $,
  $ A \subseteq [\ms a, \ms b]^{\inp} $ satisfy $ [ \min \{ \delta_1, \delta_2, \ldots, \delta_n \} > \nicefrac{1}{2} ]
    \vee [ \max \{ \delta_1, \delta_2, \ldots, \delta_n \} < 1 ] $ and $ A \neq \emptyset $.
  Then for all $ \ms c \in \R $ there exists $ \theta \in \R^{\dimension} $ such that for all
  $ \vartheta \in \{ \eta \in \R^{\dimension} \colon \mc N^{\eta} = \mc N^{\theta} \} $ it holds that
  \begin{equation}\label{eq:cor:lipnorm:range:negative}
    \eucnorm{ \vartheta } > \ms c \Bigl( \smallsum\nolimits_{i=1}^n \lipnorm{ \mc N^{\theta} }{ A }^{ \delta_i } \Bigr)
  \end{equation}
  \cfout.
\end{cor}

\begin{cproof}{cor:lipnorm:range:negative}
  Throughout this proof let $ i, j \in \{ 1, 2, \ldots, n \} $ satisfy that
  \begin{equation}
    \delta_i = \min \{ \delta_1, \delta_2, \ldots, \delta_n \}
    \qqandqq
    \delta_j = \max \{ \delta_1, \delta_2, \ldots, \delta_n \}.
  \end{equation}
  \Nobs that the assumption that $ [ \min \{ \delta_1, \delta_2, \ldots, \delta_n \} > \nicefrac{1}{2} ]
    \vee [ \max \{ \delta_1, \delta_2, \ldots, \delta_n \} < 1 ] $ \proves that $ [ \delta_i > \nicefrac{1}{2} ] \vee [ \delta_j < 1 ] $.
  \cref{thm:bound:range} \hence \proves that for all $ \ms c \in [0,\infty) $ there exists $ \theta \in \R^{\dimension} $ such that for all
  $ \vartheta \in \{ \eta \in \R^{\dimension} \colon \mc N^{\eta} = \mc N^{\theta} \} $ it holds that
  \begin{equation}
    \ms c \Bigl( \smallsum\nolimits_{k=1}^n \lipnorm{ \mc N^{\theta} }{ A }^{ \delta_k } \Bigr)
    \le \ms c n \max\nolimits_{ k \in \{ 1, 2, \ldots, n \} } \lipnorm{ \mc N^{\theta} }{ A }^{ \delta_k }
    \le \ms c n \max\nolimits \bigr\{ \lipnorm{ \mc N^{\theta} }{ A }^{ \delta_i }, \lipnorm{ \mc N^{\theta} }{ A }^{ \delta_j } \bigr\}
    < \eucnorm{ \vartheta }
  \end{equation}
  \cfload.
  This \proves that for all $ \ms c \in \R $ there exists $ \theta \in \R^{\dimension} $ such that for all
  $ \vartheta \in \{ \eta \in \R^{\dimension} \colon \mc N^{\eta} = \mc N^{\theta} \} $ it holds that
  \begin{equation}
    \eucnorm{ \vartheta } > \ms c \Bigl( \smallsum\nolimits_{k=1}^n \lipnorm{ \mc N^{\theta} }{ A }^{ \delta_k } \Bigr).
  \end{equation}
\end{cproof}

\section{Lower bounds for norms of reparameterized ANNs using \allowbreak Hölder norms and Sobolev-Slobodeckij norms}
\label{sec:lower_bounds_hoelder_norms_sob-slob_norms}

In this section, we consider different norms for the realization function than Lipschitz norms, and we prove, in \cref{cor:hoelder:sob:sum} in Subsection~\ref{subsec:lower_bounds_hoelder_norms_sob-slob_norms} below, lower bounds for norms of reparameterized \ANN\ parameter vectors using Hölder norms and Sobolev-Slobodeckij norms. As a consequence, \cref{cor:hoelder:sob:sum} implies that it is not possible to control the norm of reparameterized \ANNs\ using sums of powers of Hölder norms of the realization function or sums of powers of Sobolev-Slobodeckij norms of the realization function with arbitrary exponents. The proof of \cref{cor:hoelder:sob:sum} employs the lower bounds for reparameterized \ANNs\ demonstrated in \cref{thm:bound:hoelder:sob} in Subsection~\ref{subsec:lower_bounds_hoelder_norms_sob-slob_norms} and the well-known relationships between different Hölder norms and different Sobolev-Slobodeckij norms established in \cref{lem:hoeldernorm:inequality} and \cref{lem:sobnorm:inequality} in Subsection~\ref{subsec:hoelder_norms_sob-slob_norms} below, respectively. Only for completeness, we also include the detailed proofs of \cref{lem:hoeldernorm:inequality} and \cref{lem:sobnorm:inequality}. Moreover, we note that our proof of \cref{lem:sobnorm:inequality} makes use of the elementary integral results presented in \cref{lem:integral:gamma} and \cref{lem:double:integral:gamma} in Subsection~\ref{subsec:hoelder_norms_sob-slob_norms}.

For the convenience of the reader, we recall the notions of Hölder norms and Sobolev-Slobodeckij norms in \cref{def:hoelder_norm} and \cref{def:sobolev-slobodeckij_norm} in Subsection~\ref{subsec:hoelder_norms_sob-slob_norms}.

\subsection{Hölder norms and Sobolev-Slobodeckij norms}
\label{subsec:hoelder_norms_sob-slob_norms}

\cfclear
\begin{definition}\label{def:hoelder_norm}
  Let $ d \in \N $, $ \ms a \in \R $, $ \ms b \in (\ms a, \infty) $, $ \gamma \in [0,1] $, $ v \in [\ms a, \ms b] $ and let
  $ f \colon [\ms a, \ms b]^d \to \R $ be a function.
  Then we denote by $ \hoeldernorm{f}{\gamma}{v} \in [0,\infty] $ the extended real number given by
  \begin{equation}
    \hoeldernorm{f}{\gamma}{v} = \sup\nolimits_{ x \in [\ms a, v]^d } \abs{ f(x) }
      + \sup\nolimits_{ x, y \in [\ms a, \ms b]^d, \, x \neq y } \tfrac{ \abs{ f(x) - f(y) } }{ \eucnorm{ x - y }^{\gamma} }
  \end{equation}
  (cf.~\cref{def:scalar_product_norm}).
\end{definition}

\cfclear
\begin{definition}\label{def:sobolev-slobodeckij_norm}
  Let $ d \in \N $, $ \ms a \in \R $, $ \ms b \in (\ms a, \infty) $, $ \gamma \in [0,1] $, $ p \in [1,\infty) $ and let
  $ f \colon [\ms a, \ms b]^d \to \R $ be measurable.
  Then we denote by $ \sobnorm{f}{\gamma}{p} \in [0,\infty] $ the extended real number given by
  \begin{equation}
    \sobnorm{f}{\gamma}{p} = \biggl[ \int_{ [\ms a, \ms b]^d } \abs{ f(x) }^p \, \diff x \biggr]^{\nicefrac{1}{p}}
      + \biggl[ \int_{ [\ms a, \ms b]^d } \int_{ [\ms a, \ms b]^d }
        \tfrac{ \abs{ f(x) - f(y) }^p }{ \eucnorm{ x - y }^{\gamma p + d} } \, \diff x \, \diff y \biggr]^{\nicefrac{1}{p}}
  \end{equation}
  (cf.~\cref{def:scalar_product_norm}).
\end{definition}

\cfclear
\begin{lemma}\label{lem:hoeldernorm:inequality}
  Let $ d \in \N $, $ \ms a \in \R $, $ \ms b \in (\ms a, \infty) $, $ \gamma, \lambda \in [0,1] $, $ v, w \in [\ms a, \ms b] $ satisfy
  $ \gamma \le \lambda $ and $ v \le w $.
  Then for all functions $ f \colon [\ms a, \ms b]^d \to \R $ it holds that
  \begin{equation}
    \hoeldernorm{ f }{ \gamma }{ v }
    \le \max \bigl\{ 1, \bigl[ d^{\nicefrac{1}{2}} (\ms b - \ms a) \bigr]^{ \lambda - \gamma } \bigr\} \hoeldernorm{ f }{ \lambda }{ w }
  \end{equation}
  \cfout.
\end{lemma}

\begin{cproof}{lem:hoeldernorm:inequality}
  \Nobs that the fact that for all $ x, y \in [\ms a, \ms b]^d $ it holds that
  $ \eucnorm{ x - y } \le d^{\nicefrac{1}{2}} ( \ms b - \ms a ) $ and the assumption that $ \gamma \le \lambda $ and $ v \le w $
  \prove that for all functions $ f \colon [\ms a, \ms b]^d \to \R $ it holds that
  \begin{equation}
  \begin{split}
    \hoeldernorm{ f }{ \gamma }{ v }
    &= \sup\nolimits_{ x \in [\ms a, v]^d } \abs{ f(x) }
      + \sup\nolimits_{ x, y \in [\ms a, \ms b]^d, \, x \neq y } \tfrac{ \abs{ f(x) - f(y) } }{ \eucnorm{ x - y }^{\gamma} } \\
    &= \sup\nolimits_{ x \in [\ms a, v]^d } \abs{ f(x) }
      + \sup\nolimits_{ x, y \in [\ms a, \ms b]^d, \, x \neq y } \tfrac{ \abs{ f(x) - f(y) } }{ \eucnorm{ x - y }^{\lambda} }
      \eucnorm{ x - y }^{ \lambda - \gamma } \\
    &\le \sup\nolimits_{ x \in [\ms a, w]^d } \abs{ f(x) }
      + \bigl[ d^{\nicefrac{1}{2}} ( \ms b - \ms a ) \bigr]^{ \lambda - \gamma }
      \sup\nolimits_{ x, y \in [\ms a, \ms b]^d, \, x \neq y } \tfrac{ \abs{ f(x) - f(y) } }{ \eucnorm{ x - y }^{\lambda} } \\
    &\le \max \bigl\{ 1, \bigl[ d^{\nicefrac{1}{2}} ( \ms b - \ms a ) \bigr]^{ \lambda - \gamma } \bigr\}
      \hoeldernorm{ f }{ \lambda }{ w }
  \end{split}
  \end{equation}
  \cfload.
\end{cproof}

\cfclear
\begin{lemma}\label{lem:integral:gamma}
  Let $ d \in \N $, $ r \in (0,\infty) $, $ \gamma \in (-d,\infty) $ and let $ \Gamma \colon (0,\infty) \to \R $ satisfy for all
  $ x \in (0,\infty) $ that $ \Gamma(x) = \int_0^{\infty} t^{x-1} e^{-t} \, \diff t $.
  Then
  \begin{equation}
    \int_{ \{ y \in \R^d \colon \eucnorm{ y } \le r \} } \eucnorm{ x }^{\gamma} \, \diff x
    = \tfrac{ 2 \pi^{ \nicefrac{d}{2} } }{ (d + \gamma) \Gamma( \nicefrac{d}{2} ) } r^{ d + \gamma }
  \end{equation}
  \cfout.
\end{lemma}

\begin{cproof}{lem:integral:gamma}
  Throughout this proof let $ \mc S \colon \mc B( \R^d ) \to [0,\infty] $ be the $ (d - 1) $-dimensional spherical measure.
  \Nobs that the coarea formula and the fact that for all $ t \in (0,\infty) $ it holds that
  $ \mc S ( \{ y \in \R^d \colon \eucnorm{ y } = t \} ) = 2 \pi^{ \nicefrac{d}{2} } [ \Gamma( \nicefrac{d}{2} ) ]^{-1} t^{d-1} $ \prove that
  \begin{equation}
  \begin{split}
    &\int_{ \{ y \in \R^d \colon \eucnorm{ y } \le r \} } \eucnorm{ x }^{\gamma} \, \diff x
    = \int_{ \R^d } \eucnorm{ x }^{\gamma} \ind_{ [0,r] } ( \eucnorm{ x } ) \, \diff x \\
    &= \int_0^{\infty} \int_{ \{ y \in \R^d \colon \eucnorm{ y } = t \} } \eucnorm{ x }^{\gamma} \ind_{ [0,r] } ( \eucnorm{ x } )
      \, \mc S( \diff x ) \, \diff t \\
    &= \int_0^{\infty} \int_{ \{ y \in \R^d \colon \eucnorm{ y } = t \} } t^{\gamma} \ind_{ [0,r] } (t) \, \mc S( \diff x ) \, \diff t \\
    &= \int_0^r t^{\gamma} \mc S \bigr( \{ y \in \R^d \colon \eucnorm{ y } = t \} \bigl) \, \diff t
    = \tfrac{ 2 \pi^{ \nicefrac{d}{2} } }{ \Gamma( \nicefrac{d}{2} ) } \int_0^r t^{d+\gamma-1} \, \diff t \\
    &= \tfrac{ 2 \pi^{ \nicefrac{d}{2} } }{ \Gamma( \nicefrac{d}{2} ) } \Bigl[ \tfrac{ t^{ d + \gamma } }{ d + \gamma } \Bigr]^r_0
    = \tfrac{ 2 \pi^{ \nicefrac{d}{2} } }{ \Gamma( \nicefrac{d}{2} ) } \tfrac{ r^{d+\gamma} }{ d + \gamma }
    = \tfrac{ 2 \pi^{ \nicefrac{d}{2} } }{ (d + \gamma) \Gamma( \nicefrac{d}{2} ) } r^{ d + \gamma }
  \end{split}
  \end{equation}
  \cfload.
\end{cproof}

\cfclear
\begin{lemma}\label{lem:double:integral:gamma}
  Let $ d \in \N $, $ \ms a \in \R $, $ \ms b \in (\ms a, \infty) $, $ \gamma \in (-d,\infty) $ and let
  $ \Gamma \colon (0,\infty) \to \R $ satisfy for all $ x \in (0,\infty) $ that $ \Gamma(x) = \int_0^{\infty} t^{x-1} e^{-t} \, \diff t $.
  Then
  \begin{equation}
    \int_{ [\ms a, \ms b]^d } \int_{ [\ms a, \ms b]^d } \eucnorm{ x - y }^{\gamma} \, \diff x \, \diff y
    \le \tfrac{ 2 \pi^{\nicefrac{d}{2}} d^{ \nicefrac{ (d + \gamma) }{ 2 } } (\ms b - \ms a)^{ 2 d + \gamma } }
      { (d + \gamma) \Gamma( \nicefrac{d}{2} ) }
  \end{equation}
  \cfout.
\end{lemma}

\begin{cproof}{lem:double:integral:gamma}
  \Nobs that the fact that for all $ x, y \in [\ms a, \ms b]^d $ it holds that $ \eucnorm{ x - y } \le d^{\nicefrac{1}{2}} (\ms b - \ms a) $
  and \cref{lem:integral:gamma} (applied with
  $ d \is d $, $ r \is d^{\nicefrac{1}{2}} (\ms b - \ms a) $, $ \gamma \is \gamma $
  in the notation of \cref{lem:integral:gamma})
  \prove that for all $ y \in [\ms a, \ms b]^d $ it holds that
  \begin{equation}
  \begin{split}
    \int_{ [\ms a, \ms b]^d } \eucnorm{ x - y }^{\gamma} \, \diff x
    &= \int_{ \smallbigcup_{ x \in [\ms a, \ms b]^d } \{ x - y \} } \eucnorm{ z }^{\gamma} \, \diff z
    \le \int_{ \{ x \in \R^d \colon \eucnorm{ x } \le d^{\nicefrac{1}{2}} ( \ms b - \ms a ) \} } \eucnorm{ z }^{\gamma} \, \diff z \\
    &= \tfrac{ 2 \pi^{\nicefrac{d}{2}} }{ (d + \gamma) \Gamma( \nicefrac{d}{2} ) }
      \bigl[ d^{\nicefrac{1}{2}} (\ms b - \ms a) \bigr]^{ d + \gamma }
    = \tfrac{ 2 \pi^{\nicefrac{d}{2}} d^{ \nicefrac{ (d + \gamma) }{ 2 } } (\ms b - \ms a)^{ d + \gamma } }
      { (d + \gamma) \Gamma( \nicefrac{d}{2} ) }
  \end{split}
  \end{equation}
  \cfload.
  \Hence that
  \begin{equation}
    \int_{ [\ms a, \ms b]^d } \int_{ [\ms a, \ms b]^d } \eucnorm{ x - y }^{\gamma} \, \diff x \, \diff y
    \le \int_{ [\ms a, \ms b]^d } \tfrac{ 2 \pi^{\nicefrac{d}{2}} d^{ \nicefrac{ (d + \gamma) }{ 2 } } (\ms b - \ms a)^{ d + \gamma } }
      { (d + \gamma) \Gamma( \nicefrac{d}{2} ) } \, \diff y
    = \tfrac{ 2 \pi^{\nicefrac{d}{2}} d^{ \nicefrac{ (d + \gamma) }{ 2 } } (\ms b - \ms a)^{ 2 d + \gamma } }
      { (d + \gamma) \Gamma( \nicefrac{d}{2} ) }.
  \end{equation}
\end{cproof}

\cfclear
\begin{lemma}\label{lem:sobnorm:inequality}
  Let $ d \in \N $, $ \ms a \in \R $, $ \ms b \in (\ms a, \infty) $, $ \gamma, \lambda \in [0,1] $, $ p, q \in [1,\infty) $ satisfy
  $ \gamma < \lambda $ and $ p < q $ and let $ \Gamma \colon (0,\infty) \to \R $ satisfy for all $ x \in (0,\infty) $ that
  $ \Gamma(x) = \int_0^{\infty} t^{x-1} e^{-t} \, \diff t $.
  Then for all measurable functions $ f \colon [\ms a, \ms b]^d \to \R $ it holds that
  \begin{equation}
    \sobnorm{ f }{ \gamma }{ p }
    \le \Bigl[ \max \Bigl\{ ( \ms b - \ms a )^{\inp}, \tfrac{ 2 \pi^{d/2} d^{ ( \lambda - \gamma ) qp / 2(q-p) }
        (\ms b - \ms a)^{ d + ( \lambda - \gamma ) qp / (q-p) }  (q-p) }{ ( \lambda - \gamma ) qp \Gamma( d/2 ) } \Bigr\}
        \Bigr]^{\nicefrac{(q-p)}{qp}} \sobnorm{ f }{ \lambda }{ q }
  \end{equation}
  \cfout.
\end{lemma}

\begin{cproof}{lem:sobnorm:inequality}
  \Nobs that the Hölder inequality and the assumption that $ p < q $ \prove that for all measurable functions
  $ f \colon [\ms a, \ms b]^d \to \R $ it holds that
  \begin{equation}\label{eq:lem:sobnorm:inequality:1}
  \begin{split}
    \biggl[ \int_{ [\ms a, \ms b]^d } \abs{ f(x) }^p \, \diff x \biggr]^{\nicefrac{1}{p}}
    &\le \Biggl( \biggl[ \int_{ [\ms a, \ms b]^d } \Bigl[ \abs{ f(x) }^p \Bigr]^{ \nicefrac{q}{p} } \, \diff x \biggr]^{ \nicefrac{p}{q} }
      \biggl[ \int_{ [\ms a, \ms b]^d } \bigl[ \, 1 \, \bigr]^{ \nicefrac{q}{(q-p)} } \, \diff x \biggr]^{ \nicefrac{(q-p)}{q} }
      \Biggr)^{\nicefrac{1}{p}} \\
    &= \biggl[ \int_{ [\ms a, \ms b]^d } \abs{ f(x) }^q \, \diff x \biggr]^{\nicefrac{1}{q}}
      \Bigl[ ( \ms b - \ms a )^d \Bigr]^{ \nicefrac{(q-p)}{qp} } \\
    &= \bigl( \ms b - \ms a \bigr)^{ \nicefrac{ d(q-p) }{ qp } }
      \biggl[ \int_{ [\ms a, \ms b]^d } \abs{ f(x) }^q \, \diff x \biggr]^{\nicefrac{1}{q}}.
  \end{split}
  \end{equation}
  \Moreover \cref{lem:double:integral:gamma} (applied with
  $ d \is \inp $, $ \ms a \is \ms a $, $ \ms b \is \ms b $, $ \gamma \is ( \lambda - \gamma ) \nicefrac{ qp }{ (q-p) } - d $
  in the notation of \cref{lem:double:integral:gamma})
  and the assumption that $ \gamma < \lambda $ and $ p < q $ \prove that
  \begin{equation}
    \int_{ [\ms a, \ms b]^d } \int_{ [\ms a, \ms b]^d }
      \eucnorm{ x - y }^{ ( \lambda - \gamma ) qp / (q-p) - d } \, \diff x \, \diff y
    \le \tfrac{ 2 \pi^{d/2} d^{ ( \lambda - \gamma ) qp / 2(q-p) } (\ms b - \ms a)^{ d + ( \lambda - \gamma ) qp / (q-p) }  (q-p) }
      { ( \lambda - \gamma ) qp \Gamma( d/2 ) }
  \end{equation}
  \cfload.
  Combining this, the Hölder inequality, and the assumption that $ p < q $ \proves that for all measurable functions
  $ f \colon [\ms a, \ms b]^d \to \R $ it holds that
  \begin{equation}
  \begin{split}
    &\biggl[ \int_{ [\ms a, \ms b]^d } \int_{ [\ms a, \ms b]^d }
      \tfrac{ \abs{ f(x) - f(y) }^p }{ \eucnorm{ x - y }^{ \gamma p + d } } \, \diff x \, \diff y \biggr]^{\nicefrac{1}{p}} \\
    &= \biggl[ \int_{ [\ms a, \ms b]^d } \int_{ [\ms a, \ms b]^d }
      \tfrac{ \abs{ f(x) - f(y) }^p }{ \eucnorm{ x - y }^{ \lambda p + dp/q } }
      \eucnorm{ x - y }^{ \lambda p - \gamma p + dp/q - d } \, \diff x \, \diff y \biggr]^{\nicefrac{1}{p}} \\
    &\le \Biggl( \biggl[ \int_{ [\ms a, \ms b]^d } \int_{ [\ms a, \ms b]^d }
      \Bigl[ \tfrac{ \abs{ f(x) - f(y) }^p }{ \eucnorm{ x - y }^{ \lambda p + dp/q } } \Bigr]^{\nicefrac{q}{p}}
      \, \diff x \, \diff y \biggr]^{\nicefrac{p}{q}} \\
    &\qquad \times \biggl[ \int_{ [\ms a, \ms b]^d } \int_{ [\ms a, \ms b]^d }
      \Bigl[ \eucnorm{ x - y }^{ \lambda p - \gamma p + dp/q - d } \Bigr]^{\nicefrac{q}{(q-p)}} \, \diff x \, \diff y
      \biggr]^{\nicefrac{(q-p)}{q}} \Biggr)^{\nicefrac{1}{p}} \\
    &= \biggl[ \int_{ [\ms a, \ms b]^d } \int_{ [\ms a, \ms b]^d }
      \tfrac{ \abs{ f(x) - f(y) }^q }{ \eucnorm{ x - y }^{ \lambda q + d } } \, \diff x \, \diff y \biggr]^{\nicefrac{1}{q}}
      \biggl[ \int_{ [\ms a, \ms b]^d } \int_{ [\ms a, \ms b]^d }
      \eucnorm{ x - y }^{ ( \lambda - \gamma ) qp / (q-p) - d } \, \diff x \, \diff y \biggr]^{\nicefrac{(q-p)}{qp}} \\
    &\le \biggl[ \int_{ [\ms a, \ms b]^d } \int_{ [\ms a, \ms b]^d }
      \tfrac{ \abs{ f(x) - f(y) }^q }{ \eucnorm{ x - y }^{ \lambda q + d } } \, \diff x \, \diff y \biggr]^{\nicefrac{1}{q}}
      \Bigl[ \tfrac{ 2 \pi^{d/2} d^{ ( \lambda - \gamma ) qp / 2(q-p) } (\ms b - \ms a)^{ d + ( \lambda - \gamma ) qp / (q-p) }  (q-p) }
        { ( \lambda - \gamma ) qp \Gamma( d/2 ) } \Bigr]^{\nicefrac{(q-p)}{qp}}.
  \end{split}
  \end{equation}
  This and \cref{eq:lem:sobnorm:inequality:1} \prove that for all measurable functions $ f \colon [\ms a, \ms b]^d \to \R $ it holds that
  \begin{equation}
  \begin{split}
    &\sobnorm{ f }{ \gamma }{ p }
    = \biggl[ \int_{ [\ms a, \ms b]^d } \abs{ f(x) }^p \biggr]^{\nicefrac{1}{p}}
      + \biggl[ \int_{ [\ms a, \ms b]^d } \int_{ [\ms a, \ms b]^d }
        \tfrac{ \abs{ f(x) - f(y) }^p }{ \eucnorm{ x - y }^{ \gamma p + d } } \biggr]^{\nicefrac{1}{p}} \\
    &\le \bigl( \ms b - \ms a \bigr)^{ \nicefrac{ d(q-p) }{ qp } }
      \biggl[ \int_{ [\ms a, \ms b]^d } \abs{ f(x) }^q \, \diff x \biggr]^{\nicefrac{1}{q}} \\
    &\quad + \Bigl[ \tfrac{ 2 \pi^{d/2} d^{ ( \lambda - \gamma ) qp / 2(q-p) } (\ms b - \ms a)^{ d + ( \lambda - \gamma ) qp / (q-p) }  (q-p) }
        { ( \lambda - \gamma ) qp \Gamma( d/2 ) } \Bigr]^{\nicefrac{(q-p)}{qp}}
      \biggl[ \int_{ [\ms a, \ms b]^d } \int_{ [\ms a, \ms b]^d }
      \tfrac{ \abs{ f(x) - f(y) }^q }{ \eucnorm{ x - y }^{ \lambda q + d } } \, \diff x \, \diff y \biggr]^{\nicefrac{1}{q}} \\
    &\le \max \biggl\{ \bigl( \ms b - \ms a \bigr)^{ \nicefrac{ d(q-p) }{ qp } },
        \Bigl[ \tfrac{ 2 \pi^{d/2} d^{ ( \lambda - \gamma ) qp / 2(q-p) } (\ms b - \ms a)^{ d + ( \lambda - \gamma ) qp / (q-p) }  (q-p) }
        { ( \lambda - \gamma ) qp \Gamma( d/2 ) } \Bigr]^{\nicefrac{(q-p)}{qp}} \biggr\} \sobnorm{ f }{ \lambda }{ q } \\
    &= \Bigl[ \max \Bigl\{ ( \ms b - \ms a )^{\inp}, \tfrac{ 2 \pi^{d/2} d^{ ( \lambda - \gamma ) qp / 2(q-p) }
        (\ms b - \ms a)^{ d + ( \lambda - \gamma ) qp / (q-p) }  (q-p) }{ ( \lambda - \gamma ) qp \Gamma( d/2 ) } \Bigr\}
        \Bigr]^{\nicefrac{(q-p)}{qp}} \sobnorm{ f }{ \lambda }{ q }
  \end{split}
  \end{equation}
  \cfload.
\end{cproof}

\subsection{Lower bounds for norms of reparameterized ANNs using Hölder norms and Sobolev-Slobodeckij norms}
\label{subsec:lower_bounds_hoelder_norms_sob-slob_norms}

\cfclear
\begin{theorem}\label{thm:bound:hoelder:sob}
  Assume \cref{setting:main} and let $ \gamma \in [0,1) $, $ p \in [1,\infty) $.
  Then for all $ \ms c \in (0,\infty) $ there exists $ \theta \in \R^{\dimension} $ such that for all
  $ \vartheta \in \{ \eta \in \R^{\dimension} \colon \mc N^{\eta} = \mc N^{\theta} \} $ it holds that
  \begin{equation}\label{eq:thm:bound:hoelder:sob}
    \eucnorm{ \vartheta } > \ms c
    \qqandqq
    \max \bigl\{ \hoeldernorm{ \mc N^{\theta} }{\gamma}{\ms b}, \sobnorm{ \mc N^{\theta} }{\gamma}{p} \bigr\} < \ms c^{-1}.
  \end{equation}
  \cfout.
\end{theorem}

\begin{cproof}{thm:bound:hoelder:sob}
  Throughout this proof let $ q \in (0,\infty) $ satisfy
  \begin{equation}\label{eq:thm:bound:hoelder:sob:q}
    \gamma q < 1 - \gamma
    \qqandqq
    ( p - \inp ) q < d,
  \end{equation}
  let $ \theta = ( \theta_1, \ldots, \theta_{\dimension} ) \colon \N \to \R^{\dimension} $ satisfy for all
  $ n \in \N $, $ i \in \{ 2, 3, \ldots, \width \} $, $ j \in \{ 1, 2, \ldots, \inp \} $ that
  \begin{equation}\label{eq:thm:bound:hoelder:sob:theta}
    \theta_j (n) = n^q, \qquad
    \theta_{ \inp \width + 1 } (n) = n^{-1} - n^q \inp \ms b, \qquad
    \theta_{ \inp \width + \width + 1 } (n) = 1,
  \end{equation}
  and
  $ \theta_{ (i-1) \inp + j } (n) = \theta_{ \inp \width + i } (n) = \theta_{ \inp \width + \width + i } (n) = \theta_{\dimension} (n) = 0 $,
  let $ u = ( 1, 1, \ldots, 1 ) $, $ v = ( \ms b, \ms b, \ldots, \ms b ) \in \R^d $,
  let $ \varepsilon_n \in (0,\infty) $, $ n \in \N $, and $ A_n \subseteq [\ms a, \ms b]^{\inp} $, $ n \in \N $, satisfy for all
  $ n \in \N $ that
  \begin{equation}\label{eq:thm:bound:hoelder:sob:varepsilon_A}
    \varepsilon_n = \min\{ n^{-1-q} \inp^{-1}, \ms b - \ms a \}
    \qqandqq
    A_n = \bigl\{ x \in [\ms a, \ms b]^{\inp} \colon n^q \scalprod{ u }{ x } + n^{-1} - n^q \inp \ms b \ge 0 \bigr\},
  \end{equation}
  and let $ \Gamma \colon (0,\infty) \to \R $ satisfy for all $ x \in (0,\infty) $ that
  $ \Gamma(x) = \int_0^{\infty} t^{x-1} e^{-t} \, \diff t $
  \cfload.
  \Nobs that \cref{eq:thm:bound:hoelder:sob:theta} \proves that for all $ n \in \N $,
  $ x = ( x_1, \ldots, x_{\inp} ) \in [\ms a, \ms b]^{\inp} $ it holds that
  \begin{equation}\label{eq:thm:bound:hoelder:sob:realization:theta}
  \begin{split}
    \mc N^{\theta(n)} (x)
    &= \theta_{\dimension}(n) + \smallsum_{i=1}^{\width} \theta_{ \inp \width + \width + i }(n)
      \max \bigl\{ \theta_{ \inp \width + i }(n) + \smallsum_{j=1}^{\inp} \theta_{ (i-1) \inp + j }(n) x_j, 0 \bigr\} \\
    &= \max \bigl\{ n^{-1} - n^q \inp \ms b + \smallsum_{j=1}^{\inp} n^q x_j, 0 \bigr\}
    = \max \bigl\{ n^q \scalprod{ u }{ x } + n^{-1} - n^q \inp \ms b, 0 \bigr\}.
  \end{split}
  \end{equation}
  \Moreover for all $ n \in \N $ it holds that
  \begin{equation}
  \begin{split}
    n^q \scalprod{ u }{ v - \varepsilon_n u } + n^{-1} - n^q \inp \ms b
    &= n^q \scalprod{ u }{ v } - \varepsilon_n n^q \scalprod{ u }{ u } + n^{-1} - n^q \inp \ms b \\
    &= n^{-1} - \varepsilon_n n^q \inp
    \ge n^{-1} - n^{-1} = 0.
  \end{split}
  \end{equation}
  The fact that for all $ n \in \N $ it holds that $ \ms b - \varepsilon_n \in [\ms a, \ms b] $ \hence \proves that for all $ n \in \N $
  it holds that $ v - \varepsilon_n u \in A_n $.
  Combining this with \cref{lem:lipschitzconstant:bound} and \cref{eq:thm:bound:hoelder:sob:realization:theta} \proves that for all
  $ n \in \N $, $ \vartheta \in \{ \eta \in \R^{\dimension} \colon \mc N^{\eta} = \mc N^{\theta(n)} \} $ it holds that
  \begin{equation}
  \begin{split}
    \tfrac{1}{2} \eucnorm{ \vartheta }^2
    &\ge \sup\nolimits_{ x, y \in [\ms a, \ms b]^{\inp}, \, x \neq y } \tfrac{ \abs{ \mc N^{\vartheta}(x) - \mc N^{\vartheta}(y) } }
      { \eucnorm{ x - y } }
    \ge \abs{ \mc N^{\vartheta} ( v ) - \mc N^{\vartheta} ( v - \varepsilon_n u ) } \bigl[ \varepsilon_n \eucnorm{u} \bigr]^{-1} \\
    &= \abs{ \mc N^{\theta(n)} ( v ) - \mc N^{\theta(n)} ( v - \varepsilon_n u ) } [ \varepsilon_n ]^{-1} \inp^{-\nicefrac{1}{2}} \\
    &= \bigl( n^{-1} - n^{-1} + \varepsilon_n n^q \inp \bigr) [ \varepsilon_n ]^{-1} \inp^{-\nicefrac{1}{2}} 
    = \inp^{\nicefrac{1}{2}} n^q.
  \end{split}
  \end{equation}
  \Hence that for all $ n \in \N $,
  $ \vartheta \in \{ \eta \in \R^{\dimension} \colon \mc N^{\eta} = \mc N^{\theta(n)} \} $ it holds that
  \begin{equation}\label{eq:lem:counterpart:norm:vartheta}
    \eucnorm{ \vartheta } \ge 2^{ \nicefrac{1}{2} } \inp^{\nicefrac{1}{4}} n^{ \nicefrac{q}{2} }.
  \end{equation}
  Next \nobs that \cref{eq:thm:bound:hoelder:sob:realization:theta} and the fact that for all $ x \in [\ms a, \ms b]^d $ it holds that
  $ \scalprod{ u }{ x } \le \inp \ms b $ \prove that for all $ n \in \N $, $ x \in [\ms a, \ms b]^{\inp} $ it holds that
  \begin{equation}\label{eq:thm:bound:hoelder:sob:realization:bound}
    \abs{ \mc N^{\theta(n)} (x) }
    = \babs{ \max \bigl\{ n^q \scalprod{ u }{ x } + n^{-1} - n^q \inp b, 0 \bigr\} }
    \le \abs{ \max\{ n^{-1}, 0 \} }
    = n^{-1}.
  \end{equation}
  \Moreover \cref{eq:thm:bound:hoelder:sob:realization:theta}, the fact that for all $ x, y \in \R $ it holds that
  $ \abs{ \max \{ x, 0 \} - \max \{ y, \allowbreak 0 \} } \le \abs{ x - y } $, and the Cauchy Schwarz inequality
  \prove that for all $ n \in \N $, $ x, y \in [\ms a, \ms b]^{\inp} $ it holds that
  \begin{equation}\label{eq:thm:bound:hoelder:sob:realization:diff}
  \begin{split}
    &\abs{ \mc N^{\theta(n)} (x) - \mc N^{\theta(n)} (y) } \\
    &= \babs{ \max \bigl\{ n^q \scalprod{ u }{ x } + n^{-1} - n^q \inp b, 0 \bigr\}
      - \max \bigl\{ n^q \scalprod{ u }{ y } + n^{-1} - n^q \inp b, 0 \bigr\} } \\
    &\le \babs{ [ n^q \scalprod{ u }{ x } + n^{-1} - n^q \inp b ] - [ n^q \scalprod{ u }{ y } + n^{-1} - n^q \inp b ] } \\
    &= \abs{ n^q \scalprod{ u }{ x } - n^q \scalprod{ u }{ y } }
    = n^q \abs{ \scalprod{ u }{ x - y } }
    \le n^q \eucnorm{ u } \eucnorm{ x - y }
    = n^q \inp^{\nicefrac{1}{2}} \eucnorm{ x - y }.
  \end{split}
  \end{equation}
  \Moreover \cref{eq:thm:bound:hoelder:sob:varepsilon_A} \proves that for all $ n \in \N $, $ x = ( x_1, \ldots, x_{\inp} ) \in A_n $
  it holds that
  \begin{equation}\label{eq:thm:bound:hoelder:sob:radius:A}
  \begin{split}
    \eucnorm{ x - v }
    &\le \smallsum_{j=1}^{\inp} \abs{ x_i - \ms b }
    = \smallsum_{j=1}^{\inp} ( \ms b - x_i )
    = \inp \ms b - \scalprod{ u }{ x } \\
    &= - n^{-q} \bigl( n^q \scalprod{ u }{ x } - n^q \inp \ms b \bigr)
    \le n^{-q} n^{-1}
    = n^{-1-q}.
  \end{split}
  \end{equation}
  Combining this and \cref{eq:thm:bound:hoelder:sob:realization:diff} \proves that for all $ n \in \N $, $ x, y \in A_n $
  with $ x \neq y $ it holds that
  \begin{equation}\label{eq:thm:bound:hoelder:sob:sup:1}
  \begin{split}
    \tfrac{ \abs{ \mc N^{\theta(n)} (x) - \mc N^{\theta(n)} (y) } }{ \eucnorm{ x - y }^{\gamma} }
    &\le n^q \inp^{\nicefrac{1}{2}} \tfrac{ \eucnorm{ x - y } }{ \eucnorm{ x - y }^{\gamma} }
    = n^q \inp^{\nicefrac{1}{2}} \eucnorm{ x - y }^{1-\gamma} \\
    &\le n^q \inp^{\nicefrac{1}{2}} \bigl[ \eucnorm{ x - v } + \eucnorm{ y - v } \bigr]^{1-\gamma}
    \le n^q \inp^{\nicefrac{1}{2}} \bigl[ 2 n^{-1-q} \bigr]^{1-\gamma} \\
    &= 2^{1-\gamma} \inp^{\nicefrac{1}{2}} n^{ q + (1-\gamma)(-1-q) }
    = 2^{1-\gamma} \inp^{\nicefrac{1}{2}} n^{ \gamma q + \gamma - 1 } \\
    &\le \max \bigl\{ 2^{ 1 - \gamma } \inp^{ \nicefrac{1}{2} }, \inp^{ \nicefrac{\gamma}{2} } \bigr\} n^{ \gamma q + \gamma - 1 }.
  \end{split}
  \end{equation}
  \Moreover \cref{eq:thm:bound:hoelder:sob:varepsilon_A} and the Cauchy Schwarz inequality \prove that for all
  $ x \in A_n $, $ y \in [\ms a, \ms b]^{\inp} \backslash A_n $ it holds that
  \begin{equation}
  \begin{split}
    &n^{-q} d^{ -\nicefrac{1}{2} } \bigl( n^q \scalprod{ u }{ x } + n^{-1} - n^q \inp \ms b \bigr) \\
    &\le n^{-q} d^{ -\nicefrac{1}{2} } \Bigl( \bigl[ n^q \scalprod{ u }{ x } + n^{-1} - n^q \inp \ms b \bigr]
      - \bigl[ n^q \scalprod{ u }{ y } + n^{-1} - n^q \inp \ms b \bigr] \Bigr] \\
    &= n^{-q} d^{ -\nicefrac{1}{2} } \bigl( n^q \scalprod{ u }{ x - y } \bigr)
    = d^{ -\nicefrac{1}{2} }\scalprod{ u }{ x - y }
    \le d^{ -\nicefrac{1}{2} } \eucnorm{ u } \eucnorm{ x - y }
    = \eucnorm{ x - y }.
  \end{split}
  \end{equation}
  Combining this with \cref{eq:thm:bound:hoelder:sob:realization:theta} \proves that for all $ n \in \N $, $ x \in A_n $,
  $ y \in [\ms a, \ms b]^{\inp} \backslash A_n $ it holds that
  \begin{equation}
  \begin{split}
    \tfrac{ \abs{ \mc N^{\theta(n)} (x) - \mc N^{\theta(n)} (y) } }{ \eucnorm{ x - y }^{\gamma} }
    &= \tfrac{ n^q \scalprod{ u }{ x } + n^{-1} - n^q \inp \ms b }{ \eucnorm{ x - y }^{\gamma} }
    \le \tfrac{ n^q \scalprod{ u }{ x } + n^{-1} - n^q \inp \ms b }
      { n^{ -\gamma q } \inp^{ - \nicefrac{\gamma}{2} } [ n^q \scalprod{ u }{ x } + n^{-1} - n^q \inp \ms b ]^{\gamma} } \\
    &= n^{ \gamma q } \inp^{ \nicefrac{\gamma}{2} } [ n^q \scalprod{ u }{ x } + n^{-1} - n^q \inp \ms b ]^{1-\gamma}
    \le n^{ \gamma q } \inp^{ \nicefrac{\gamma}{2} } \bigl[ n^{-1} \bigr]^{1-\gamma} \\
    &= \inp^{ \nicefrac{\gamma}{2} } n^{ \gamma q + \gamma - 1 }  
    \le \max \bigl\{ 2^{ 1 - \gamma } \sqrt{\inp}, d^{ \nicefrac{\gamma}{2} } \bigr\} n^{ \gamma q + \gamma - 1 }.
  \end{split}
  \end{equation}
  This, \cref{eq:thm:bound:hoelder:sob:sup:1}, and the fact that for all $ n \in \N $, $ x, y \in [\ms a, \ms b]^{\inp} \backslash A_n $
  it holds that $ \abs{ \mc N^{\theta(n)} (x) - \mc N^{\theta(n)} (y) } = 0 $ \prove that for all
  $ n \in \N $, $ x, y \in [\ms a, \ms b]^{\inp} $ with $ x \neq y $ it holds that
  \begin{equation}
    \tfrac{ \abs{ \mc N^{\theta(n)} (x) - \mc N^{\theta(n)} (y) } }{ \eucnorm{ x - y }^{\gamma} }
    \le \max \bigl\{ 2^{ 1 - \gamma } \sqrt{\inp}, d^{ \nicefrac{\gamma}{2} } \bigr\} n^{ \gamma q + \gamma - 1 }.
  \end{equation}
  Combining this with \cref{eq:thm:bound:hoelder:sob:realization:bound} \proves that for all $ n \in \N $ it holds that
  \begin{equation}
  \begin{split}
    \hoeldernorm{ \mc N^{\theta(n)} }{\gamma}{\ms b}
    &= \sup\nolimits_{ x \in [\ms a, \ms b]^{\inp} } \abs{ \mc N^{\theta(n)} (x) }
      + \sup\nolimits_{ x, y \in [\ms a, \ms b]^{\inp}, \, x \neq y } \tfrac{ \abs{ \mc N^{\theta(n)} (x) - \mc N^{\theta(n)} (y) } }
      { \eucnorm{ x - y }^{\gamma} } \\
    &\le n^{-1} + \max \bigl\{ 2^{ 1 - \gamma } \sqrt{\inp}, d^{ \nicefrac{\gamma}{2} } \bigr\} n^{ \gamma q + \gamma - 1 }
  \end{split}
  \end{equation}
  \cfload.
  The assumption that $ \gamma q < 1 - \gamma $ \hence \proves that $ \gamma q + \gamma - 1 < 0 $ and
  \begin{equation}\label{eq:lem:counterpart:hoelder:limit}
    \lim\nolimits_{ n \to \infty } \hoeldernorm{ \mc N^{\theta(n)} }{\gamma}{\ms b} = 0.
  \end{equation}
  \Moreover \cref{eq:thm:bound:hoelder:sob:realization:bound} \proves that for all $ n \in \N $ it holds that
  \begin{equation}\label{eq:lem:counterpart:lp}
    \biggl[ \int_{ [\ms a, \ms b]^{\inp} } \abs{ \mc N^{\theta(n)} (x) }^p \, \diff x \biggr]^{\nicefrac{1}{p}}
    \le \biggl[ \int_{ [\ms a, \ms b]^{\inp} } [ n^{-1} ]^p \, \diff x \biggr]^{\nicefrac{1}{p}}
    = \Bigl[ n^{-p} ( \ms b - \ms a )^{\inp} \Bigr]^{\nicefrac{1}{p}}
    = n^{-1} ( \ms b - \ms a )^{\nicefrac{\inp}{p}}.
  \end{equation}
  \Moreover Fubini's theorem and \cref{eq:thm:bound:hoelder:sob:realization:theta} \prove that for all $ n \in \N $ it holds that
  \begin{equation}\label{eq:lem:counterpart:double:integral}
  \begin{split}
    &\int_{ [\ms a, \ms b]^{\inp} } \int_{ [\ms a, \ms b]^{\inp} }
      \tfrac{ \abs{ \mc N^{\theta(n)} (x) - \mc N^{\theta(n)} (y) }^p }{ \eucnorm{ x - y }^{ \gamma p + \inp } } \, \diff x \, \diff y \\
    &= \int_{ A_n } \int_{ A_n }
      \tfrac{ \abs{ \mc N^{\theta(n)} (x) - \mc N^{\theta(n)} (y) }^p }{ \eucnorm{ x - y }^{ \gamma p + \inp } } \, \diff x \, \diff y
      + \int_{ A_n } \int_{ [\ms a, \ms b]^{\inp} \backslash A_n }
      \tfrac{ \abs{ \mc N^{\theta(n)} (x) - \mc N^{\theta(n)} (y) }^p }{ \eucnorm{ x - y }^{ \gamma p + \inp } } \, \diff x \, \diff y \\
    &\quad + \int_{ [\ms a, \ms b]^{\inp} \backslash A_n } \int_{ A_n }
      \tfrac{ \abs{ \mc N^{\theta(n)} (x) - \mc N^{\theta(n)} (y) }^p }{ \eucnorm{ x - y }^{ \gamma p + \inp } } \, \diff x \, \diff y \\
    &\quad + \int_{ [\ms a, \ms b]^{\inp} \backslash A_n } \int_{ [\ms a, \ms b]^{\inp} \backslash A_n }
      \tfrac{ \abs{ \mc N^{\theta(n)} (x) - \mc N^{\theta(n)} (y) }^p }{ \eucnorm{ x - y }^{ \gamma p + \inp } } \, \diff x \, \diff y \\
    &= \int_{ A_n } \int_{ A_n }
      \tfrac{ \abs{ \mc N^{\theta(n)} (x) - \mc N^{\theta(n)} (y) }^p }{ \eucnorm{ x - y }^{ \gamma p + \inp } } \, \diff x \, \diff y
      + 2 \int_{ A_n } \int_{ [\ms a, \ms b]^{\inp} \backslash A_n }
      \tfrac{ \abs{ \mc N^{\theta(n)} (x) - \mc N^{\theta(n)} (y) }^p }{ \eucnorm{ x - y }^{ \gamma p + \inp } } \, \diff x \, \diff y \\
    &\le 3 \int_{ A_n } \int_{ [\ms a, \ms b]^{\inp} }
      \tfrac{ \abs{ \mc N^{\theta(n)} (x) - \mc N^{\theta(n)} (y) }^p }{ \eucnorm{ x - y }^{ \gamma p + \inp } } \, \diff x \, \diff y.
  \end{split}
  \end{equation}
  \Moreover the fact that for all $ x, y \in [\ms a, \ms b]^{\inp} $ it holds that
  $ \eucnorm{ x - y } \le \inp^{\nicefrac{1}{2}} ( \ms b - \ms a ) $ and \cref{lem:integral:gamma} (applied with
  $ d \is \inp $, $ r \is \inp^{\nicefrac{1}{2}} ( \ms b - \ms a ) $, $ \gamma \is (1-\gamma)p - \inp $
  in the notation of \cref{lem:integral:gamma})
  \prove that for all $ n \in \N $, $ y \in [\ms a, \ms b]^{\inp} $ it holds that
  \begin{equation}\label{eq:thm:bound:hoelder:sob:integral:diff}
  \begin{split}
    &\int_{ [\ms a, \ms b]^{\inp} } \eucnorm{ x - y }^{ (1-\gamma)p - \inp } \, \diff x
    = \int_{ \smallbigcup_{ x \in [\ms a, \ms b]^{\inp} } \{ x - y \} } \eucnorm{ z }^{ (1-\gamma)p - \inp } \, \diff z \\
    &\le \int_{ \{ x \in \R^d \colon \eucnorm{ x } \le \inp^{\nicefrac{1}{2}} ( \ms b - \ms a ) \} }
      \eucnorm{ z }^{ (1-\gamma)p - \inp } \, \diff z
    = \tfrac{ 2 \pi^{ \nicefrac{\inp}{2} } }{ (1-\gamma)p \Gamma( \nicefrac{\inp}{2} ) }
      \bigl[ \inp^{\nicefrac{1}{2}} ( \ms b - \ms a ) \bigr]^{ (1-\gamma)p } \\
    &= \tfrac{ 2 \pi^{ \nicefrac{\inp}{2} } ( \ms b - \ms a )^{ (1-\gamma)p } }{ (1-\gamma)p \Gamma( \nicefrac{\inp}{2} ) }
        \inp^{ \frac{(1-\gamma)p}{2} }.
  \end{split}
  \end{equation}
  \Moreover \cref{eq:thm:bound:hoelder:sob:radius:A} \proves that for all $ n \in \N $ it holds that
  $ A_n \subseteq \{ x \in \R^d \colon \eucnorm{ x - v } \le n^{-1-q} \} $.
  \cref{lem:integral:gamma} (applied for every $ n \in \N $ with
  $ d \in \inp $, $ r \is n^{-1-q} $, $ \gamma \is 0 $
  in the notation of \cref{lem:integral:gamma})
  \hence \proves that for all $ n \in \N $ it holds that
  \begin{equation}
    \int_{ A_n } 1 \, \diff y
    \le \int_{ \{ x \in \R^d \colon \eucnorm{ x - v } \le n^{-1-q} \} } 1 \, \diff y
    = \int_{ \{ x \in \R^d \colon \eucnorm{ x } \le n^{-1-q} \} } 1 \, \diff y
    = \tfrac{ 2 \pi^{ \nicefrac{\inp}{2} } }{ \Gamma( \nicefrac{\inp}{2} ) } n^{ -\inp - \inp q}.
  \end{equation}
  Combining this with \cref{eq:thm:bound:hoelder:sob:realization:diff} and \cref{eq:thm:bound:hoelder:sob:integral:diff} \proves that
  for all $ n \in \N $ it holds that
  \begin{equation}
  \begin{split}
    &3 \int_{ A_n } \int_{ [\ms a, \ms b]^{\inp} }
      \tfrac{ \abs{ \mc N^{\theta(n)} (x) - \mc N^{\theta(n)} (y) }^p }{ \eucnorm{ x - y }^{ \gamma p + \inp } } \, \diff x \, \diff y
    \le 3 \int_{ A_n } \int_{ [\ms a, \ms b]^{\inp} }
      \tfrac{ [ n^q \inp^{\nicefrac{1}{2}} \eucnorm{ x - y } ]^p }{ \eucnorm{ x - y }^{ \gamma p + \inp } } \, \diff x \, \diff y \\
    &= 3 n^{pq} \inp^{\nicefrac{p}{2}} \int_{ A_n } \int_{ [\ms a, \ms b]^{\inp} }
      \eucnorm{ x - y }^{ (1-\gamma)p - \inp }  \, \diff x \, \diff y
    \le 3 n^{pq} \inp^{\nicefrac{p}{2}} \int_{ A_n }
      \tfrac{ 2 \pi^{ \nicefrac{\inp}{2} } ( \ms b - \ms a )^{ (1-\gamma)p } }{ (1-\gamma)p \Gamma( \nicefrac{\inp}{2} ) }
        \inp^{ \frac{(1-\gamma)p}{2} } \, \diff y \\
    &= 3 n^{pq} \inp^{\nicefrac{p}{2}} \tfrac{ 2 \pi^{ \nicefrac{\inp}{2} } ( \ms b - \ms a )^{ (1-\gamma)p } }
        { (1-\gamma)p \Gamma( \nicefrac{\inp}{2} ) } \inp^{ \frac{(1-\gamma)p}{2} } \int_{ A_n } 1 \, \diff y
    \le n^{pq} \tfrac{ 6 \pi^{ \nicefrac{\inp}{2} } ( \ms b - \ms a )^{ (1-\gamma)p } }
        { (1-\gamma)p \Gamma( \nicefrac{\inp}{2} ) } \inp^{ p - \frac{\gamma p}{2} }
        \tfrac{ 2 \pi^{ \nicefrac{\inp}{2} } }{ \Gamma( \nicefrac{\inp}{2} ) } n^{ -\inp - \inp q} \\
    &= \tfrac{ 12 \pi^{ \inp } ( \ms b - \ms a )^{ (1-\gamma)p } }{ (1-\gamma)p [ \Gamma( \nicefrac{\inp}{2} ) ]^2 }
        \inp^{ p - \frac{\gamma p}{2} } n^{ (p-\inp)q - d }.
  \end{split}
  \end{equation}
  This and \cref{eq:lem:counterpart:double:integral} \prove that for all $ n \in \N $ it holds that
  \begin{equation}
  \begin{split}
    &\biggl[ \int_{ [\ms a, \ms b]^{\inp} } \int_{ [\ms a, \ms b]^{\inp} }
      \tfrac{ \abs{ \mc N^{\theta(n)} (x) - \mc N^{\theta(n)} (y) }^p }{ \eucnorm{ x - y }^{ \gamma p + \inp } } \, \diff x \, \diff y
      \biggr]^{\nicefrac{1}{p}}
    \le \biggl[ 3 \int_{ A_n } \int_{ [\ms a, \ms b]^{\inp} }
      \tfrac{ \abs{ \mc N^{\theta(n)} (x) - \mc N^{\theta(n)} (y) }^p }{ \eucnorm{ x - y }^{ \gamma p + \inp } } \, \diff x \, \diff y
      \biggr]^{\nicefrac{1}{p}} \\
    &\le \Bigl[ \tfrac{ 12 \pi^{ \inp } ( \ms b - \ms a )^{ (1-\gamma)p } }{ (1-\gamma)p [ \Gamma( \nicefrac{\inp}{2} ) ]^2 }
        \inp^{ p - \frac{\gamma p}{2} } \Bigr]^{\nicefrac{1}{p}} n^{ \frac{(p-\inp)q - d}{p} }.
  \end{split}
  \end{equation}
  Combining this and \cref{eq:lem:counterpart:lp} \proves that for all $ n \in \N $ it holds that
  \begin{equation}
  \begin{split}
    \sobnorm{ \mc N^{\theta(n)} }{\gamma}{p}
    &= \biggl[ \int_{ [\ms a, \ms b]^{\inp} } \abs{ \mc N^{\theta(n)} (x) }^p \, \diff x \biggr]^{\nicefrac{1}{p}}
      + \biggl[ \int_{ [\ms a, \ms b]^{\inp} } \int_{ [\ms a, \ms b]^{\inp} }
      \tfrac{ \abs{ \mc N^{\theta(n)} (x) - \mc N^{\theta(n)} (y) }^p }{ \eucnorm{ x - y }^{ \gamma p + \inp } } \, \diff x \, \diff y
      \biggr]^{\nicefrac{1}{p}} \\
    &\le n^{-1} ( \ms b - \ms a )^{\nicefrac{\inp}{p}}
      + \Bigl[ \tfrac{ 12 \pi^{ \inp } ( \ms b - \ms a )^{ (1-\gamma)p } }{ (1-\gamma)p [ \Gamma( \nicefrac{\inp}{2} ) ]^2 }
        \inp^{ p - \frac{\gamma p}{2} } \Bigr]^{\nicefrac{1}{p}} n^{ \frac{(p-\inp)q - d}{p} }
  \end{split}
  \end{equation}
  \cfload.
  The assumption that $ ( p - \inp ) q < \inp $ \hence \proves that $ \tfrac{ ( p - \inp ) q - d }{ p } < 0 $ and
  \begin{equation}\label{eq:lem:counterpart:sob:limit}
    \lim\nolimits_{ n \to \infty } \sobnorm{ \mc N^{\theta(n)} }{\gamma}{p} = 0.
  \end{equation}
  Combining this, \cref{eq:lem:counterpart:norm:vartheta}, and \cref{eq:lem:counterpart:hoelder:limit} \proves that for all
  $ \ms c \in (0,\infty) $ there exists $ n \in \N $ such that for all
  $ \vartheta \in \{ \eta \in \R^{\dimension} \colon \mc N^{\eta} = \mc N^{\theta(n)} \} $ it holds that
  \begin{equation}
    \eucnorm{ \vartheta } \ge 2^{ \nicefrac{1}{2} } \inp^{\nicefrac{1}{4}} n^{ \nicefrac{q}{2} } > \ms c
    \qqandqq
    \max \bigl\{ \hoeldernorm{ \mc N^{\theta(n)} }{\gamma}{\ms b}, \sobnorm{ \mc N^{\theta(n)} }{\gamma}{p} \bigr\} < \ms c^{-1}.
  \end{equation}
  \Hence that for all $ \ms c \in (0,\infty) $ there exists $ \theta \in \R^{\dimension} $ such that for all
  $ \vartheta \in \{ \eta \in \R^{\dimension} \colon \mc N^{\eta} = \mc N^{\theta} \} $ it holds that
  \begin{equation}
    \eucnorm{ \vartheta } > \ms c
    \qqandqq
    \max \bigl\{ \hoeldernorm{ \mc N^{\theta} }{\gamma}{\ms b}, \sobnorm{ \mc N^{\theta} }{\gamma}{p} \bigr\} < \ms c^{-1}.
  \end{equation}
\end{cproof}

\cfclear
\begin{cor}\label{cor:hoelder:sob:sum}
  Assume \cref{setting:main} and let $ n \in \N $, $ \gamma_1, \gamma_2, \ldots, \gamma_n \in [0,1) $,
  $ v_1, v_2, \ldots, v_n \in [\ms a, \ms b] $, $ p_1, p_2, \ldots, p_n \in [1,\infty) $,
  $ \delta_1, \delta_2, \ldots, \delta_n \in [0,\infty) $.
  Then for all $ \ms c \in \R $ there exists $ \theta \in \R^{\dimension} $ such that for all
  $ \vartheta \in \{ \eta \in \R^{\dimension} \colon \mc N^{\eta} = \mc N^{\theta} \} $ it holds that
  \begin{equation}\label{eq:cor:hoelder:sob:sum}
    \eucnorm{ \vartheta } > \ms c \Bigl( \smallsum\nolimits_{i=1}^n \hoeldernorm{ \mc N^{\theta} }{\gamma_i}{v_i}^{\delta_i} \Bigr)
    \qqandqq
    \eucnorm{ \vartheta } > \ms c \Bigl( \smallsum\nolimits_{i=1}^n \sobnorm{ \mc N^{\theta} }{\gamma_i}{p_i}^{\delta_i} \Bigr)
  \end{equation}
  \cfout.
\end{cor}

\begin{cproof}{cor:hoelder:sob:sum}
  Throughout this proof let $ \lambda \in [0,1) $, $ q \in [1, \infty) $ satisfy
  \begin{equation}\label{eq:cor:hoelder:norm:gamma}
    \lambda = \tfrac{ 1 + \max \{ \gamma_1, \gamma_2, \ldots, \gamma_n \} }{ 2 }
    \qqandqq
    q = 1 + \max \{ p_1, p_2, \ldots, p_n \}.
  \end{equation}
  \Nobs that \cref{lem:hoeldernorm:inequality} (applied for every $ i \in \{ 1, 2, \ldots, n \} $ with
  $ d \is \inp $, $ \ms a \is \ms a $, $ \ms b \is \ms b $, $ \gamma \is \gamma_i $, $ \lambda \is \lambda $, $ v \is v_i $, $ w \is \ms b $
  in the notation of \cref{lem:hoeldernorm:inequality})
  and \cref{lem:sobnorm:inequality} (applied for every $ i \in \{ 1, 2, \ldots, n \} $ with
  $ d \is \inp $, $ \ms a \is \ms a $, $ \ms b \is \ms b $, $ \gamma \is \gamma_i $, $ \lambda \is \lambda $, $ p \is p_i $, $ q \is q $
  in the notation of \cref{lem:sobnorm:inequality})
  \prove that for all $ i \in \{ 1, 2, \ldots, n \} $ there exist $ \ms c, \ms C \in (0,\infty) $ such that for all
  $ \theta \in \R^{\dimension} $ it holds that
  \begin{equation}
    \hoeldernorm{ \mc N^{\theta} }{ \gamma_i }{ v_i } \le \ms c \hoeldernorm{ \mc N^{\theta} }{ \lambda }{ \ms b }
    \qqandqq
    \sobnorm{ \mc N^{\theta} }{ \gamma_i }{ p_i } \le \ms C \sobnorm{ \mc N^{\theta} }{ \lambda }{ q }
  \end{equation}
  \cfload.
  \Hence that there exists $ \ms C \in (0,\infty) $ which satisfies that for all $ i \in \{ 1, 2, \ldots, n \} $, $ \theta \in \R^{\dimension} $
  it holds that
  \begin{equation}\label{eq:lem:hoelder:sob:sum:C}
    \hoeldernorm{ \mc N^{\theta} }{ \gamma_i }{ v_i } \le \ms C \hoeldernorm{ \mc N^{\theta} }{ \lambda }{ \ms b }
    \qqandqq
    \sobnorm{ \mc N^{\theta} }{ \gamma_i }{ p_i } \le \ms C \sobnorm{ \mc N^{\theta} }{ \lambda }{ q }.
  \end{equation}
  \Moreover \cref{thm:bound:hoelder:sob} (applied for every $ \ms c \in \R $ with
  $ \ms c \is \max\{ \ms c n, \ms C \} $, $ \gamma \is \lambda $, $ p \is q $
  in the notation of \cref{thm:bound:hoelder:sob})
  \proves that for all $ \ms c \in \R $ there exists $ \theta \in \R^{\dimension} $ such that for all
  $ \vartheta \in \{ \eta \in \R^{\dimension} \colon \mc N^{\eta} = \mc N^{\theta} \} $ it holds that
  \begin{equation}
    \eucnorm{ \vartheta } > \max\{ \ms c n, \ms C \} \ge \ms c n
    \qandq
    \max \bigl\{ \hoeldernorm{ \mc N^{\theta} }{ \lambda }{ \ms b }, \sobnorm{ \mc N^{\theta} }{ \lambda }{ q } \bigr\}
      < [ \max\{ \ms c n, \ms C \} ]^{-1} \le \ms C^{-1}
  \end{equation}
  \cfload.
  Combining this and \cref{eq:lem:hoelder:sob:sum:C} \proves that for all $ \ms c \in [0,\infty) $ there exists
  $ \theta \in \R^{\dimension} $ such that for all $ \vartheta \in \{ \eta \in \R^{\dimension} \colon \mc N^{\eta} = \mc N^{\theta} \} $
  it holds that
  \begin{equation}
    \ms c \Bigl( \smallsum_{i=1}^n \hoeldernorm{ \mc N^{\theta} }{ \gamma_i }{ v_i }^{ \delta_i } \Bigr)
    \le \ms c \Bigl( \smallsum_{i=1}^n \bigl[ \ms C \hoeldernorm{ \mc N^{\theta} }{ \lambda }{ \ms b } \bigr]^{ \delta_i } \Bigr)
    \le \ms c \Bigl( \smallsum_{i=1}^n 1^{ \delta_i } \Bigr)
    = \ms c n
    < \eucnorm{ \vartheta }
  \end{equation}
  and
  \begin{equation}
    \ms c \Bigl( \smallsum_{i=1}^n \sobnorm{ \mc N^{\theta} }{ \gamma_i }{ p_i }^{ \delta_i } \Bigr)
    \le \ms c \Bigl( \smallsum_{i=1}^n \bigl[ \ms C \sobnorm{ \mc N^{\theta} }{ \lambda }{ q } \bigr]^{ \delta_i } \Bigr)
    \le \ms c \Bigl( \smallsum_{i=1}^n 1^{ \delta_i } \Bigr)
    = \ms c n
    < \eucnorm{ \vartheta }.
  \end{equation}
  \Hence that for all $ \ms c \in \R $ there exists $ \theta \in \R^{\dimension} $ such that for all
  $ \vartheta \in \{ \eta \in \R^{\dimension} \colon \mc N^{\eta} = \mc N^{\theta} \} $ it holds that
  \begin{equation}
    \eucnorm{ \vartheta } > \ms c \Bigl( \smallsum\nolimits_{i=1}^n \hoeldernorm{ \mc N^{\theta} }{\gamma_i}{v_i}^{\delta_i} \Bigr)
    \qqandqq
    \eucnorm{ \vartheta } > \ms c \Bigl( \smallsum\nolimits_{i=1}^n \sobnorm{ \mc N^{\theta} }{\gamma_i}{p_i}^{\delta_i} \Bigr).
  \end{equation}
\end{cproof}

\subsection*{Acknowledgements}
This work has been partially funded by the European Union (ERC, MONTECARLO, 101045811). The views and the opinions expressed in this work are however those of the authors only and do not necessarily reflect those of the European Union or the European Research Council (ERC). Neither the European Union nor the granting authority can be held responsible for them. Moreover, this work has been partially funded by the Deutsche Forschungsgemeinschaft (DFG, German Research Foundation) under Germany's Excellence Strategy EXC 2044--390685587, Mathematics M\"unster: Dynamics--Geometry--Structure.

\bibliographystyle{acm}
\bibliography{references}

\end{document}